\documentclass{article}

    \PassOptionsToPackage{numbers, compress}{natbib}

 \usepackage[preprint]{neurips_2019}

\usepackage[utf8]{inputenc} 
\usepackage[T1]{fontenc}    
\usepackage{hyperref}       
\usepackage{url}            
\usepackage{booktabs}       
\usepackage{amsfonts}       
\usepackage{nicefrac}       
\usepackage{microtype}      

\usepackage{graphicx}
\usepackage{subfigure}
\usepackage{float}
\usepackage{amsmath}
\usepackage{xcolor}
\usepackage{multirow}
\usepackage{bm}
\usepackage{epstopdf}
\usepackage{tikz}
    \usetikzlibrary{fit, positioning, shapes, snakes, calc}
    \usepackage{relsize}
\usepackage{algorithm}
\usepackage{algorithmic}
\usepackage{wrapfig}

\renewcommand{\algorithmiccomment}[1]{  $\triangleright${\textit{\small{{#1}}}}}

\newcommand{\tth}{\text{th}}
\newcommand{\iidsim}{\overset{iid}{\sim}}
\newcommand{\indsim}{\overset{ind}{\sim}}
\newcommand{\Norm}{\text{N}} 
\newcommand{\Gam}{\text{Gam}} 
\newcommand{\DP}{\text{DP}} 
\newcommand{\EDP}{\text{EDP}} 
\newcommand{\GP}{\text{GP}} 
\newcommand{\1}{\textbf{1}} 

\usepackage{amsmath}

\makeatother

\urldef\adniinfo\url{www.adni-info.org}
\urldef\adnidream\url{https://www.synapse.org/#!Synapse:syn2290704/wiki/60828}
\urldef\finalresults\url{https://www.synapse.org/#!Synapse:syn2290704/wiki/70719}
\urldef\adni\url{adni.loni.ucla.edu}
\urldef\adniinvest\url{http://adni.loni.ucla.edu/wp-content/uploads/how_to_apply/ADNI_Acknowledgement_List.pdf}
\urldef\fnih\url{www.fnih.org}
\urldef\adniContribs\url{adni-info.org/Scientists/ADNISponsors.aspx}

\title{Enriched Mixtures of Gaussian Process Experts}

\author{%
  Charles W.L. Gadd \\
  School of Engineering\\ 
  University of Warwick\\
   Coventry, United Kingdom\\
  \texttt{cwlgadd@gmail.com } \\
   \And
   Sara Wade\thanks{Equal contribution}\\
School of Mathematics\\
 University of Edinburgh\\
  Edinburgh, United Kingdom\\
  \texttt{sara.wade@ed.ac.uk} \\
   \And
  Alexis Boukouvalas\\
PROWLER.io\\
 Cambridge, United Kingdom\\
  \texttt{alexis@prowler.io} \\
}

\begin{document}

\maketitle

\begin{abstract}
Mixtures of experts probabilistically divide the input space into regions, where the assumptions of each expert, or conditional model, need only hold locally. Combined with Gaussian process (GP) experts, this results in a powerful and highly flexible model. We focus on alternative mixtures of GP experts, which  model the joint distribution of the inputs and targets explicitly. We highlight issues of this approach in multi-dimensional input spaces, namely,  poor scalability and the need for an unnecessarily large number of experts, degrading the predictive performance and increasing uncertainty. We construct a novel model to address these issues through a nested partitioning scheme that automatically infers the number of components at both levels. Multiple response types are accommodated through a generalised GP framework, while multiple input types are included through a factorised exponential family structure. 
We show the effectiveness of our approach in estimating a parsimonious probabilistic description of both  synthetic data of increasing dimension and an Alzheimer's challenge dataset.
\end{abstract}

\section{Introduction}\label{sec:intro}

The Gaussian process \citep{Ras05} is a powerful and popular prior for nonparametric regression, 
due to its flexibility, analytic tractability, and interpretable hyperparameters.  The GP assumes that the unknown function evaluated at any finite set of inputs has a Gaussian distribution with consistent parameters. It is fully specified by a mean function and symmetric positive definite covariance (or kernel) function, which together encapsulate any prior knowledge and/or assumptions of the regression function, such as smoothness and periodicity. 
While GP regression has been successfully applied to various problems,
  it only allows for flexibility in the regression function, assuming homoskedastic Gaussian errors. Many datasets further require flexibility in the errors,
such as multi-modality or different variances across the input space. Moreover, for computational purposes, a stationary assumption of the GP is typically employed, which is inappropriate in many examples, by limiting the ability to recover changing behaviour of the function across the input space, e.g. different smoothness levels. 
	
	Density regression refers to the general problem of estimating the conditional density of the targets across the input space, or equivalently, flexible estimation of both the regression function and input-dependent error distribution.  
	Mixtures of experts \cite{jacobs1991adaptive}  address the density regression problem by probabilistically partitioning the input space. Each expert is a conditional model, 
	and a gating network maps experts to local regions of the input space. 
	Scalability is enhanced since each expert considers only its local region, and simplifying assumptions of the regression function need only hold locally in each region.  
	Experts may range from simple linear models to flexible non-linear approaches. \citet{tresp2001mixtures} combines mixtures of experts with GPs, resulting in a flexible nonparametric approach for both the experts and gating networks. 
GP experts allow the model to infer different behaviours, such as smoothness and variability, in local regions of the input space. 
	In this work, we focus on alternative mixtures of GP experts \citep{meeds2006alternative}, which explicitly model the joint distribution of the inputs and targets. We highlight issues of this approach for multi-dimensional inputs, namely, poor scalability and the need for an unnecessary number of experts, and we construct a novel model to address these issues.
	
	The paper is organised as follows. Related work is reviewed in Section \ref{sec:review}. In Section \ref{sec:eMoE}, we construct a novel mixture of generalised GP experts, that extends  \citet{meeds2006alternative} for multiple response and input types and additionally utilises a nested partitioning scheme to improve prediction and uncertainty quantification. Posterior inference is described in Section \ref{sec:enrichedGibbs}. Section \ref{sec:examples} illustrates the benefits in a non-linear toy example  and a case study to predict cognitive decline in Alzheimer's.  

	\section{Related work}\label{sec:review}

	Infinite mixtures of GP experts  \cite{rasmussen2002infinite} 
	allow the number of experts to be determined by the data and
	  grow unboundedly as more data points are observed. 	 
	 The alternative infinite mixture of GP experts \cite{meeds2006alternative} models the joint distribution of the inputs and targets explicitly, 
	 and advantages include
	 the ability to handle missing data and answer inverse problems, as well as simplified computations, relying on established algorithms for infinite mixtures of exchangeable data \cite[e.g.][]{neal2000markov}. 
\citet{meeds2006alternative} 
assume a local multivariate Gaussian distribution for the inputs, 
and in multi-dimensions, complexity in the marginal distribution of the inputs 
may lead to the creation of an unnecessary number of experts, degrading the predictive performance and increasing uncertainty, due to small sample sizes for each expert. 
 \citet{yuan2009variational} remove this constraint by using a Gaussian mixture for the local input density; however, a finite approximation to the infinite mixture is used at both levels. Moreover, the local 
multivariate Gaussian input model 
scales poorly with the input dimension $D$  due to the computational cost of dealing with the full $D$ by $D$ matrix. 

The treed-GP \citep[TGP,][]{gramacy2008bayesian} is another example of a mixture of GP experts, where the input space is partitioned into axis-aligned rectangular regions. 
However, this axis-aligned approach also scales poorly in multi-dimensional input spaces, again leading to an unnecessarily large number of experts. 
More flexible partitioning approaches exist, such as Voronoi tessellations \cite{pope2018modelling}; 
however, inference is more computationally expensive, especially as the input dimension increases.

Infinite mixtures of generalised linear experts \cite{hannah2011dirichlet} provide a unifying framework 
to model multiple response types. 
In this linear setting, the problems associated with an overly large number of experts is highlighted in \citet{wade2014improving}, where a loss of predictive accuracy and increased uncertainty is demonstrated, particularly as $D$ increases.  Due to the greater flexibility of GPs over linear experts, these problems are exacerbated for mixtures of GP experts. In the following section, we construct a novel model to overcome these issues.

\section{Enriched mixtures of generalised GP experts}\label{sec:eMoE}

A mixture model for the joint density of the output $y \in \mathcal{Y}$ and $D$-dimensional input $x \in \mathcal{X}$ assumes
\begin{align}
f_Q(y,x)= \int p(y|x,\theta)p(x|\psi)dQ(\theta,\psi). \label{eq:DPM}
\end{align}
The three key elements are 1) the \textbf{local expert} $p(y |x, \theta)$, a family of densities on $\mathcal{Y}$ for $\theta \in \Theta$; 2) the \textbf{local input model} $p(x|\psi)$, a family of densities on $\mathcal{X}$ for $\psi \in \Psi$; and 3) the \textbf{mixing measure} $Q$, a probability measure on $\Theta\times\Psi$.
  In the following, we define these three key elements for our model. 
  
    \subsection{Local experts}
We provide a framework for multiple output types by defining the local expert $p(y|x,\theta)$ to be an extension of the generalised linear model (GLM) used in \citet{hannah2011dirichlet}. Specifically, $p(y|x,\theta)$ 
belongs to the exponential family, which in canonical form assumes 
\begin{align*}
p(y|x,\theta)= \exp\left( \frac{y\eta -b(\eta)}{a(\phi)} + c(y, \phi) \right).
\end{align*}
The functions $a$, $b$, and $c$ are known and specific to the exponential family; $\phi$ is the scale parameter; and  $\eta$ is the canonical parameter with $ b'(\eta) = \mu(x) =\mathbb{E}[y|x]$ and $g(\mu(x))=m(x)$, where $g$ is a chosen link function that maps $\mu(x)$  to the real line. 
In GLMs \citep{MN89}, 
a linear function of $x$ determines the canonical parameter through a set of transformations, i.e. $m(x)=\alpha+x\beta$. 

Instead, we consider  a general non-linear function and assign a GP prior to the unknown function:
$$m(\cdot)|\beta_{0}, \lambda \sim \GP(\beta_{0},K_{\lambda}),$$ 
with constant mean function, $\mathbb{E}[ m(x)]=\beta_{0}$, and kernel function $K_\lambda$ with hyperparameters $\lambda$, defining the covariance of the function at any two inputs, $\text{Cov}[ m(x),m(x_*)]=K_\lambda(x,x_*)$. 
The parameters of this generalized Gaussian process \citep[GGP,][]{generalizedGP} are $\theta =(m(\cdot),\beta_{0},\lambda,\phi)$. 
In many examples, it is common to use a zero-centred GP, which is made appropriate by subtracting the overall mean from the response. However, in our case, we must include a constant mean, as the partitioning structure is unknown and the data cannot be centred for each expert. 
Additionally, by including $ \lambda$ in the set of mixing parameters $\theta$, we can recover non-stationary behaviour, e.g. different length-scales in local regions of the input space. 
The GGP experts used in Section \ref{sec:examples} are 1) \textbf{Gaussian} with identity link,
$p(y|x,\theta)= \Norm(y|m(x), \sigma^2);$ 
and 2) \textbf{Ordinal} with probit link for ordered categories $l=0,\ldots, L$, $$\mathbb{P}(y \leq l|x,\theta)=\Phi\left[\frac{\varepsilon_l-m(x)}{\sigma}\right],$$ and cutoffs $0=\varepsilon_0<\varepsilon_1 <\ldots < \varepsilon_{L-1} $, which may be fixed 
 due to the nonparametric nature of the model \citep{kottas2005}. 
The ordinal model can be equivalently formulated through a latent Gaussian response: 
$$ \tilde{y}|m(x), \sigma^2 \sim \Norm(m(x), \sigma^2), \quad  p(y|\tilde{y})=\left\lbrace \begin{array}{ll}
\1(\tilde{y} \leq 0) & \text{if } l=0 \\
\1( \varepsilon_{l-1} <\tilde{y} \leq \varepsilon_l) & \text{if } l=1,\ldots,L-1 \\
\1( \tilde{y}> \varepsilon_{L-1}) & \text{if } l=L 
\end{array} \right.,$$
with the ordered probit recovered after marginalisation of the latent $\tilde{y}$.
A list of GGP experts is provided in the Supplementary Material (SM), for studies with other output types.

\subsection{Local input models}
We assume a factorised exponential family structure for the local input model. Specifically, it factorises across $d=1,\ldots,D$, 
where each $p(x_d|\psi_d)$ belongs to the exponential family, that is, $$p(x_d|\psi_d)=\exp(\psi_d^{T}t_d(x_d)-a_d(\psi_d)+b_d(x_d)),$$
and $t_d$, $a_d$, and $b_d$ are known functions specified by the choice within the exponential family. 
The standard conjugate prior for $\psi$ assumes independence of $\psi_d$ across $d=1,\ldots,D$ with 
$$ \pi(\psi_d) \propto \exp(\psi_d^{T}\tau_d-\nu_d a_d(\psi_d)),$$
and parameters $\tau_d$ and $\nu_d$ determining the location and scale of the prior, respectively. 
In this conjugate setting, $\psi$ can be marginalised, and the local marginal and predictive likelihood of the inputs are available analytically (specific calculations are provided in the SM).
Examples used in Section \ref{sec:examples} are the  1) \textbf{Gaussian} for $x_{d} \in \mathbb{R}$, with local input model
 $ \Norm(x_{d}|u_{d},s_{d}^2)$; 2) \textbf{Categorical} for $x_{d}$ taking \textit{unordered} values $g=0,1,\ldots,G_d$, with local input model $\text{Cat}(x_d|\psi_d)$, where  $\psi_d=(\psi_{d,0},\ldots,\psi_{d,G_d})$ is a probability vector;
and 3) \textbf{Binomial} for $x_{d}$ taking \textit{ordered} values $g=0,1,\ldots,G_d$ with local input model $\text{Bin}(x_g| G_d,\psi_d)$ for $\psi_d \in (0,1)$.

Advantages of this factorised exponential form include improved scalability, inclusion of multiple input types, and richer parametrisation. 
Indeed, the mixtures of GP experts in \citep{meeds2006alternative,yuan2009variational,nguyen2014fast} consider only continuous inputs with a local multivariate Gaussian density and conjugate inverse Wishart prior on the covariance matrix. However, even for moderately large $D$, this approach becomes unfeasible. Specifically, the computational cost of dealing with the full covariance matrix is $O(D^3)$, which is reduced to $O(D)$ in our factorised form. Furthermore, \citet{CV01} highlight the poor parametrisation of the Wishart prior; in particular, there is a single parameter to control variability. In our model, flexibility of the conjugate prior is enhanced, as it includes a  scale parameter $\nu_d$ for each of the $D$ variances.
We emphasize that although the inputs are locally independent, globally, they may be dependent. For example, a highly-correlated, elliptically-shaped Gaussian can be accurately approximated with a mixture of several smaller spherical Gaussians.

\subsection{Mixing measure}

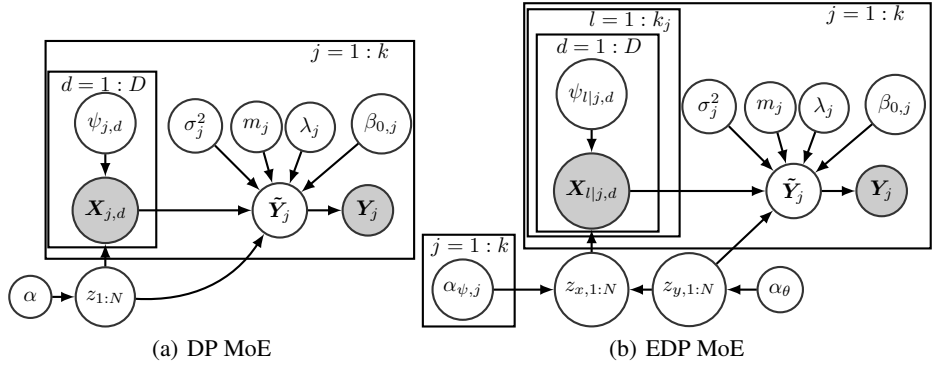
\begin{figure}[!t]
\begin{center}
\subfigure[DP MoE]{\begin{tikzpicture}[thick,scale=0.8, every node/.style={scale=0.8}]
\tikzstyle{main}=[circle, minimum size = 7mm, thick, draw =black!80, node distance = 3mm]
\tikzstyle{connect}=[-latex, thick]
\tikzstyle{box}=[rectangle, draw=black!100]
  \node[main, fill = white!100] (alpha)  {$\alpha$ };
  \node[main] (zir) [right=of alpha] {$z_{1:N}$ };
  \node[main, fill= black!20] (xir) [above=of zir] {$\bm{X}_{j,d}$};
  \node[main] (yrt) [right=15mm of xir] {$\bm{\tilde{Y}}_j$};
  \node[main] (psi_temp) [above =3mm of xir] {$\psi_{j,d}$};
 \node[main] (sigma) [above left=6mm and 6mm of yrt] {$\sigma^2_j$};
  \node[main] (theta) [above left=6mm and -2mm of yrt] {$m_j$};
  \node[main] (mu) [above right=6mm and -1mm of yrt] {$\lambda_j$};
  \node[main, fill= black!20] (yr) [right=4.5mm of yrt] {$\bm{Y}_j$};
  \node[main] (beta) [above right=6mm and 8mm of yrt] {$\beta_{0,j}$};
  \path (alpha) edge [connect] (zir)
        (zir) edge [connect] (xir)
        (xir) edge [connect] (yrt)
        (zir) edge [connect, bend right] (yrt)
        (psi_temp) edge [connect] (xir)
        (sigma) edge [connect] (yrt)
        (theta) edge [connect] (yrt)
        (mu) edge [connect] (yrt)
        (beta) edge [connect] (yrt)
        (yrt) edge [connect] (yr);
  \node[rectangle, inner sep=4.5mm,draw=black!100, fit= (xir) (psi_temp) , label={[xshift=0.25mm, yshift=-4.6mm]:$d=1:D$}, xshift=-0.5mm, yshift=1.4mm] {};
    \node[rectangle, inner sep=8mm,draw=black!100, fit= (xir) (yrt) (yr) (psi_temp) (sigma) (theta) (beta), label={[xshift=20mm, yshift=-5mm]:$j=1:k$}, xshift=-2mm, yshift=3mm] {};
\end{tikzpicture}\label{fig:gm_dp}}
\subfigure[EDP MoE]{\begin{tikzpicture}[thick,scale=0.8, every node/.style={scale=0.8}]
\tikzstyle{main}=[circle, minimum size = 7mm, thick, draw =black!80, node distance = 3mm]
\tikzstyle{connect}=[-latex, thick]
\tikzstyle{box}=[rectangle, draw=black!100]
  \node[main, fill = white!100] (alpha) {$\alpha_{\psi,j}$ };
  \node[main] (zir) [right=8mm of alpha] {$z_{x,1:N}$ };
\node[main] (zpsi) [right=of zir] {$z_{y,1:N}$ };
  \node[main, fill = white!100] (alphapsi) [right=4mm of zpsi] {$\alpha_\theta$ };
  \node[main, fill= black!20] (xir) [above=of zir] {$\bm{X}_{l|j,d}$};
   \node[main] (psi_temp) [above =3mm of xir] {$\psi_{l|j,d}$};
  \node[main] (yrt) [right=18mm of xir] {$\bm{\tilde{Y}}_j$};
  \node[main] (sigma) [above left=6mm and 6mm of yrt] {$\sigma^2_j$};
  \node[main] (theta) [above left=6mm and -2mm of yrt] {$m_j$};
  \node[main] (mu) [above right=6mm and -1mm of yrt] {$\lambda_j$};
  \node[main, fill= black!20] (yr) [right=4.5mm of yrt] {$\bm{Y}_j$};
  \node[main] (beta) [above right=6mm and 8mm of yrt] {$\beta_{0,j}$};
  \path 
        (alpha) edge [connect] (zir)
        (alphapsi) edge [connect] (zpsi)
        (zpsi) edge [connect] (zir)
        (zir) edge [connect] (xir)
        (xir) edge [connect] (yrt)
        (zpsi) edge [connect] (yrt)
        (psi_temp) edge [connect] (xir)
        (sigma) edge [connect] (yrt)
        (theta) edge [connect] (yrt)
        (mu) edge [connect] (yrt)
        (beta) edge [connect] (yrt)
        (yrt) edge [connect] (yr);
  \node[rectangle, inner sep=9mm,draw=black!100, fit= (xir) (yrt) (yr) (psi_temp) (sigma) (theta) (beta), label={[xshift=23mm, yshift=-5mm]:$j=1:k$}, xshift=-2mm, yshift=3mm] {};
  \node[rectangle, inner sep=7.5mm,draw=black!100, fit= (xir) (psi_temp), label={[xshift=4.5mm, yshift=-5.2mm]:$l=1:k_j$}, xshift=2mm, yshift=3.5mm] {};
  \node[rectangle, inner sep=5mm,draw=black!100, fit= (xir) (psi_temp), label={[xshift=0.25mm, yshift=-4.6mm]:$d=1:D$}, xshift=1mm, yshift=1.8mm] {};
    \node[rectangle, inner sep=3.5mm,draw=black!100, fit= (alpha), label={[xshift=0.25mm, yshift=-5mm]:$j=1:k$}, xshift=1mm, yshift=1.5mm] {};
\end{tikzpicture}\label{fig:gm_edp}}

\caption{Mixture of experts (MoE) with \ref{fig:gm_dp} DP prior and \ref{fig:gm_edp} EDP prior on the mixing measure $Q$. 
Here, $\bm{Y}_j$ and $\bm{\tilde{Y}}_j$ denote the observed and latent outputs in cluster $j$, with $\bm{X}_j$ denoting the inputs in cluster $j$ for the DP and $\bm{X}_{l|j}$ denoting the inputs in $x$-cluster $l$ nested in $y$-cluster $j$ for the EDP.}
\label{fig:gm_moe}
\end{center}
\vspace{-5mm}
\end{figure}
  
The Bayesian model is completed with a prior on the mixing measure $Q$, and the Dirichlet process \cite[DP,][]{Ferg} is a popular nonparametric choice. Indeed, it is utilised in \citep{rasmussen2002infinite,meeds2006alternative,hannah2011dirichlet}, among many others. Instead, we propose to use the enriched Dirichlet process  \citep[EDP,][]{wade2011} and highlight its advantages for improved prediction, better uncertainty quantification, and more interpretable clustering. 

\paragraph{Dirichlet process.} The parameters of the DP consist of the concentration parameter $\alpha>0$ and the base measure $Q_0$, a probability measure on $\Theta\times\Psi$. 
The DP is discrete with probability one, and realisations place positive mass on a countably infinite number of atoms. When utilised as a prior for the mixing measure $Q \sim \DP(\alpha, Q_0)$, this implies a countably infinite mixture for the joint density in  \eqref{eq:DPM}.
For $N$ data points $(y_n,x_n)$, $n=1,\ldots, N$, 
this induces a random partition of the data points into clusters. Introducing the latent variable $z_n$ denoting the cluster allocation of data point $n$, in order of appearance, and the parameters $(\theta_j,\psi_j)$ denoting the parameters of the $j^{\tth}$ observed cluster, the mixing measure $Q$ can be marginalised. In this case, the model can be expressed as
\begin{align*}
(y_n,x_n)|z_n=j,\theta_j,\psi_{j} &\indsim p(y_n|x_n,\theta_j)p(x_n|\psi_{j}), 
\end{align*}
where $(\theta_j,\psi_j) \iidsim Q_0$.
 The law of allocation variables is defined by the predictive distributions \citep{blackwell1973ferguson}: 
\begin{align*}
z_{N+1}|z_{1:N} \sim \frac{\alpha}{\alpha+N} \delta_{k+1}+ \sum_{j=1}^{k} \frac{N_{j}}{\alpha+N} \delta_{j},
 \end{align*}
where  $k$ is the number of clusters and $N_{j}$ is the number of data points allocated to cluster $j$. In this setting, the number of clusters is determined by and can grow with the data. 

\paragraph{Enriched Dirichlet process.}
The EDP defines a prior for the joint measure $Q$ on $\Theta \times \Psi$ by decomposing it in terms of the marginal  $Q_\theta$ and conditionals $Q_{\psi|\theta}(\cdot|\theta)$. 
The parameters consist of the base measure $Q_0$ on $\Theta \times \Psi$; a concentration parameter $\alpha_\theta$; and a collection of concentration parameters $\alpha_{\psi}(\theta)$ for  $\theta \in \Theta$. The EDP assumes 1) $ Q_\theta \sim \DP(\alpha_\theta Q_{0\,\theta})$; 2) $Q_{\psi|\theta}(\cdot|\theta) \sim \DP(\alpha_\psi(\theta) Q_{0\,\psi|\theta}(\cdot|\theta))$ for all $\theta \in \Theta$; and 3) independence of $Q_{\psi|\theta}(\cdot|\theta) $ across $\theta \in \Theta$ and from $Q_\theta$.
When utilised as a prior for the mixing measure 
$Q \sim \EDP(\alpha_\theta,  \alpha_{\psi}(\theta),Q_0)$,
this induces a random nested partition of data points in $y$-clusters and $x$-subclusters within each $y$-cluster. The latent cluster allocation of each data point consists of two terms $z_n=(z_{y,n},z_{x,n})$, where $z_{y,n}=j$ if the $n\text{th}$ data point belongs to $j\tth$ $y$-cluster with parameter $\theta_j$ and $z_{x,n}=l$ if the $n\text{th}$ data point belongs to $l\tth$ $x$-cluster with parameter $\psi_{l|j}$ within the $j\tth$ $y$-cluster. After marginalising $Q$, the model can be expressed as 
\begin{align*} 
(y_n,x_n)|z_{n}=(j,l),\theta_j, \psi_{l|j} &\indsim p(y_n|x_n,\theta_j) p(x_{n}|\psi_{l|j}),
\end{align*}
where $\theta_j \iidsim Q_{0\, \theta}$ and $\psi_{l|j}| \theta_j \iidsim Q_{0\, \psi|\theta}(\cdot|\theta_j)$. 
The law of allocation variables 
is defined by: 
 \begin{align*}
 &z_{N+1}|z_{1:N} \sim \frac{\alpha_\theta}{\alpha_\theta+N} \delta_{(k+1,1)} +\sum_{j=1}^{k} \frac{N_{j}}{\alpha_\theta+N} \left(\frac{\alpha_{\psi,j}}{\alpha_{\psi,j}+N_j} \delta_{(j,k_j+1)}+\sum_{l=1}^{k_j} \frac{N_{l|j}}{\alpha_{\psi,j}+N_j} \delta_{(j,l)}\right), 
 \end{align*}
where $k$ denotes the number of $y$-clusters of sizes $N_{j}$ and $k_j$ denotes the number $x$-clusters within the $j^\tth$ $y$-cluster of sizes $N_{l|j}$. Further, hyperpriors on the concentration parameters assume $ \alpha_\theta \sim \Gam(u_\theta,v_\theta)$ and  $  \alpha_{\psi,j} =\alpha_\psi(\theta_j)$ are independent with $\alpha_{\psi,j} \sim \Gam(u_\psi,v_\psi)$.


	A graphical comparison of the MoE with the DP and EDP priors is provided in Figure \ref{fig:gm_moe}. The DP mixture of GGP experts allocates data points to similar groups to obtain a good approximation to the joint density, with similarity measured by the local expert and input model. 
	The local factorised exponential family for the inputs is crucial for scaling to multi-dimensions and  inclusion of multiple input types. 
	However, this results in a rigid similarity measure between inputs, 
and as $D$ increases $x$ tends to dominate the partitioning structure, typically requiring many small clusters to capture increasing departures from the local input model. 
	This occurs despite the flexible nature of GPs, often requiring only a few GP experts to approximate the conditional of $y$ given $x$, 
and results in degradation of regression and conditional density estimates, wide credible intervals, and uninterpretable clustering due to the small sample sizes for each expert.
By replacing the DP with the EDP, the nested partitioning scheme allows the data to determine if the conditional of $y$ given $x$ can be recovered with fewer experts. 
The $y$-clustering is determined by similarity measured through the local expert and a more flexible local input model, which can itself be a mixture. Moreover, a simple analytically computable allocation rule is maintained, allowing the construction of efficient inference algorithms. 

%
	
	\section{Posterior inference}\label{sec:enrichedGibbs}
	
	\begin{algorithm}[!t]
   \caption{Non-conjugate collapsed Gibbs sampler}
   \label{alg:gibbs}
\begin{algorithmic}
   \STATE {\bfseries Input:} data $(y_n, x_n)_{n=1}^N$
      \STATE Initialize: $(z_{1:N}^{(0)},\sigma^{2\, (0)}_{1:k}, \beta_{0,1:k}^{(0)}, \lambda_{1:k}^{(0)},\alpha_\theta^{(0)}, \alpha_{\psi,1:k}^{(0)}, \tilde{y}_{1:N}^{(0)})$ \COMMENT{ by sampling from the prior.}
   \FOR{$m=1$ {\bfseries to} $M$}
    \FOR{$n=1$ {\bfseries to} $N$}
   \STATE Local updates to: $z_{n}^{(m)}|z_1^{(m)},\ldots,z_{n-1}^{(m)},z_{n+1}^{(m-1)},\ldots,z_N^{(m-1)}$  \COMMENT{ extending and combining Algorithm 3 and Algorithm 8  of  \citet{neal2000markov} for the nested partition.}
      \ENDFOR
   \STATE Global split-merge $y$-cluster updates to: $z_{1:N}^{(m)}$ \COMMENT{ Metropolis-Hastings step to move an $x$-cluster to be nested within a new or different $y$-cluster.}
      \STATE Global split-merge $x$-cluster updates to: $z_{x,1:N}^{(m)}$ \COMMENT{ extending  \citep{JN04,wang2015} to split or merge $x$-clusters nested within a common $y$-cluster.}
   \STATE Sample $y$-cluster parameters $(\sigma^{2\, (m)}_{1:k}, \beta_{0,1:k}^{(m)}, \lambda_{1:k}^{(m)})$ \COMMENT{ using Hamiltonian Monte Carlo \citep{duane1987hybrid}.}
   \STATE Sample concentration parameters $(\alpha_\theta^{(m)}, \alpha_{\psi,1:k}^{(m)})$ \COMMENT{ with auxiliary variable techniques \citep{EW95}.}
    \STATE Sample latent outputs $  \tilde{y}_{1:N}^{(m)}$ \COMMENT{ if present, through Gibbs sampling and CDF inversion \citep{kotecha1999gibbs}.}
   \ENDFOR
\end{algorithmic}
\end{algorithm}
	
	
	For inference, we resort to Markov chain Monte Carlo (MCMC) and derive a collapsed Gibbs algorithm to sample 
	the latent allocation variables $z_{1:N}$ and unique $y$-cluster parameters $(\theta_j)$, with the $x$-cluster parameters  $(\psi_{l|j})$ marginalised. 
Additionally, we focus on the case when the functions $m_j(\cdot)$ can be marginalised; this includes the Gaussian likelihood, but also the ordered probit, among others, through data augmentation. In the latter, the data is augmented with latent Gaussian outputs $\tilde{y}_{1:N}$, which have a deterministic relationship with the observed outputs. 
Algorithm \ref{alg:gibbs} gives an overview of the MCMC scheme (with a full description in the SM). To improve mixing, two novel split-merge updates are developed  for global changes to the allocation variables.

\begin{figure}[!t]
\begin{center}
\subfigure[DP PSM, $D=1$]{\includegraphics[width=0.23\textwidth]{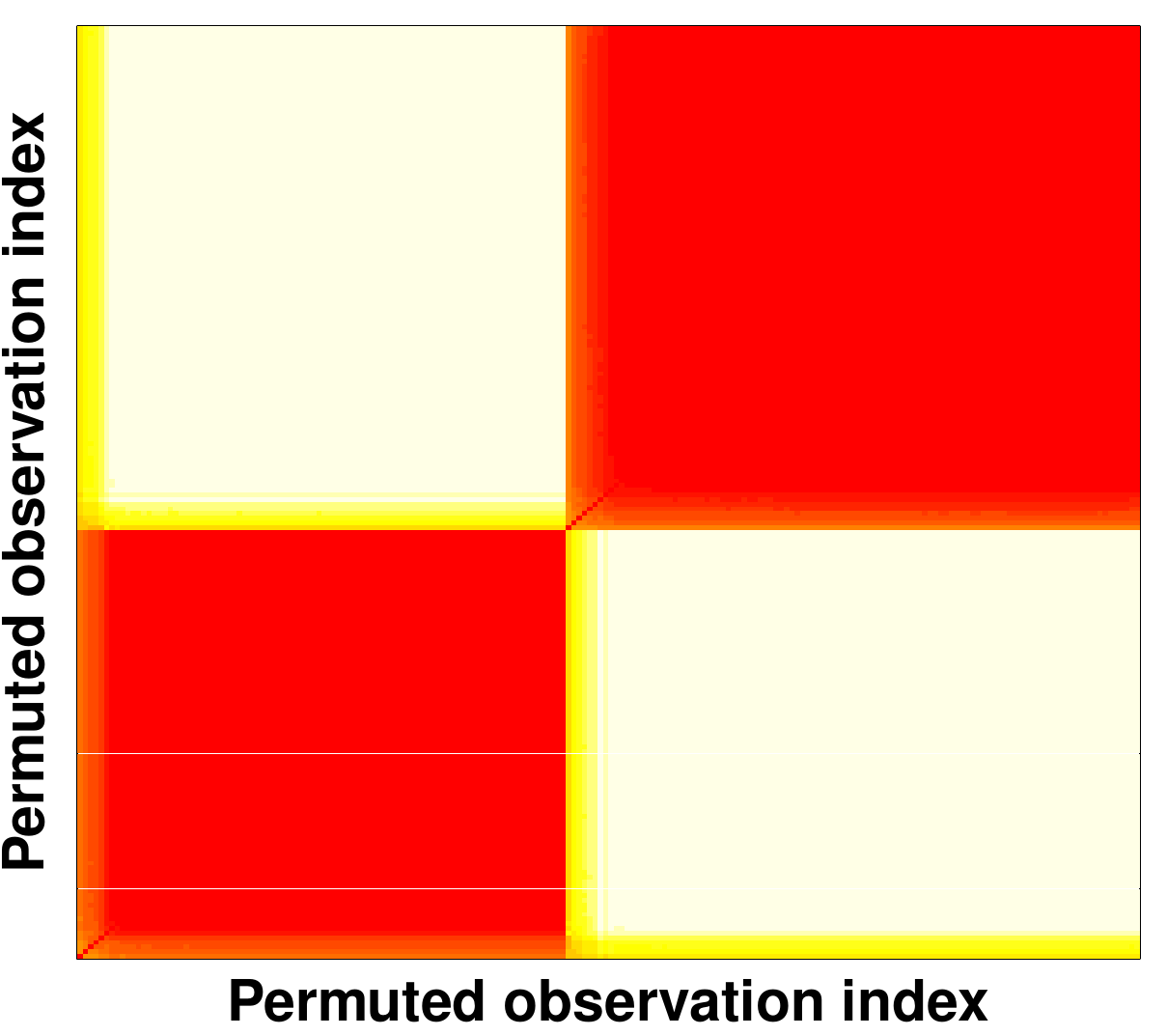}\label{fig:JmE1_psm}}
\subfigure[DP PSM, $D=5$]{\includegraphics[width=0.23\textwidth]{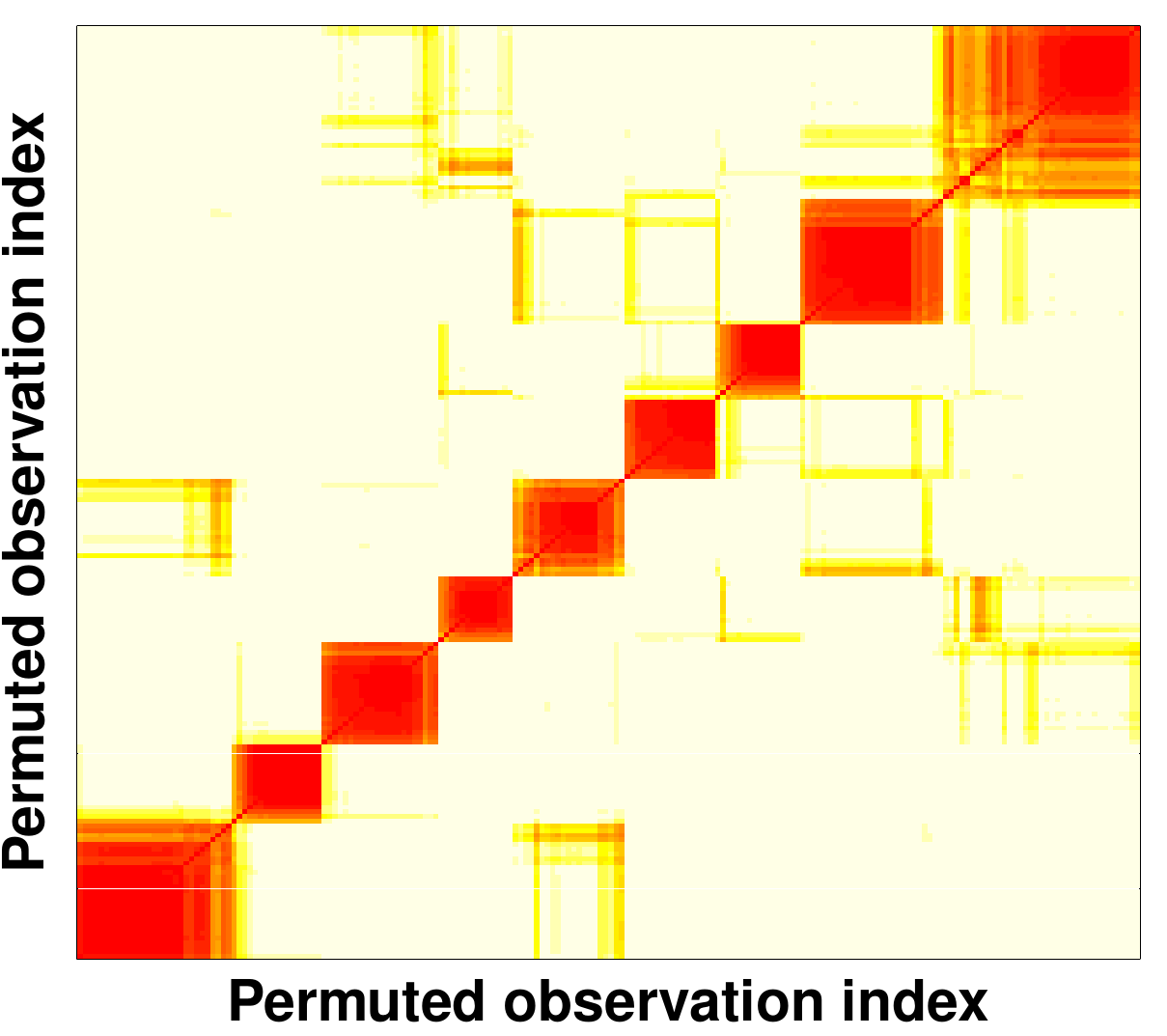}\label{fig:JmE5_psm}}
\subfigure[DP clustering, $D=1$]{\includegraphics[width=0.26\textwidth]{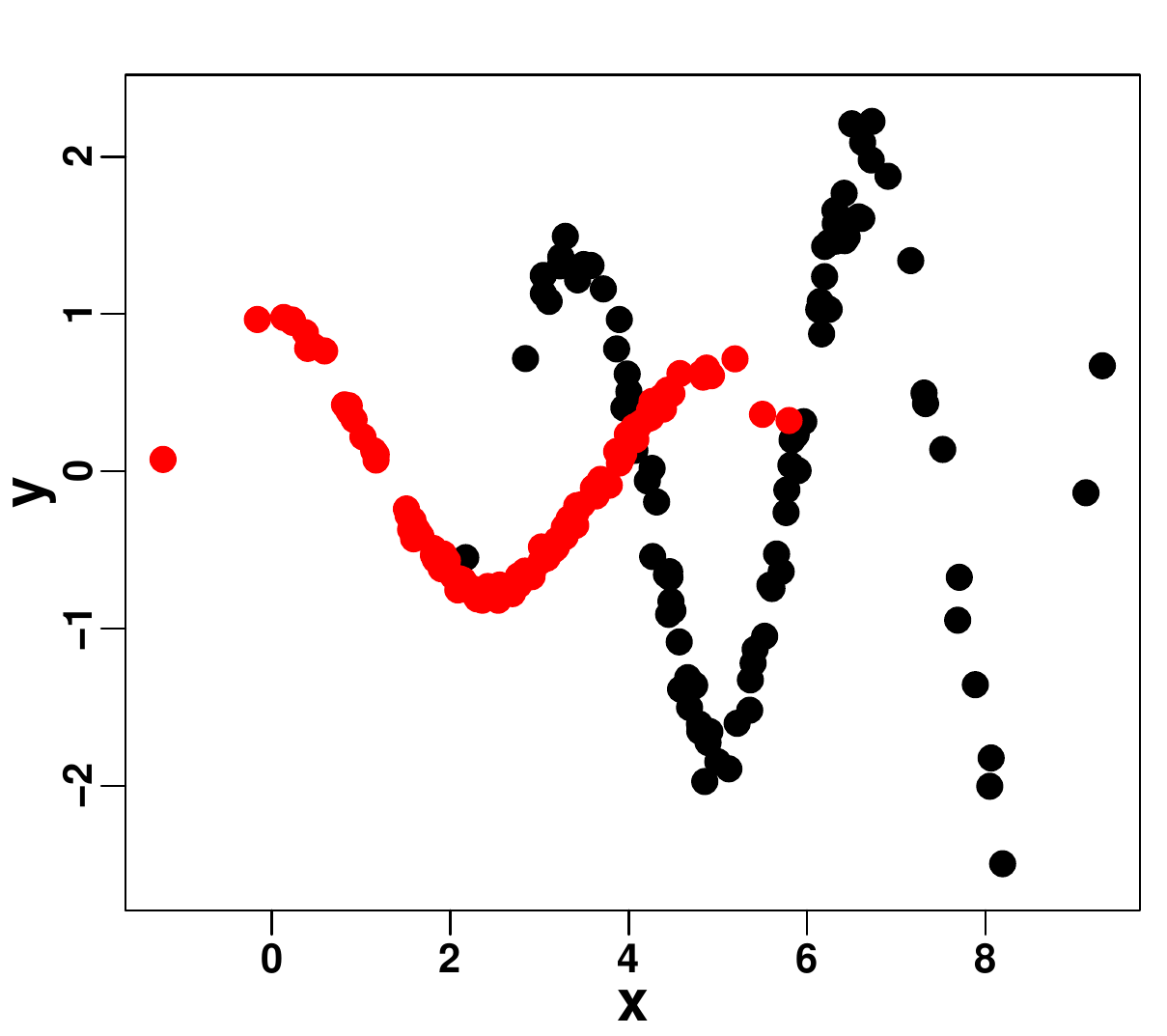}\label{fig:JmE1_VI}}
\subfigure[DP clustering, $D=5$]{\includegraphics[width=0.26\textwidth]{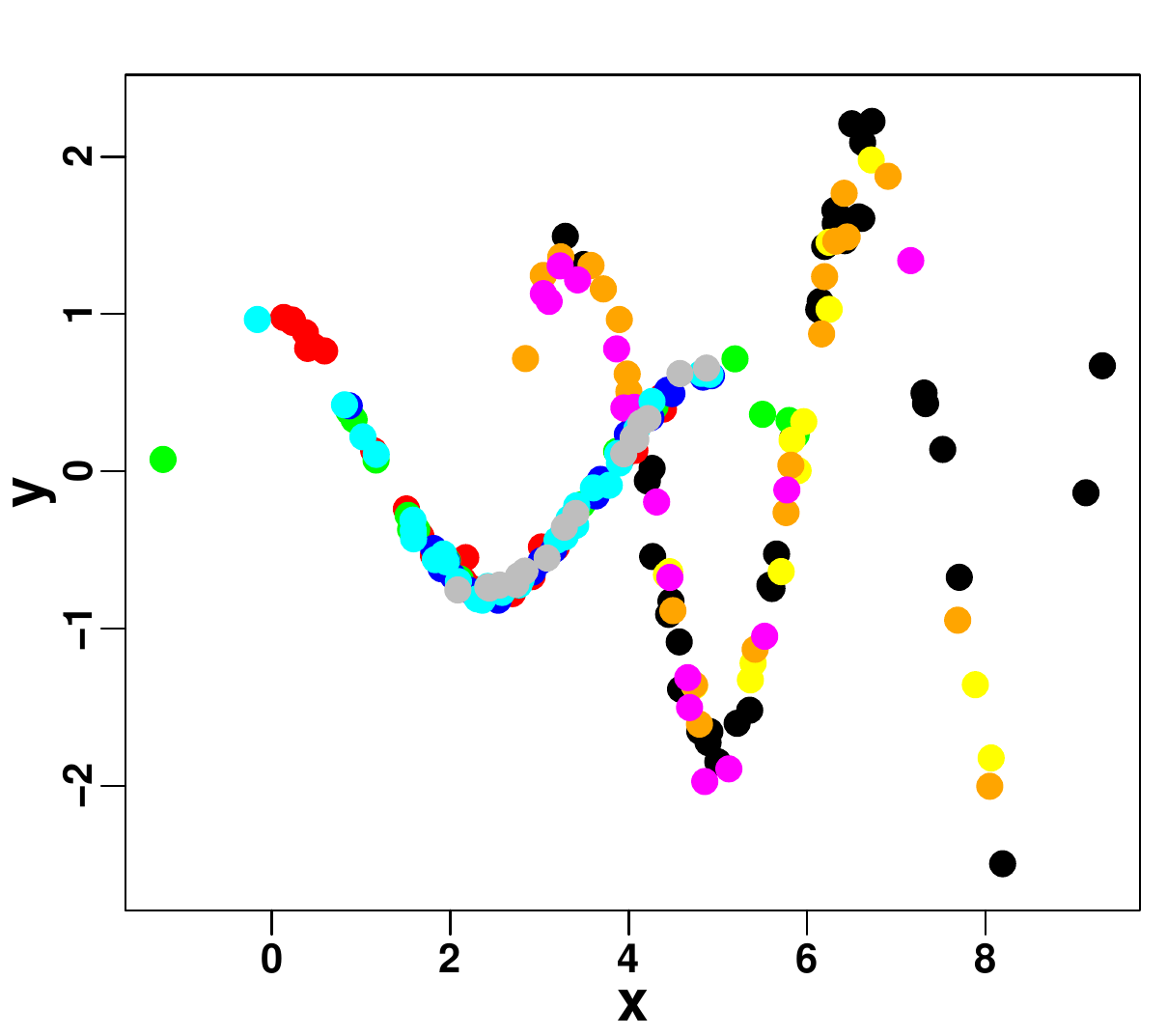}\label{fig:JmE5_VI}}\\
\subfigure[EDP PSM, $D=1$]{\includegraphics[width=0.23\textwidth]{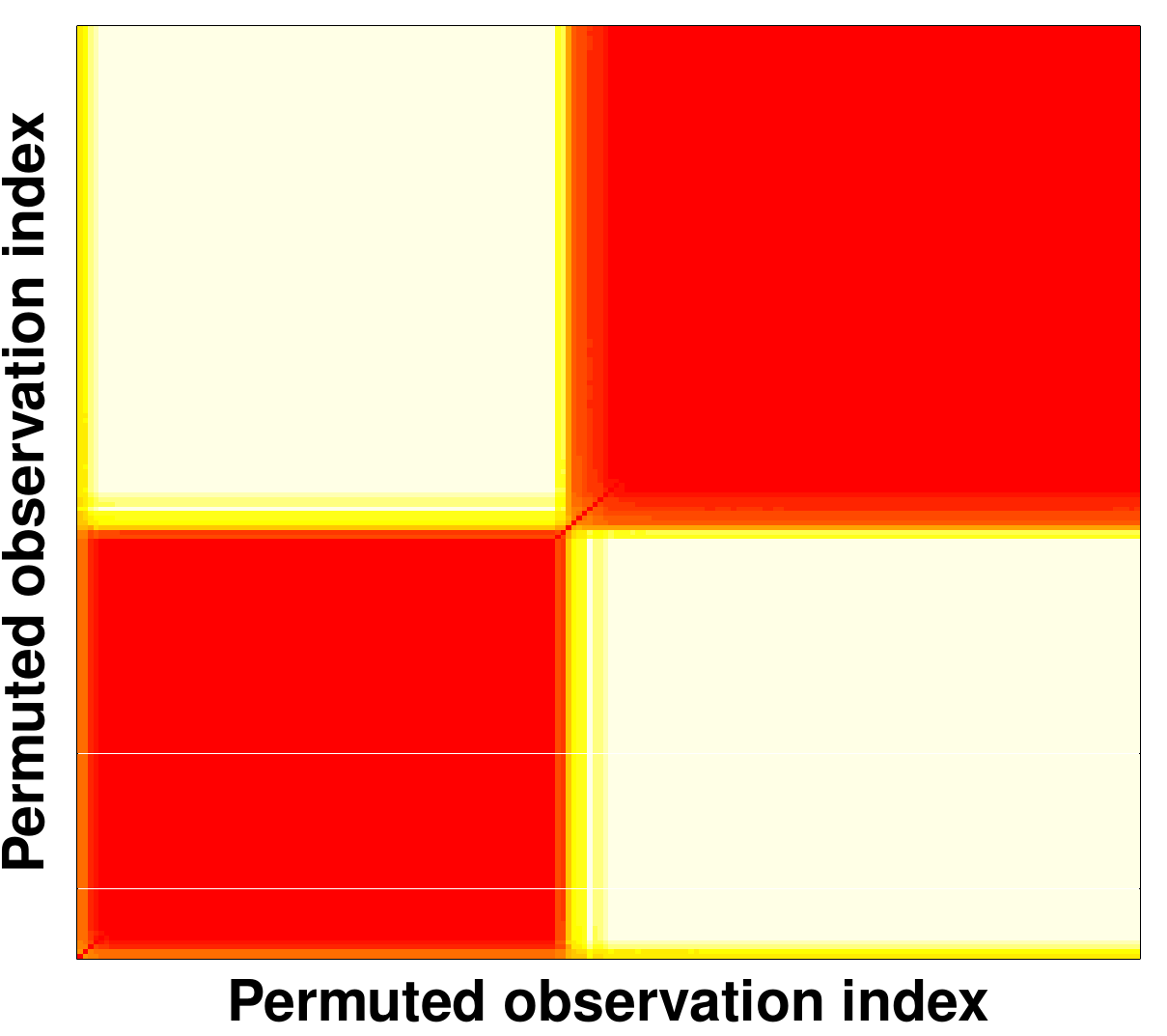}\label{fig:EmE1_psm}}
\subfigure[EDP PSM, $D=5$]{\includegraphics[width=0.23\textwidth]{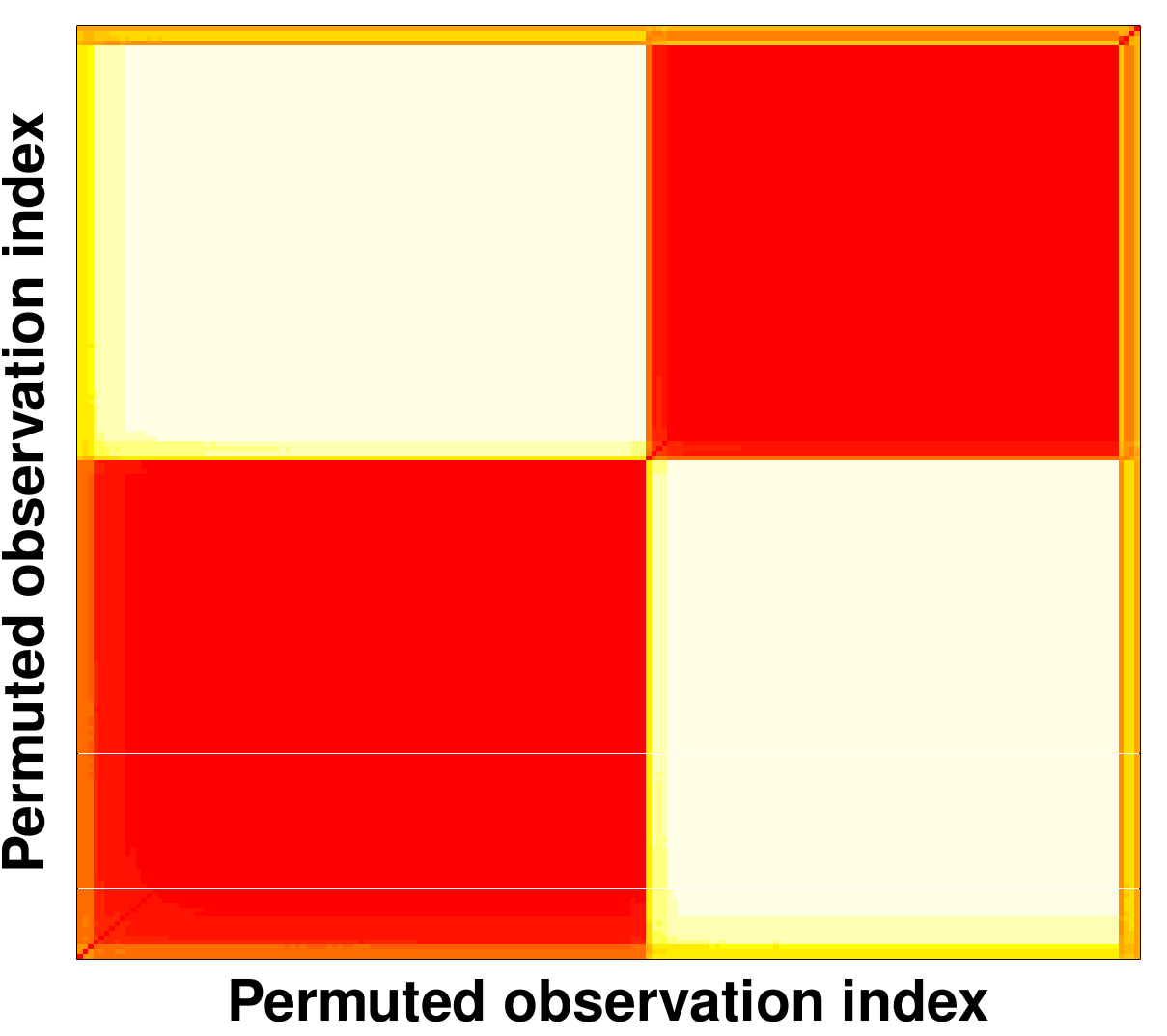}\label{fig:EmE5_psm}}
\subfigure[EDP clustering, $D=1$]{\includegraphics[width=0.26\textwidth]{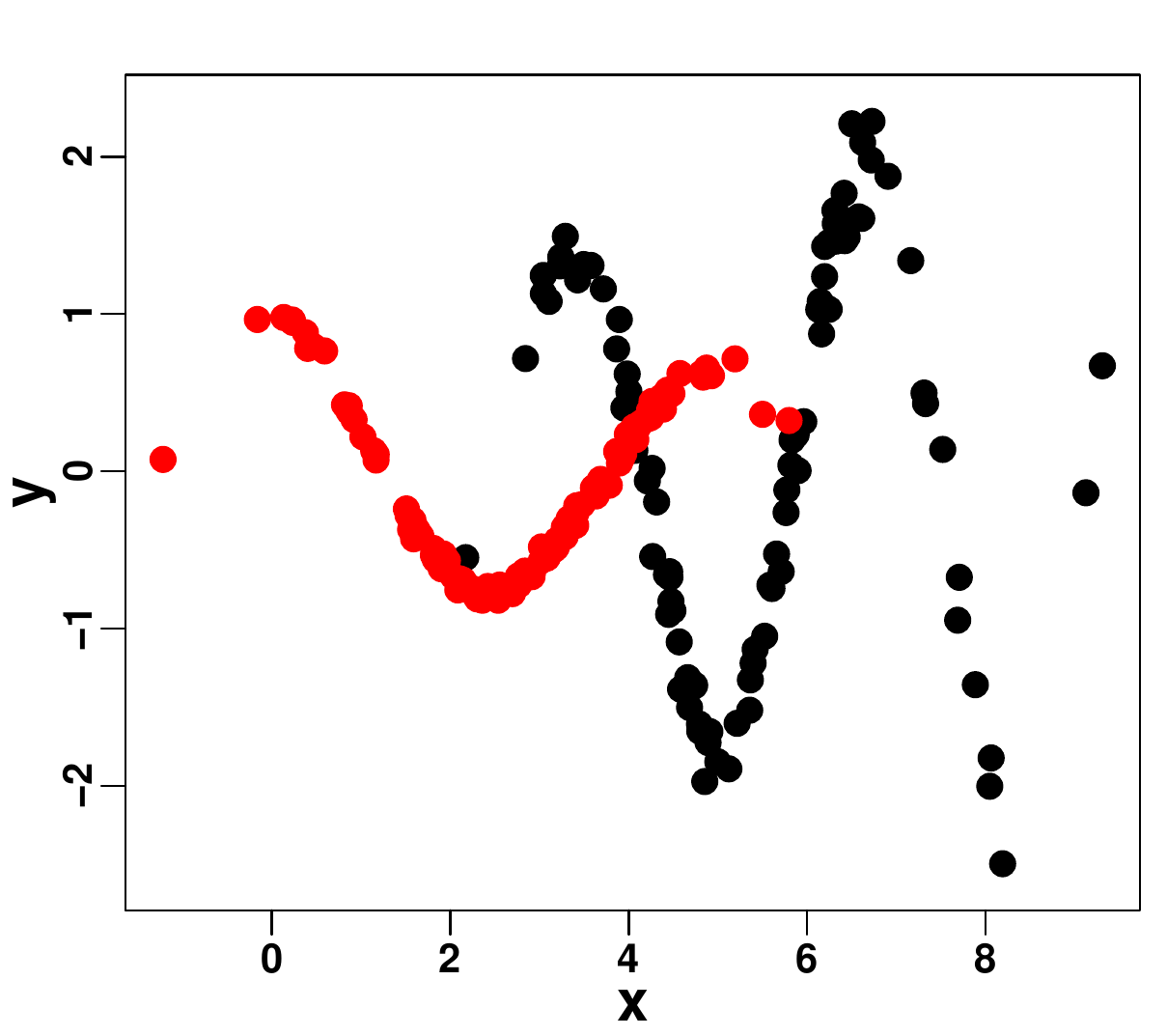}\label{fig:EmE1_VI}}
\subfigure[EDP clustering, $D=5$]{\includegraphics[width=0.26\textwidth]{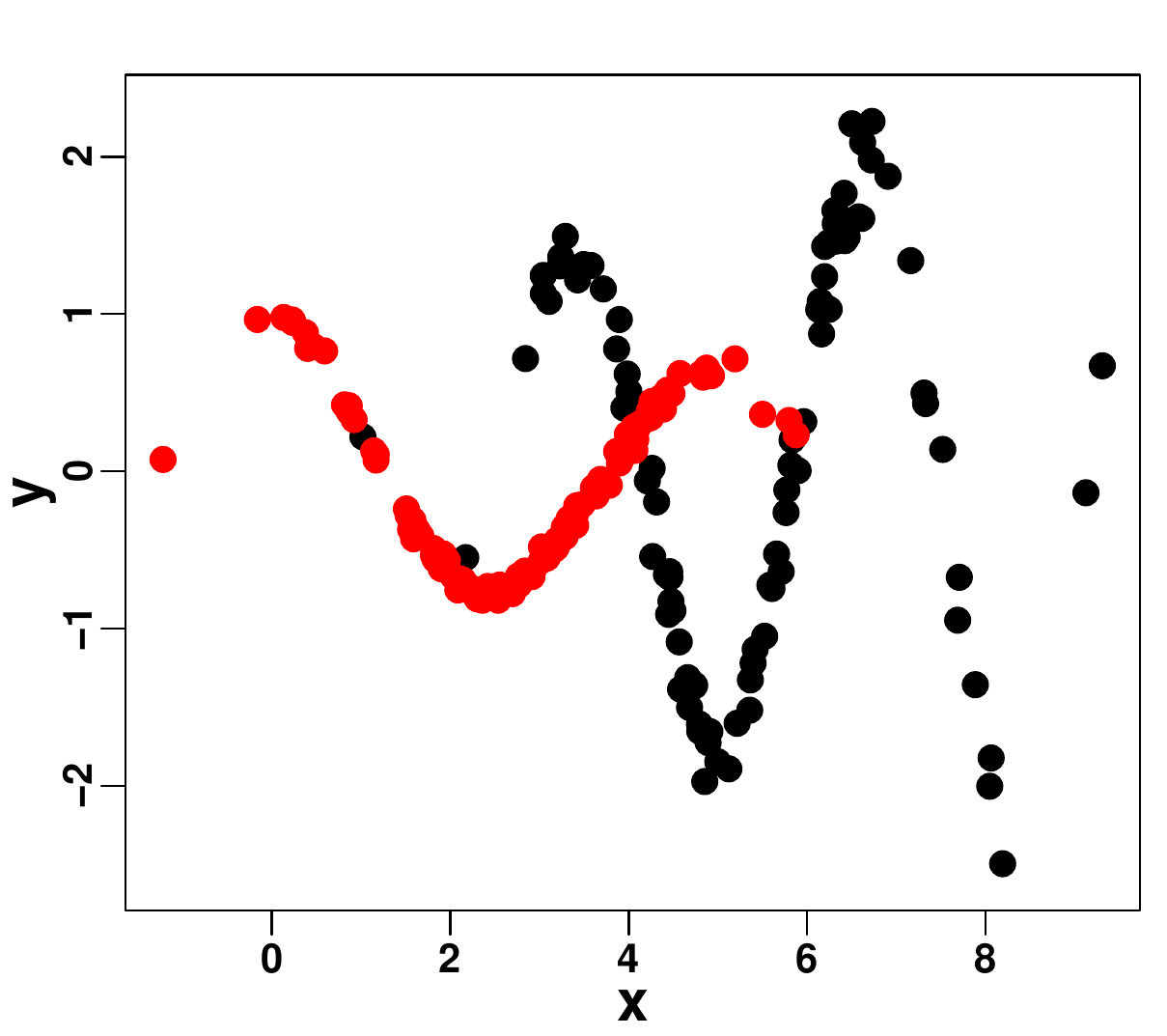}\label{fig:EmE5_VI}}
\caption{Simulated example. \textit{Left}: Heat map of the posterior similarity matrix (PSM), with observations indices permuted based on hierarchical clustering to improve visualisation.  \textit{Right:} VI clustering estimate with data points  $(y_n,x_{n,1})$ coloured by cluster membership. Rows correspond to DP and EDP MoE, respectively. For the EDP, plots correspond to the $y$-level clustering.}  
\label{fig:santner_psm}
\end{center}
\vspace{-5mm}
\end{figure}
	
	
\paragraph{Predictions.} 	
From the $M$ MCMC samples, 
we compute predictions for the new output $y_*$ given $x_*$. In the Gaussian case, the posterior expectation of $y_*$ 
is  approximated by 
$$\mathbb{E}[y_*|x_*,y_{1:N},x_{1:N}] =C^{-1}\left( \sum_{m=1}^M  p^{(m)}_{k^{(m)}+1}(x_*) \mu_\beta+\sum_{j=1}^{k^{(m)}}  p^{(m)}_{j}(x_*) \widehat{m}_j^{(m)}(x_*) \right),$$
%
where $C$ 
is the normalising constant; and for a new cluster, the predictive mean is simply the prior expectation of $\beta_{0,k^{(m)}+1}$, denoted by $\mu_\beta$, while 
for an existing cluster, 
the GP predictive mean in cluster $j$ is denoted by $\widehat{m}_j^{(m)}(x_*)$. 
Thus, for each sample, the expectation is a weighted average of the GP predictions from each cluster and a new cluster, with input-dependent weights that flexibly measure the similarity between the new input and the inputs of each cluster through a mixture.  Specifically, the weights of a new cluster and existing cluster $j$, for $j=1,\ldots,k^{(m)}$, are, respectively,
\begin{align}
p^{(m)}_{k^{(m)}+1}(x_*) &= \frac{\alpha^{(m)}_\theta}{\alpha^{(m)}_\theta+N} h(x_*),\nonumber\\
 p^{(m)}_{j}(x_*) &= \frac{ N_j^{(m)}}{\alpha^{(m)}_\theta+N} \left[\frac{\alpha_{\psi,j}^{(m)}}{\alpha_{\psi,j}^{(m)}+N_j^{(m)}} h(x_*) + \sum_{l=1}^{k_j^{(m)}} \frac{N_{l|j}^{(m)}}{\alpha_{\psi,j}^{(m)}+N_j^{(m)}} h(x_*|\bm{X}_{l|j}^{(m)}) \right].\label{eq:enrichedweights}
\end{align}
Here,  $h(x_*)$ is the marginal density of $x^*$ and  $h(x_*|\mathbf{X}_{l|j})$ is the predictive marginal density of  $x_*$ given $\mathbf{X}_{l|j}$, which contains the $x_n$ such that $z_n=(j,l)$. In contrast, for the DP, the weight of an existing cluster in \eqref{eq:enrichedweights} is more rigidly defined based on a single predictive marginal density, arising from the factorised exponential family. 
The predictive density or appropriate quantities for other output types are similarly computed. Another advantage of the joint approach includes the calculation of predictions  based only on a subset of inputs. Full derivations are provided in the SM.

\paragraph{Clustering.}
The enriched MoE induces a nested clustering of the data points into $y$-clusters and nested $x$-clusters. This latent clustering may be of interest to identify similar groups of data points 
and to improve understanding of the model. 
The MCMC samples from the posterior over this nested clustering; to summarise the samples, we obtain the point estimate from minimising the posterior expected variation of information \citep[VI,][]{WG17}, first estimating the $y$-level partition and then the nested $x$-level partition. 
To visualise uncertainty in the clustering structure, 
 we also compute the posterior similarity matrix with elements 
$ p(z_{y,n}=z_{y,n'}|y_{1:N},x_{1:N})$ 
representing the posterior probability that two points are clustered together and approximated by the fraction of times this occurred in the chain. 

\section{Examples}\label{sec:examples}
	
	We demonstrate the advantages of the enriched MoE, namely, improved predictive accuracy, smaller credible intervals, and more interpretable clustering, in two examples. The first demonstrates increasing improvement over the DP as $D$ increases, and the second shows the range of applicability of our model for ordinal outputs with multiple input types. Code to implement the model and reproduce the results is publicly available at \textit{GitHub link}, 
	alongside further plots and videos. Prior parameter specification, further algorithm details, and epoch times are detailed in the SM.
	
	\subsection{Simulated mixture of damped cosine functions}\label{sec:simulated}

\begin{figure}[!t]
\begin{center}	
\subfigure[Number of clusters]{\includegraphics[width=0.32\textwidth]{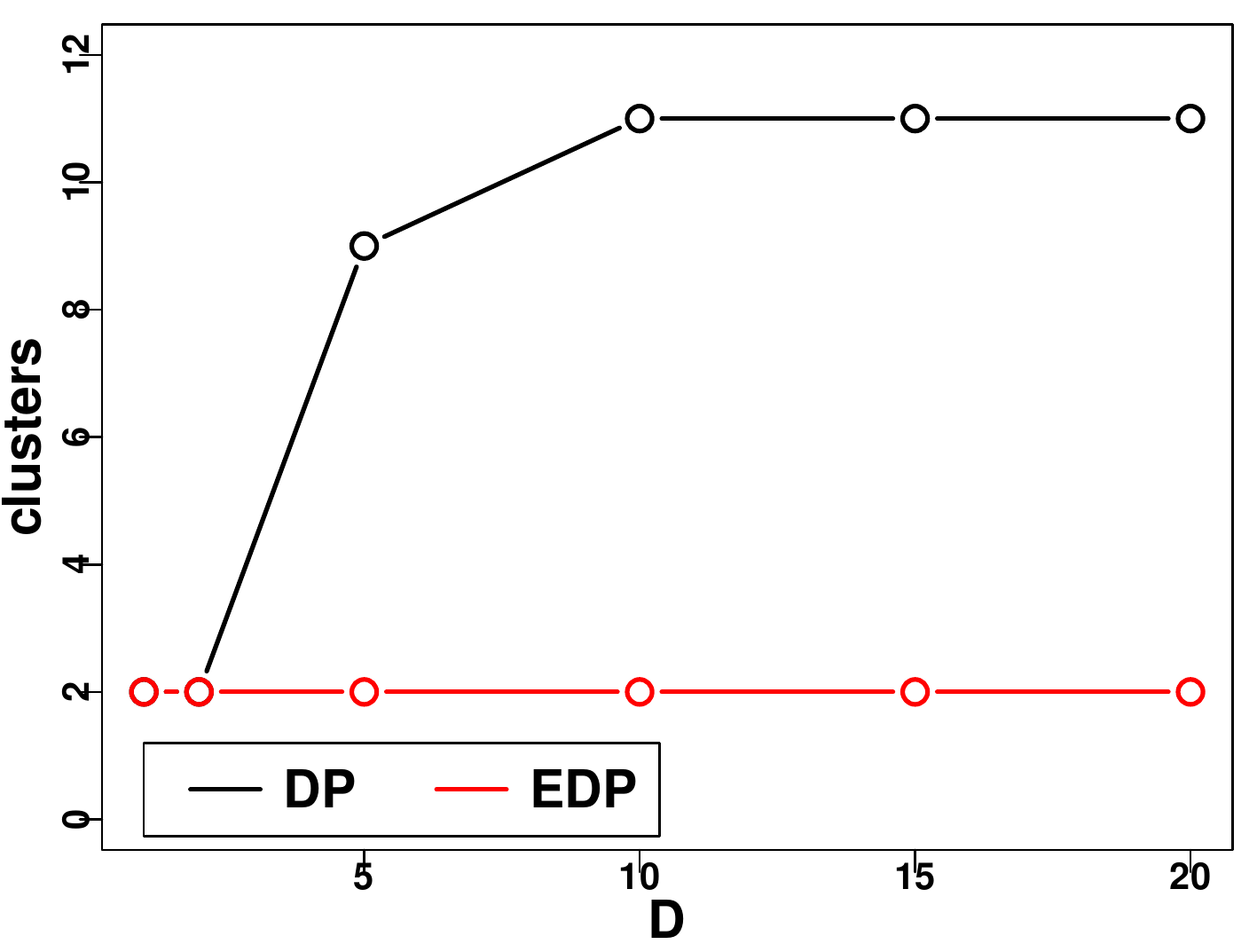}\label{fig:clusters}}
\subfigure[$L_1$ error]{\includegraphics[width=0.32\textwidth]{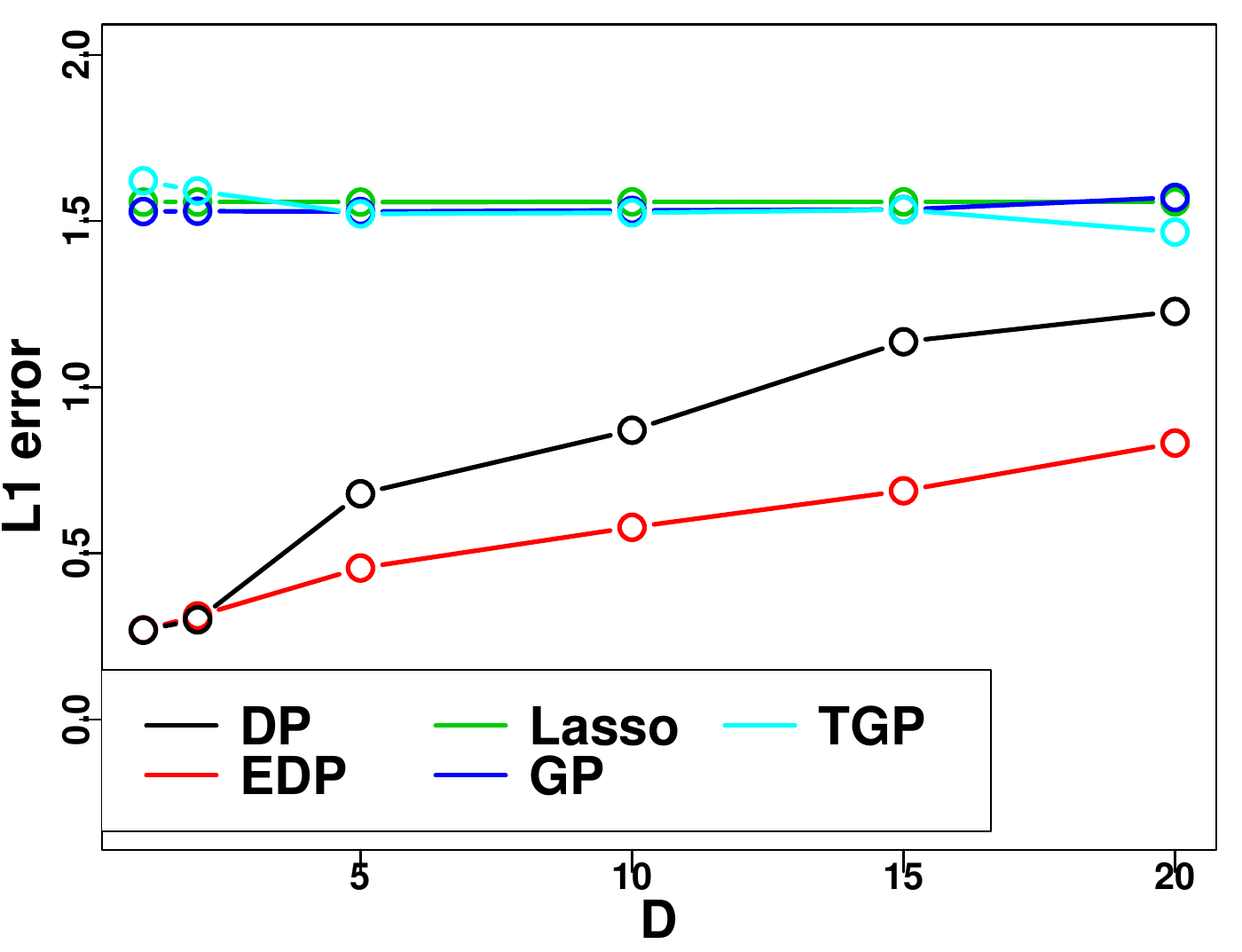}\label{fig:l1error}}
\subfigure[Average CI length]{\includegraphics[width=0.32\textwidth]{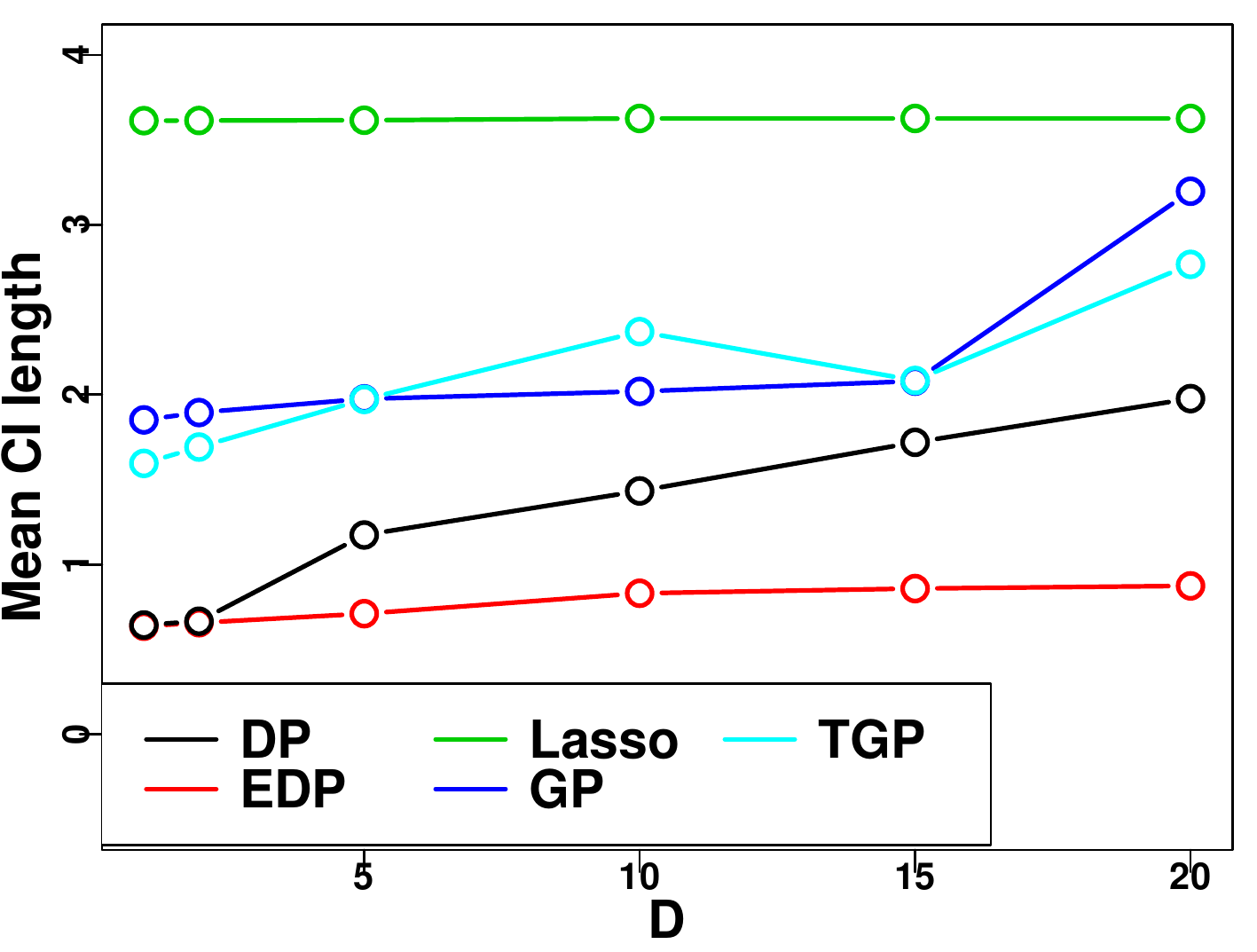}\label{fig:cilength}}
\caption{Simulated example. Comparison of the EDP MoE with the DP MoE, Lasso, GP, and TGP in terms of number of clusters in the VI estimate,  approximate $L_1$ distance of the estimated and true conditional densities, and average length of the 95\% credible intervals (CI). }
\label{fig:simcomparison}
	\end{center}
	\vspace{-5mm}
\end{figure}

In the first example, $N=200$ points are generated with only the first input as a predictor. The true output model  is a highly non-linear regression obtained as a mixture of two  damped cosines \citep{santner2003design}.  The inputs are independently sampled from a multivariate normal, 
with the additional inputs positively correlated among each other but independent of the first input. 
We compare the DP and EDP, with automatic relevance determination (ARD) squared exponential kernels for the GP experts. 

A heat map of the posterior similarity matrix from the DP MoE in Figures \ref{fig:JmE1_psm}-\ref{fig:JmE5_psm} highlights data points with a high probability of clustering together in red; the need for a  greater number of clusters as $D$ increases is clearly evident. Indeed,  the VI clustering estimates in Figures \ref{fig:JmE1_VI}-\ref{fig:JmE5_VI} contain two clusters for $D=1$ and $9$ clusters for $D=5$. Conversely, the corresponding plots for the $y$-level clustering of the EDP in Figures \ref{fig:EmE1_psm}-\ref{fig:EmE5_VI}  highlight two 
$y$-clusters. Figure \ref{fig:clusters} emphasizes this
improvement of the EDP in recovering the true number of $y$-clusters for increasing $D$.
This leads to more accurate predictions and tighter credible intervals. 
We quantify the predictive accuracy with the approximate $L_1$ distance between the estimated predictive response density and true data generating density, averaged across test samples. 
These errors are depicted in Figure \ref{fig:l1error}, alongside the average length of 95\% credible intervals  in Figures \ref{fig:cilength}, comparing the EDP with the DP, Lasso, GP, and TGP.  
While the $L_1$ errors generally increase with $D$, the EDP is the most robust. As expected, the Lasso, an effective tool for sparse linear regression, performs poorly in this highly non-linear example, but interestingly, the GP and TGP perform just as bad due to the inability to cope with bimodality. Moreover, the EDP produces tighter credible intervals across $D$, compared with the other methods, while maintaining similar coverage. 


\subsection{Alzheimer's challenge}\label{sec:ADNI}

	 \begin{wrapfigure}{R}{0.45\textwidth}
    \begin{minipage}{0.45\textwidth}
    \vspace{-4mm}
\begin{table}[H]
	\caption{Alzheimer's challenge. Comparison of EDP with  DP and competition winners by 1) number of clusters;  2) mean absolute \textit{test} error;  
  3) empirical coverage and 4) average length  of 95\% credible intervals. }
		\label{tab:adni}
		\begin{center}
	\begin{tabular}{ccccc}
	\toprule
		& $\widehat{k}$ &  $\text{MAE}_{\text{test}}$  &$\text{EC}_{95}$&$\bar{\text{CI}}_{95}$ \\ 
		\midrule
				EDP& 3 & \textbf{2.112}
		& 0.948 & 8.96
		 \\ 
		DP & 7   & 2.149 &  0.950 &  9.10 \\ 
		GL & 3 & 2.153  & - &  -  \\
		GL2 & 3  & 2.208 & 0.945 & 11.06 \\ 
		ADDT   & - & 2.158  &0.867  & 8.29 \\ 
		\bottomrule
	\end{tabular}
\end{center}
\end{table}
\end{minipage}
    \vspace{-3mm}
\end{wrapfigure}

Motivated by the Alzheimer's Disease Big Data DREAM Challenge (\adnidream), this study aims to predict cognitive scores 24 months after initial assessment. This can potentially assist in early diagnosis of Alzheimer's disease (AD) and provide personalised predictions with uncertainty for patients and their families. 
Training data is extracted from the Alzheimer's Disease Neuro-Initiative (ADNI) database (\adniinfo).
We emphasise that the competition test data can no longer be accessed, and the \textit{test} results presented here are based on a random split of the data into training and test sets of sizes $N=384$ and $N^*=383$. The ordinal response $y_n$ is the mini-mental state exam (MMSE) score at a 24 month follow-up visit; MMSE is an extensively used clinical measure of cognitive decline, defined on a 30 point scale with lower scores reflecting increased impairment. The $D=6$ inputs include baseline age (in fraction of years); gender; baseline MMSE; education; APOE genotype, with values 0, 1, or 2, reflecting the number of copies of the type 4 allele; and diagnosis at baseline of cognitively normal (CN), early mild cognitive impairment (EMCI), late mild cognitive impairment (LMCI), and AD, respectively. 
The winners of this subchallenge were GuanLab (GL) and ADDT. GL \citep{GuanLab} separated training data into three groups of CN, MCI or AD and trained support vector machines (SVM) within each group; SVMs provide non-probabilistic predictions, and for comparison, we also train linear regression models within group to obtain prediction intervals in GL2.  ADDT \citep{ADDT} used robust regression based on M-estimation, optimally combined diagnosis and APOE4, and included interactions. 

\begin{figure}[!t]
\begin{center}
	\subfigure[ CN and APOE4=0]{\includegraphics[width=0.35\textwidth]{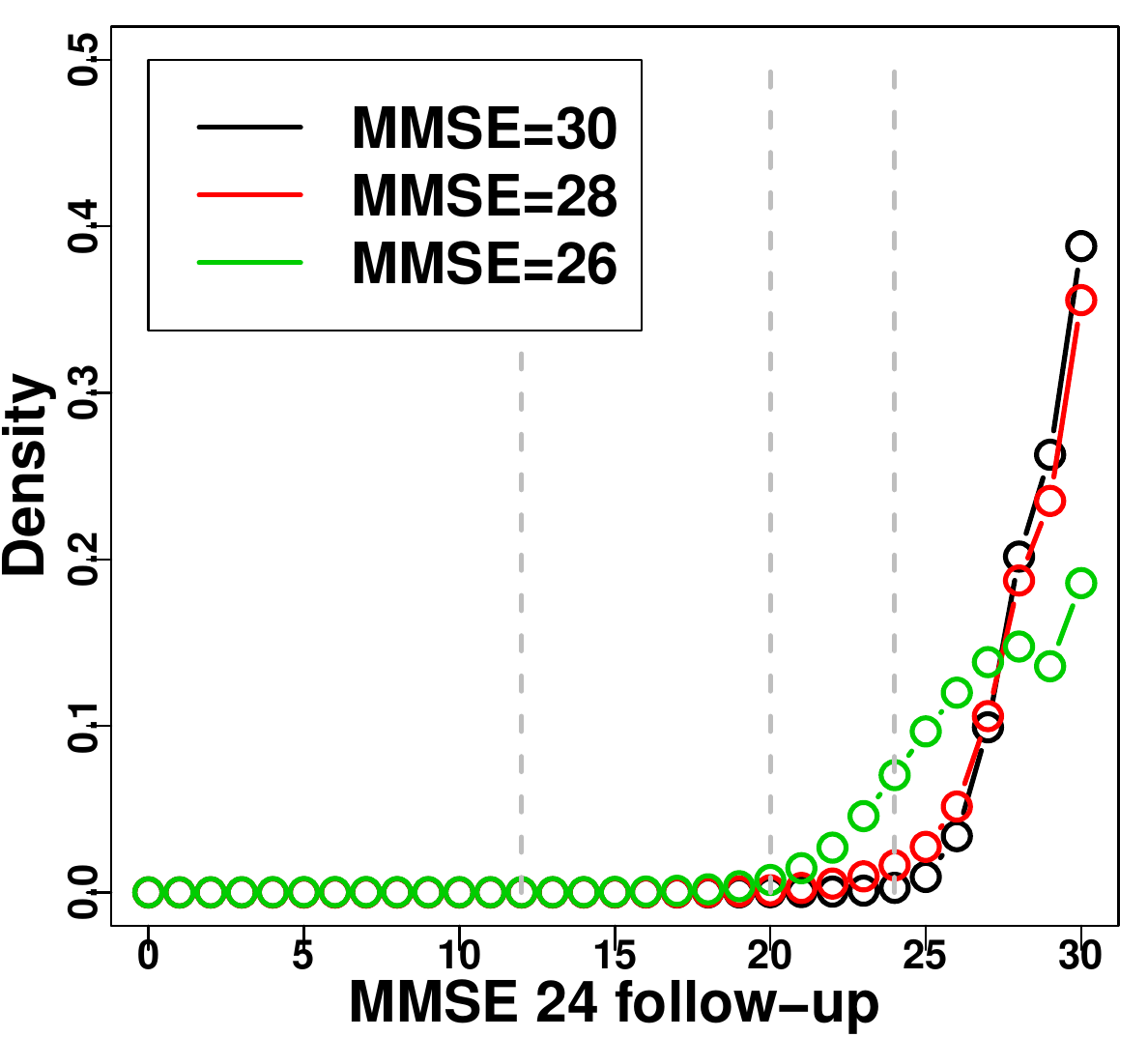}\label{fig:CN0_conditional}}
	\subfigure[ CN and APOE4=2]{\includegraphics[width=0.35\textwidth]{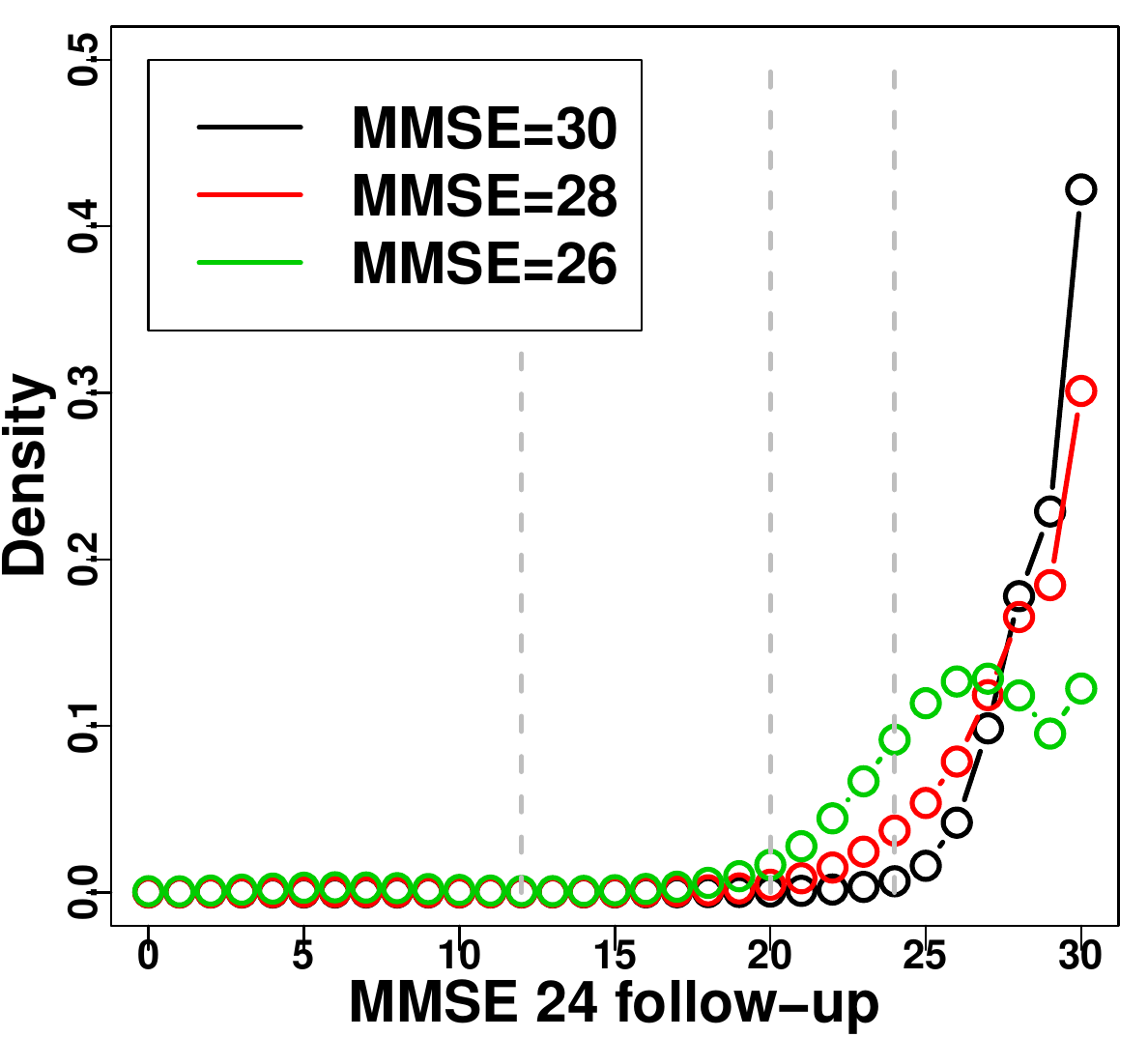}\label{fig:CN2_conditional}}\\
		\subfigure[ AD and APOE4=0]{\includegraphics[width=0.35\textwidth]{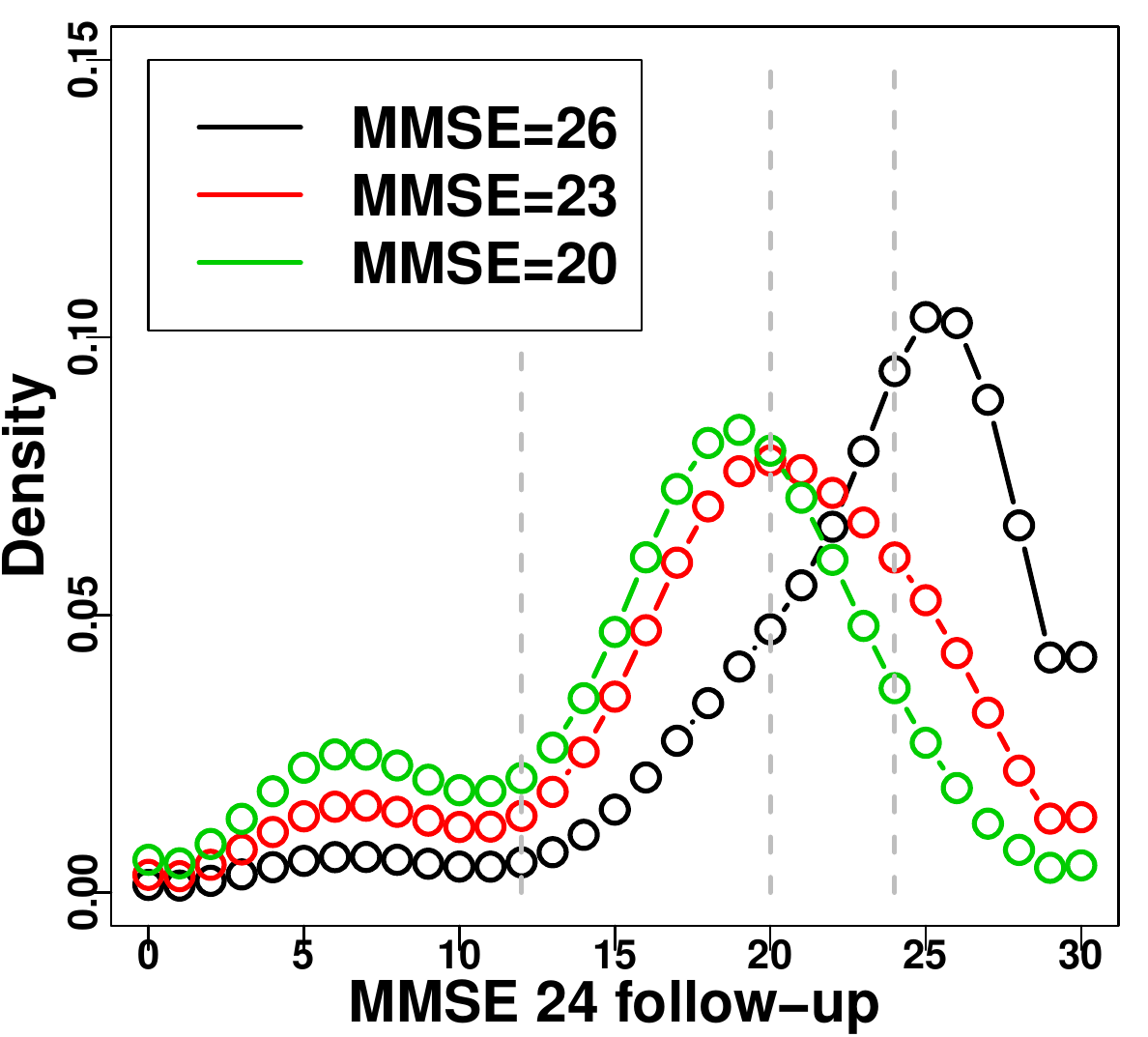}\label{fig:AD0_conditional}}
	\subfigure[ AD and APOE4=2]{\includegraphics[width=0.35\textwidth]{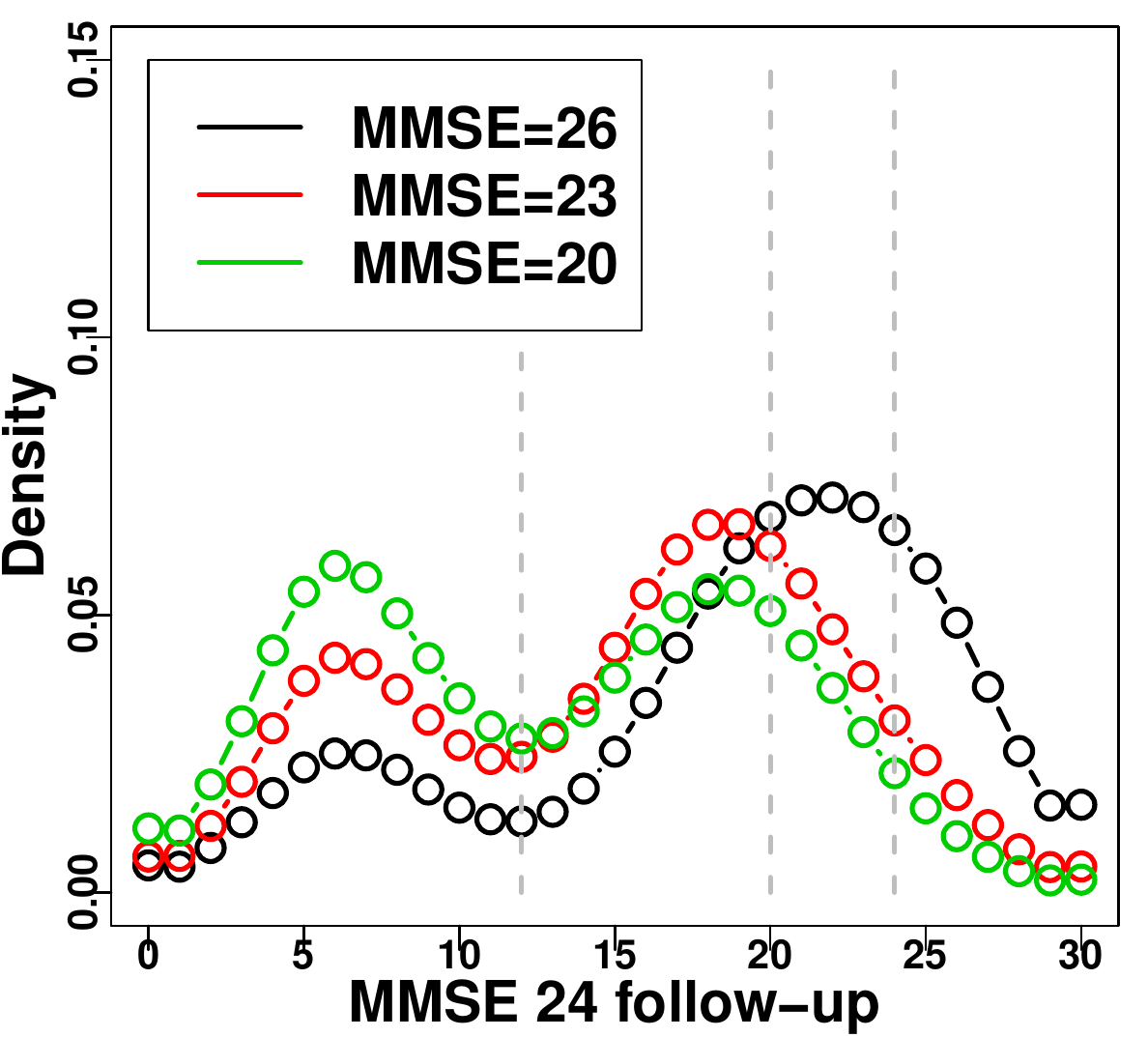}\label{fig:AD2_conditional}}
	\caption{Alzheimer's challenge. Marginal predictive density of MMSE 24-month follow-up scores for different combinations of MMSE baseline, APOE4, and baseline diagnosis from the EDP mixture of experts. Dashed lines indicate established cutoffs for MMSE: $\geq 25$ suggests no dementia; $20-24$ suggests mild dementia; $13-19$ suggests moderate dementia; $\leq 12$ suggests severe dementia.  }
	\label{fig:adni_predictiveEME}
	\end{center}
	\vspace{-5mm}
\end{figure}

The EDP MoE can flexibly recover non-linear trajectories of the cognitive decline, while also clustering patients into input-dependent groups of similar trajectories. We consider the ordered probit GGP with fixed cutoffs $0=\varepsilon_0<\varepsilon_1 =1<\varepsilon_2 =2\ldots < \varepsilon_{29}=29 $. 
Table \ref{tab:adni} summarises the \textit{test} performance of the methods via the mean absolute error and 
empirical coverage and average length of the $95\%$ credible intervals. 
Compared to the DP, the EDP performs slightly better in mean absolute \textit{test} error 
and has smaller uncertainty, reflected in a reduced average credible interval length, while maintaining good coverage ($\approx0.95$). 
This improvement is due to its ability to capture the relationship between $y$ and $x$ with fewer clusters, also leading to more interpretable clustering. Indeed, the VI estimate of the $y$-clustering of the EDP has only three clusters, while the DP has seven. 
Similar to GL, the EDP identifies three clusters of mostly  CN, MCI, and AD individuals, with some adjustments for other variables, particularly, MMSE scores. 
 The SM contains a deeper discussion on the clustering.


The EDP MoE produces flexible nonparametric density estimates of MMSE follow-up scores that change smoothly with the inputs. Specifically, Figure \ref{fig:adni_predictiveEME} shows how the densities become less peaked with larger variability for decreased baseline MMSE, increased APOE4, and increased severity in diagnosis. Instead, GL and ADDT are not able to capture this behaviour, e.g. with a minimum prediction interval length of 8 for ADDT, despite the high probability of follow-up MMSE close to 30 for CN individuals with a baseline MMSE of 30 in Figure \ref{fig:adni_predictiveEME}. Also, note the apparent difference across APOE genotype for AD patients, with an increased probability of progressing to severe dementia for carriers in Figure \ref{fig:adni_predictiveEME}. Thus, the EDP provides much improved uncertainty in predictions, which is particularly important in clinical settings and in relation to established cutoffs for MMSE.

\section{Discussion}

Infinite mixtures of GP experts are flexible models, that can capture non-stationary functions and departures from the typical homoscedastic normality assumptions on the errors. In this work, we proposed a novel enriched mixture of GGP experts,  with local independence of the inputs, to increase scalability and allow inclusion of multiple input types, and a nested partitioning scheme, to improve predictive accuracy, uncertainty, and interpretability of the clustering. Moreover, through the generalised GP framework, we can account for different output types. 

A number of proposals extend mixtures of linear experts for high-dimensional inputs using regularisation or variable selection, e.g. \citep{peralta2014, barcella2017}. Here, we focus on multi-dimensional input spaces, with GP experts and ARD kernels, that allow determination of the local relevance of each input. An important future research direction will incorporate methods to scale the GP experts to higher-dimensions, for example, using simple isotropic kernels or dimension reduction techniques \citep{snelson2012variable}.
To scale to larger datasets, future research will also focus on fast approximate inference such as MAP techniques \citep{raykov2014simple}.

%

\small
\bibliographystyle{plainnat}
\bibliography{bibFile}

%

\end{document}


\maketitle

\section{Generalised Gaussian process experts}
\label{app:GGPM}

Examples of generalised GP experts include:
%
%
\paragraph{Gaussian:} for $y \in \mathbb{R}$, with identity link function,
$$p(y|x,\theta_j)= \Norm(y|m_j(x), \sigma_j^2).$$
%
\paragraph{Bernoulli:} for $y \in\lbrace0,1 \rbrace$,
$$p(y|x,\theta_j)= \Bern(y|g^{-1}(m_j(x))),$$
where the link function maps $(0,1)$ to the real line, e.g. logistic, probit. For the logistic link function,
$$\mathbb{P}(y=1|x,\theta_j)=\frac{\exp(m_j(x))}{1+\exp(m_j(x))}.$$
For the probit link function,
$$\mathbb{P}(y=1|x,\theta_j)=\Phi(m_j(x)),$$
where $\Phi$ denotes the standard normal cumulative distribution function. In this case, the model can be equivalently formulated through a latent response $\tilde{y}$ that is Gaussian distributed with mean $m_j(x)$ and unit variance. In particular, 
$ \tilde{y}| m_j(x) \sim \Norm(m_j(x), 1)$ and  $$\quad p(y| \tilde{y})=\left\lbrace \begin{array}{ll}
\1(\tilde{y} \leq 0) & \text{if } l=0 \\
\1( \tilde{y}> 0) & \text{if } l=1 
\end{array} \right..$$
The probit model is recovered by marginalising the latent $\tilde{y}$.
%
%
%
\paragraph{Categorical:} for $y$ taking \textit{unordered} values $l=0,\ldots, L$,
$$p(y|x,\theta_j)= \Cat(y|g^{-1}(m_j(x))),$$
where the link function maps the $L$-dimensional simplex to $\mathbb{R}^L$. For the multivariate logistic link function,
$$\mathbb{P}(y=l|x,\theta_j)=\frac{\exp(m_{j,l}(x))}{1+\sum_{l=1}^L \exp(m_{j,l}(x))},$$ 
for $ l=1,\ldots,L.$ 
For the multinomial probit link function,
\begin{align*}
&\mathbb{P}(y=l|x,\theta_j)=\mathbb{P}(\tilde{y}_l > \max(\tilde{y}_1,\ldots, \tilde{y}_{l-1}, \tilde{y}_{l+1}, \ldots, \tilde{y}_L,0)), 
\end{align*}
for $l=1,\ldots,L,$ 
where $\tilde{y}$ takes values in $\mathbb{R}^L$ and has a  multivariate Gaussian distribution with mean $m_{j}(x)=(m_{j,1}(x),\ldots,m_{j,L}(x))^T$ and covariance matrix $\Sigma_j$, which may be the identity matrix, or treated as a more general scale parameter (in this case, care should be taken to avoid identifiability issues). The prior on the vector-valued unknown function $m_j(x)$ can be extended to independent GPs across $l=1,\ldots,L$ or, more generally, a matrix-variate GP. 
%
%
\paragraph{Ordinal:}  for $y$ taking \textit{ordered} values $l=0,\ldots, L$ and cutoffs $0=\varepsilon_0<\varepsilon_1 <\ldots < \varepsilon_{L-1} $,
$$\mathbb{P}(y \leq l|x,\theta_j)=g^{-1}(\varepsilon_l-m_j(x)),$$ 
where the link function maps $(0,1)$ to the real line. Due to the nonparametric nature of the model we consider fixed cutoffs  $\varepsilon_1, \ldots, \varepsilon_{L-1} $ \citep{kottas2005}. For the logistic link function,
$$\mathbb{P}(y \leq l|x,\theta_j)=\frac{\exp(\varepsilon_l-m_j(x))}{1+\exp(\varepsilon_l-m_j(x))}.$$
For the probit link function,
\begin{align}
\mathbb{P}(y \leq l|x,\theta_j)=\Phi\left(\frac{\varepsilon_l-m_j(x)}{\sigma_j}\right), \label{eq:orderedprobit}
\end{align}
with additional scale parameter $\sigma_j^2$ for $L\geq 2$. In this case, the model can be equivalently formulated through a latent response $\tilde{y}$ that is Gaussian distributed with mean $m_j(x)$ and variance $\sigma_j^2$. In particular, 
$ \tilde{y} | m_j(x), \sigma_j^2 \sim \Norm(m_j(x), \sigma_j^2)$ and $$p(y|\tilde{y})=\left\lbrace \begin{array}{ll}
\1(\tilde{y} \leq 0) & \text{if } l=0 \\
\1( \varepsilon_{l-1} <\tilde{y} \leq \varepsilon_l) & \text{if } l=1,\ldots,L-1 \\
\1( \tilde{y}> \varepsilon_{L-1}) & \text{if } l=L 
\end{array} \right..$$
The ordered probit model is recovered by marginalising the latent $\tilde{y}$.
%
%
\paragraph{Poisson:} for $y \in \lbrace0,1,2,\ldots\rbrace,$
$$p(y|x,\theta_j)= \Pois(y|g^{-1}(m_j(x))),$$
where the link function maps $(0,\infty)$ to $\mathbb{R}$. For the log link function with $\lambda_j(x)=\exp(m_{j}(x))$,
$$\mathbb{P}(y=l|x,\theta_j)=\frac{\exp(-\lambda_j(x)) \lambda_{j}(x)^{l}}{l!}.$$ 
Alternatively, a non-negative integer-valued output $y \in \lbrace0,1,2,\ldots\rbrace$ can be modelled through a discretised  latent Gaussian as in \eqref{eq:orderedprobit}, with fixed cutoffs $\epsilon_1=1, \epsilon_2=2, \ldots$.

\section{Local input models}
\label{app:IM}
	
Other types of inputs can be easily handled through the assumptions of local independence 
\begin{align*}
p(x|\psi)=\prod_{d=1}^D p(x_d|\psi_d), 
\end{align*}
and that each parametric model $p(x_d|\psi_d)$ belongs to the exponential family, that is,
$$p(x_d|\psi_d)=\exp(\psi_d^{T}t_d(x_d)-a_d(\psi_d)+b_d(x_d)),$$
and $t_d$, $a_d$, and $b_d$ are known functions specified by the choice within the exponential family. 
The standard conjugate prior for $\psi$ assumes independence of $\psi_d$ across $d=1,\ldots,D$ with 
$$ \pi(\psi_d)\propto \exp(\psi_d^{T}\tau_d-\nu_d a_d(\psi_d)).$$ In this conjugate setting, the parameters $\psi$ can be marginalised  in each cluster analytically. Specifically, for the collapsed Gibbs sampler,  we need 1) the marginal likelihood $h(x_{n})$ and 2) the predictive likelihood $h(x_n| \mathbf{X}_{l|j}^{-n})$, where $\mathbf{X}_{l|j}^{-n}$ contains $x_{n'}$ such that $n' \neq n, z_{n'}=(j,l)$. Additionally, for the spilt and merge moves we require the the joint marginal likelihood of $h(\mathbf{X}_{l|j})$.  
We note that due to the assumption of local independence
\begin{align*}
 h(x_{n})&= \int p(x_n|\psi) \pi(\psi) d\psi= \prod_{d=1}^D h(x_{n,d}),\\
 h(x_n| \mathbf{X}_{l|j}^{-n})&= \int p(x_n|\psi) \pi(\psi| \mathbf{X}_{l|j}^{-n}) d\psi = \prod_{d=1}^D h(x_{n,d}| \mathbf{X}_{l|j,d}^{-n}),\\
h(\mathbf{X}_{l|j})&= \int \prod_{n:z_n=(j,l)} p(x_n|\psi) \pi(\psi) d\psi = \prod_{d=1}^D h(\mathbf{X}_{l|j,d}).
\end{align*} 
Examples (used in this paper) include:

\paragraph{Gaussian:} for continuous input $x_{n,d}$ taking values in $\mathbb{R}$ with
 $$ p(x_{n,d}|\psi_d)=\Norm(x_{n,d}|u_{d},s_{d}^2),$$
where $\psi_{d}=(u_{d},s_{d}^2)$. The standard conjugate prior is the normal-inverse gamma distribution, 
$$ u_{d}|s_{d}^2 \indsim \Norm(u_{0,d},c^{-1}_d s_{d}^2), \quad s_{d}^2 \indsim \IG(a_{x,d},b_{x,d}),$$
which we denote by $(u_{d},s_{d}^2) \indsim \NIG(u_{0,d},c_d,a_{x,d},b_{x,d})$. 
In this case, marginally $x_{n,d}$ has a non-central $t$-distribution, 
$$h(x_{n,d})= t\left(x_{n,d}| u_{0,d},\frac{b_{x,d}}{a_{x,d}}\frac{c_d+1}{c_d} , 2a_{x,d}\right).$$
The predictive distribution of $x_{n,d}$ given $z_n=(j,l)$ is a non-central $t$-distribution,
$$  h(x_{n,d}| \mathbf{X}_{l|j,d}^{-n})=  t\left(x_{n,d}| \widehat{u}_{l|j,d}^{-n}, \frac{\widehat{b}_{x,l|j,d}^{-n}}{\widehat{a}_{x,l|j,d}^{-n}}\frac{\widehat{c}_{l|j,d}^{-n}+1}{\widehat{c}_{l|j,d}^{-n}}, 2\widehat{a}_{x,l|j,d}^{-n}\right),$$ 
with  $\widehat{c}_{l|j,d}^{-n}=c_{d}+N_{l|j}^{-n}$,  $\widehat{a}_{x,l|j,d}^{-n}=a_{x,d}+N_{l|j}^{-n}/2$, 
$$ \widehat{u}_{l|j,d}^{-n}= \frac{1}{c_d+N_{l|j}^{-n}}(c_d u_{0,d} + N_{l|j}^{-n}\bar{x}_{l|j,d}^{-n}),$$
$$ \widehat{b}_{x,l|j,d}^{-n}=b_{x,d}+\frac{1}{2}\left(c_d u_{0,d}^2-\widehat{c}_{l|j,d}^{-n}(\widehat{u}_{l|j,d}^{-n})^2+\sum_{n' \neq n: z_{n'}=(j,l)} x_{n',d}^2\right),$$
and 	$\bar{x}_{l|j,d}^{-n} =1/N_{l|j}^{-n} \sum_{n' \neq n: z_{n'}=(j,l)} x_{n',d}$.
The joint marginal likelihood of $\mathbf{X}_{l|j,d}$ follows a multivariate $t$ with mean $u_{0,d} 1_{N_{l|j}}$, variance matrix  $$\Sigma_{l|j,d}=\frac{b_{x,d}}{a_{x,d}} \left(I_{N_{l|j}}-\frac{1}{c_d+N_{l|j}} 1_{N_{l|j}} 1_{N_{l|j}}'\right)^{-1},$$ and degrees of freedom $ 2a_{x,d}$, that is
$$h(\mathbf{X}_{l|j,d})=t(\mathbf{X}_{l|j,d}| u_{0,d} 1_{N_{l|j}}, \Sigma_{l|j,d}, 2a_{x,d}).$$

\paragraph{Categorical:} for discrete inputs $x_{n,d}$ taking \textit{unordered} values $g=0,1,\ldots,G_d$ with
$$ p(x_{n,d}|\psi_d)=\psi_{d,x_{n,d}},$$
where $ \psi_d$ is a $G_d+1$ vector of probabilities such that $\sum_{g=0}^{G_d} \psi_{d,g}=1$. The standard conjugate prior is the Dirichlet distribution with parameter $\gamma_{d}=(\gamma_{d,0},\ldots,\gamma_{d,G_d})$. In this case, the marginal likelihood is the Dirichlet-multinomial with
$$ h(x_{n,d})= \frac{\Gamma \left( \sum_{g=0}^{G_d} \gamma_{d,g}\right)}{\Gamma\left(\sum_{g=0}^{G_d} \gamma_{d,g}+1\right)} \frac{\Gamma\left( \gamma_{d,x_{n,d}}+1 \right) }{\Gamma\left( \gamma_{d,x_{n,d}} \right)}.$$
The predictive likelihood of  $x_{n,d}$ given $z_n=(j,l)$ is the Dirichlet-multinomial with 
\begin{align*}
&h(x_{n,d}| \mathbf{X}_{l|j,d}^{-n}) = \frac{\Gamma \left( \sum_{g=0}^{G_d} \gamma_{d,g}+ N_{l|j}^{-n} \right)}{\Gamma\left(\sum_{g=0}^{G_d} \gamma_{d,g}+ N_{l|j}^{-n}+1\right)} \frac{\Gamma\left( \gamma_{d,x_{n,d}}+N_{l|j,x_{n,d}}^{d,-n}+1 \right) }{\Gamma\left( \gamma_{d,x_{n,d}}+N_{l|j,x_{n,d}}^{d,-n} \right)},
\end{align*} 
where $N_{l|j,g}^{d,-n}=\sum_{n' \neq n : z_{n'}=(j,l)} \1( x_{n',d}=g)$. The joint marginal likelihood of $\mathbf{X}_{l|j,d}$ follows a Dirichlet-multinomial with 
$$h(\mathbf{X}_{l|j,d})= \frac{\Gamma \left( \sum_{g=0}^{G_d} \gamma_{d,g} \right)}{\Gamma\left(\sum_{g=0}^{G_d} \gamma_{d,g}+ N_{l|j}\right)}  \prod_{g=0}^{G_d}\frac{\Gamma\left( \gamma_{d,g}+N_{l|j,g}^{d}\right) }{\Gamma\left( \gamma_{d,g} \right)},$$
and $N_{l|j,g}^{d}=\sum_{ n : z_{n}=(j,l)} \1( x_{n,d}=g)$.

\paragraph{Binomial:} for discrete inputs $x_{n,d}$ taking \textit{ordered} values $g=0,1,\ldots,G_d$ with 
$$ p(x_{n,d}|\psi_d)= \left( \begin{array}{c} G_d \\ x_{n,d} \end{array}\right) \psi_{d}^{x_{n,d}} (1-\psi_{d})^{G_d-x_{n,d}} ,$$
where $\psi_d \in (0,1)$. The standard conjugate prior is the beta distribution with parameter $\gamma_d=(\gamma_{d,0},\gamma_{d,1})$. In this case, the marginal likelihood is the beta-binomial with 
\begin{align*}
h(x_{n,d}) &= \left( \begin{array}{c} G_d \\ x_{n,d} \end{array}\right) \frac{\Gamma \left( \gamma_{d,0}+ \gamma_{p,1}\right)}{\Gamma\left( \gamma_{d,0}\right) \Gamma\left( \gamma_{d,1}\right)} \frac{\Gamma\left( \gamma_{d,0} +x_{n,d}\right) \Gamma\left( \gamma_{d,1}+ G_d-x_{n,d}\right) }{\Gamma \left( \gamma_{d,0}+ \gamma_{d,1}+ G_d\right)}.
\end{align*} 
The predictive likelihood of  $x_{n,d}$ given $z_n=(j,l)$ is the beta-binomial with 
\begin{small}
\begin{align*}
 h(x_{n,d}| \mathbf{X}_{l|j,d}^{-n}) &= \left( \begin{array}{c} G_d \\ x_{n,d} \end{array}\right) \frac{\Gamma \left( \gamma_{d,0}+ \gamma_{d,1}+G_d N_{l|j}^{-n}\right)}{\Gamma\left( \widehat{\gamma}_{d,0,l|j}\right) \Gamma\left( \widehat{\gamma}_{d,1,l|j}\right)} \frac{\Gamma\left( \widehat{\gamma}_{d,0,l|j}+x_{n,d}\right) \Gamma\left( \widehat{\gamma}_{d,1,l|j}+ G_d-x_{n,d}\right) }{\Gamma \left( \gamma_{d,0}+ \gamma_{d,1}+ G_d(N_{l|j}^{-n}+1)\right)},
\end{align*}
\end{small}
where $\widehat{\gamma}_{d,0,l|j}= \gamma_{d,0} +  N_{l|j}^{-n} \bar{x}_{l|j,d}^{-n} $ and $\widehat{\gamma}_{d,1,l|j}= \gamma_{d,1} +  N_{l|j}^{-n}(G_d- \bar{x}_{l|j,d}^{-n})$.
The joint marginal likelihood of $\mathbf{X}_{l|j,d}$ follows a Beta-binomial with 
{\small $$h(\mathbf{X}_{l|j,d})= \left[\prod_{n:z_n=(j,l)} \left( \begin{array}{c} G_d \\ x_{n,d} \end{array}\right) \right] \frac{\Gamma \left( \gamma_{d,0}+ \gamma_{d,1}\right)}{\Gamma\left( \gamma_{d,0}\right) \Gamma\left(\gamma_{d,1}\right)} \frac{\Gamma\left( \gamma_{d,0}+N_{l|j} \bar{x}_{l|j,d}\right) \Gamma\left( \gamma_{d,1}+ N_{l|j}(G_d- \bar{x}_{l|j,d})\right) }{\Gamma \left( \gamma_{d,0}+ \gamma_{d,1}+ N_{l|j}G_d\right)}.$$}

\section{Posterior inference}
\label{app:Gibbsenriched}

We present the algorithm for a general setting, when the observed outputs $y_n$ are a deterministic function of latent Gaussian outputs $\tilde{y}_n$. This includes the probit, ordered probit and multinomial probit, as well as the Gaussian example with $y=\tilde{y}$, among others. 
The MCMC algorithm targets the posterior 
\begin{small}
  \begin{align*}
&\pi(z_{1:N},\sigma^{2}_{1:k},\beta_{0,1:k}, \lambda_{1:k},\alpha_\theta, \alpha_{\psi,1:k},\tilde{y}_{1:N} \mid y_{1:N},x_{1:N}) \propto  \prod_{j=1}^k h(\mathbf{\tilde{Y}}_j| \sigma^{2}_{j},\beta_{0,j}, \lambda_{j}) \prod_{l=1}^{k_j} h( \mathbf{X}_{l|j}) \prod_{n=1}^N p(y_n|\tilde{y}_n)\\
&\quad  *\frac{\Gamma(\alpha_\theta)}{\Gamma(\alpha_\theta+N)} \alpha_\theta^k \pi(\alpha_\theta) \prod_{j=1}^k  \alpha_{\psi,j}^{k_j} \frac{ \Gamma(\alpha_{\psi,j})\Gamma(N_j)}{ \Gamma(\alpha_{\psi,j}+N_j)}  \pi(\sigma^2_j) \pi(\beta_{0,j}) \pi(\lambda_j) \pi(\alpha_{\psi,j})\prod_{l=1}^{k_j} \Gamma(N_{l|j}) ,
\end{align*} 
\end{small}  
where we make use of the notation $\mathbf{\tilde{Y}}_j$ to denote the latent outputs $\tilde{y}_n$ such that $z_{y,n}=j$ and $\mathbf{X}_{l|j}$ to denote the inputs $x_n$  such that $z_n=(j,l)$. 
The marginal likelihood of $\mathbf{\tilde{Y}}_j$ given $\beta_{0,j}$, $\lambda_j$ and $\sigma_j^2$, obtained from marginalising the unknown functions $m_j$, is Gaussian, e.g. for the ordered probit,
\begin{align*}
h(\mathbf{\tilde{Y}}_j| \sigma^{2}_{j},\beta_{0,j}, \lambda_{j}) =\Norm(\mathbf{\tilde{Y}}_j \mid  \beta_{0,j} 1_{N_j} , \sigma_j^2I_{N_j}+K_{\lambda_j}),
\end{align*}
where $K_{\lambda_j}$ denotes the $N_j$ by $N_j$ matrix of the kernel function evaluated at every pair of inputs in $y$-cluster $j$. 
The marginal likelihood of $\mathbf{X}_{l|j}$, obtained from marginalising $\psi_{l|j}$, is also available in closed form and factorises over $D$, with examples in Section \ref{app:IM}. 
The term $p(y_n|\tilde{y}_n)$ represents the deterministic function specifying the observed output $y_n$ given the latent Gaussian output $\tilde{y}_n$; examples are provided in Section~\ref{app:GGPM}.

The algorithm is a Gibbs sampler, which alternatively samples each set of parameters, 1) the allocation variables $z_{1:N}$, 2) the unique cluster parameters $(\sigma^2_j,\beta_{0,j},\lambda_j)_{j=1}^k$, 3) the concentration  parameters $\alpha_\theta$  and $\alpha_{\psi,1:k}$ and 4) the latent outputs $\tilde{y}_{1:N}$ (if needed). 
Computations involving the GP are evaluated using \pkg{GPy} in \pkg{Python} \cite{gpy2014}.

\paragraph{Allocation variables.} A non-conjugate collapsed Gibbs sampler is employed, combining Algorithm 3, when cluster parameters can be integrated, and Algorithm 8, when cluster parameters cannot be integrated, of \citet{neal2000markov}, and extending this for the nested partitioning scheme. This consists of $N$ Gibbs steps, where the allocation variable $z_n$  for each data point is updated conditioned on all others $z_1,\ldots, z_{n-1},z_{n+1},\ldots,z_N$. This procedure allows local changes to the allocation variables, and to improve mixing in high-dimensional input spaces, we additionally develop two novel split-merge updates for global changes to the nested partition. Throughout, we make use of the superscript notation $-n$ to denote the data points, parameters, and latent variables with the $n^\text{th}$ data point removed. 

The \textbf{local updates} are described in the following steps:

\begin{enumerate}
\item Remove singleton cluster: 
\begin{itemize}
\item Singleton $y$-cluster: If $z_{y,n} \neq z_{y,n'}$ for all $n' \neq n$, i.e. data point $n$ is in a singleton $y$-cluster, remove that cluster and set $(\sigma_{k^{-n}+1}^2,  \beta_{0,k^{-n}+1},\lambda_{k^{-n}+1}, \alpha_{\psi,k^{-n}+1})$ equal to the values of the singleton cluster parameters.
\item Singleton $x$-cluster within a non-singleton $y$-cluster: If $z_{y,n} = z_{y,n'}$ for some $n' \neq n$ and  $z_{x,n} \neq z_{x,n'}$ for all $n' \neq n$ such that $z_{y,n} = z_{y,n'}$, i.e. data point $n$ is in a singleton $x$-cluster within a non-singleton $y$-cluster, remove that cluster.
\end{itemize}
	\item Calculate the allocation probability for each occupied cluster: $j \in \{1,\ldots,k^{-n}\}$ and $l \in \{1,\ldots,k_j^{-n}\}$
		\begin{align*}
		p(z_n =(j,l) | z^{-n}_{1:N}, \ldots)
		\propto \frac{N_j^{-n} N_{l|j}^{-n}}{\alpha_{\psi,j} + N_j^{-n}  } h(\tilde{y}_n | \mathbf{\tilde{Y}}_j^{-n},\sigma_j^2,\lambda_j, \beta_{0,j}) h(x_n | \mathbf{X}_{l|j}^{-n}).  
	\end{align*}
\item Calculate the allocation probability for a new $x$-cluster within each occupied $y$-cluster: $j \in \{1,\ldots,k^{-n}\}$ 
	\begin{align*}
	p(z_n =(j,k_j^{-n}+1) | z^{-n}_{1:N},\ldots)
	\propto  \frac{N_j^{-n}\alpha_{\psi,j} }{\alpha_{\psi,j} + N_j^{-n}  } h(\tilde{y}_n |\mathbf{\tilde{Y}}_j^{-n},\sigma_j^2,\lambda_j,\beta_{0,j}) h(x_n).  
		\end{align*}
	\item Calculate the allocation probability for $m$ new $y$-clusters:  sample $m$ new parameters (or $m-1$ new parameters if $z_{y,n}$ was in a singleton $y$-cluster) from the prior $(\sigma_{k^{-n}+j}^2, \beta_{0,k^{-n}+j}, \lambda_{k^{-n}+j} , \alpha_{\psi,k^{-n}+j} ) \sim \pi(\sigma^2)\pi(\beta_0) \pi(\lambda) \Gam(u_\psi,v_\psi). $
	Then, for $j=k^{-n}+1,\ldots,k^{-n}+m$, compute 
		\begin{align*}
		p(z_n = (j,1)| \sigma_{j}^2,\beta_{0,j}, \lambda_{j},\alpha_\theta, \tilde{y}_n,x_n)  \propto\frac{\alpha_\theta}{m}   h(\tilde{y}_n| \sigma_{j}^2, \beta_{0,j}, \lambda_{j})h(x_n).  
	\end{align*}	
	\item Update the allocation variable $z_n$ using the allocation probabilities. 
	All empty clusters are removed, and if one of the $m$ new clusters is selected, set $z_n=(k^{-n}+1,1)$ and the parameters $(\sigma_{k^{-n}+1}^2, \beta_{0, k^{-n}+1}, \lambda_{k^{-n}+1}, \alpha_{\psi,k^{-n}+1})$ equal to the parameters of the selected new cluster.
	\end{enumerate}
	After the full Gibbs sweep for the $N$ allocation variables, two Metropolis-Hastings steps are performed to improve mixing and allow global changes to the allocation variables. The first proposes to move an $x$-cluster to be nested within a different or new $y$-cluster and is a `smarter' version of the move described in \citet{wade2014improving}, by proposing moves that are more likely to be accepted. This step is separated into three possible moves: 1) an $x$-cluster, among those within $y$-clusters with more than one $x$-cluster, is moved to a different $y$-cluster; 2) an $x$-cluster, among those within $y$-clusters with more than one $x$-cluster, is moved to a new $y$-cluster; 3) an $x$-cluster, among those within $y$-clusters with only one $x$-cluster, is moved to a different $y$-cluster.
%
	Define
	$$ k_{x,2+}=\sum_{j=1}^k k_j\1(k_j>1) \quad \text{and} \quad k_{x,1}=\sum_{j=1}^k \1(k_j=1).$$ 
	At every iteration, Move 1 is performed if $k_{x,2+}>0$. Next, with probability $1/2$, Move 2 is performed, otherwise, Move 3 is performed (with the exception that when $k_{x,1}=0$, Move 2 is performed with probability 1, or when $k_{x,2+}=0$,  Move 3 is performed with probability 1).

The \textbf{global updates} to the \textbf{$y$-clusters} are described in the following steps:
	\begin{enumerate}
	\item \textbf{Move 1:} an $x$-cluster (nested within a $y$-cluster with more than one $x$-cluster) is uniformly selected with probability $k_{x,2+}^{-1}$ and moved to be nested within a different $y$-cluster selected with probability proportional to the conditional marginal likelihood. Specifically, suppose $x$-cluster $l$ in $y$-cluster $j$ is first selected, then it is moved to be nested within $y$-cluster $h$ with probability proportional to $h(\bm{\tilde{Y}}_{l|j} | \bm{\tilde{Y}}_h , \sigma^2_h,\beta_{0,h},\lambda_h)$.  Let $z_{1:N}^*$ denote the proposed allocations defined by moving $x$-cluster $l$ in $y$-cluster $j$ to be nested within $y$-cluster $h$ for $h \in \lbrace 1,\ldots,j-1,j+1,\ldots, k\rbrace$. The acceptance probability is $\min(1,p)$, where
		\begin{align*}
	p&=\frac{\Gamma(N_j-N_{l|j})\Gamma(N_h+N_{l|j})}{\Gamma(N_j)\Gamma(N_h)} \frac{\Gamma(\alpha_{\psi,j}+N_j)\Gamma(\alpha_{\psi,h}+N_h)}{\Gamma(\alpha_{\psi,j}+N_j-N_{l|j})\Gamma(\alpha_{\psi,h}+N_h+N_{l|j})}\\
	&* \frac{\alpha_{\psi,h}}{\alpha_{\psi,j}}  \frac{k_{x,2+}}{k_{x,2+}^*} \frac{\sum_{h' \neq j}h(\bm{\tilde{Y}}_{l|j} | \bm{\tilde{Y}}_{h'} , \sigma^2_{h'},\beta_{0,h'},\lambda_{h'})}{\sum_{h' \neq h} h(\bm{\tilde{Y}}_{l|j} | \bm{\tilde{Y}}_{h'}^* , \sigma^2_{h'},\beta_{0,h'},\lambda_{h'})},
	\end{align*}
	where $\mathbf{\tilde{Y}}_{h'}^*$ contains the outputs under the proposed allocation with $z_{y,n}^*=h'$, e.g. $\mathbf{\tilde{Y}}_j^*$ contains the $N_j-N_{l|j}$ outputs 
	with the $N_{l|j}$ points removed from $y$-cluster $j$. 
	The notation $k_{x,2+}^*$ represents the number of $x$-clusters within a $y$-cluster with more than one $x$-cluster under the proposed partition, i.e. $k_{x,2+}^*=k_{x,2+}-\1(k_j=2)+\1(k_h=1)$.
	\item \textbf{Move 2}: an $x$-cluster (nested within a $y$-cluster with more than one $x$-cluster) is uniformly selected with probability $k_{x,2+}^{-1}$ and moved to be nested within a new $y$-cluster. In this case, we propose new parameters $(\sigma_{k+1}, \beta_{0,k+1},\lambda_{k+1}, \alpha_{\psi,k+1})$ for the new $y$-cluster from the prior. 
The acceptance probability is $\min(1,p)$, where
	\begin{align*}
	p&=\frac{\Gamma(N_j-N_{l|j})\Gamma(N_{l|j})}{\Gamma(N_j)} \frac{\Gamma(\alpha_{\psi,j}+N_j)\Gamma(\alpha_{\psi,k+1})}{\Gamma(\alpha_{\psi,j}+N_j-N_{l|j})\Gamma(\alpha_{\psi,k+1}+N_{l|j})}  \\
	&* \alpha_\theta \frac{\alpha_{\psi,k+1}}{\alpha_{\psi,j}}  \frac{k_{x,2+}}{k_{x,1}} \frac{h(\bm{\tilde{Y}}_{k+1}^* |  \sigma^2_{k+1},\beta_{0,k+1},\lambda_{k+1})}{ \sum_{h=1}^k  h(\bm{\tilde{Y}}_{l|j} | \bm{\tilde{Y}}_{h}^* , \sigma^2_{h},\beta_{0,h},\lambda_{h})}, 
	\end{align*}
where $k_{x,1}^*=k_{x,1}+1+\1(k_j=2)$ represents the number of $x$-clusters within a $y$-cluster with only one $x$-cluster under the proposed partition.
	\item \textbf{Move 3}: an $x$-cluster (nested within a $y$-cluster with only one $x$-cluster) is uniformly selected with probability $k_{x,1}^{-1}$ and moved to be nested within a different $y$-cluster selected with probability proportional to the conditional marginal likelihood. Specifically, suppose $x$-cluster $l$ in $y$-cluster $j$ is first selected, then it is moved to be nested within $y$-cluster $h$ with probability proportional to $h(\bm{\tilde{Y}}_{j} | \bm{\tilde{Y}}_h , \sigma^2_h,\beta_{0,h},\lambda_h)$ . Let $z_{1:N}^*$ denote the proposed allocations defined by moving $x$-cluster $l$ in $y$-cluster $j$ to be nested within $y$-cluster $h$ for $h \in \lbrace 1,\ldots,j-1,j+1,\ldots, k \rbrace$. The acceptance probability is $\min(1,p)$, where
		\begin{align*}
	p&=\frac{\Gamma(N_h+N_{j})}{\Gamma(N_h)\Gamma(N_{j})} \frac{\Gamma(\alpha_{\psi,j}+N_j)\Gamma(\alpha_{\psi,h}+N_h)}{\Gamma(\alpha_{\psi,h}+N_h+N_j)\Gamma(\alpha_{\psi,j})} \frac{1}{\alpha_\theta} \frac{\alpha_{\psi,h}}{\alpha_{\psi,j}}\\
	&* \frac{k_{x,1}}{k_{x,2+}^*}  \frac{ \sum_{h' \neq j}  h(\bm{\tilde{Y}}_{j} | \bm{\tilde{Y}}_{h'} , \sigma^2_{h'},\beta_{0,h'},\lambda_{h'})}{h(\bm{\tilde{Y}}_j | \sigma^2_j,\beta_{0,j} ,\lambda_j)},
	\end{align*}
\end{enumerate}

The second set of split-merge updates consists of the pair of 'smart-split' and 'dumb-merge' moves and the pair of 'dumb-split' and 'smart-split' moves, inspired from \cite{wang2015}, but tailored for the nested clustering structure of the EDP to propose global updates to the $x$-clusters. In this split moves, one $x$-cluster is selected and split into two $x$-clusters, still contained within the same $y$-cluster. In the merge moves, two $x$-clusters, within the same $y$-cluster are merged. The 'smart' moves propose clustering allocations that are more likely and are paired with the corresponding 'dumb' moves, with random cluster allocations, to increase the probability of the reverse move and acceptance of the smart moves. In the first pair of moves, a smart-split or dumb-merge is proposed with probability $1/2$, and in the second pair of moves, a dumb-split or smart-merge is proposed with probability $1/2$ (unless, there are only singleton $x$-clusters or only one $x$-cluster within each $y$-cluster). Again, we define $k_{x,2+}$ as the number of $x$-clusters within a $y$-cluster with more than one $x$-cluster, i.e. the number $x$-clusters than may be merged, and additionally define 
$$ k_{x,1+}=\sum_{j=1}^k \sum_{l=1}^k \1(N_{l|j}>1),$$
as the number of $x$-clusters with more than one data point, i.e. the number $x$-clusters than may be split. 

The \textbf{global updates} to the $x$-clusters are described in the following steps: 
\begin{itemize}
\item \textbf{Smart-Split/Dumb-Merge:} with probability $1/2$, one of the following two moves is proposed.
\begin{enumerate}
\item \textbf{Smart-Split:} $x$-cluster $l$ within $y$-cluster $j$ is selected among the $k_{x,1+}$ $x$-clusters containing more than one data point with probability proportional to $1/h(\bm{X}_{l|j})$. 
The proposed allocation $z^*_{1:N}$ is constructed sequentially by reallocating the data points currently allocated to  $x$-cluster $l$ within $y$-cluster $j$, in order of observation, to $x$-cluster $l$ or a new $x$-cluster $k_j+1$ with sequential probabilities
\begin{align*}
z_n^*| z_{1:n-1}^*=\left\lbrace \begin{array}{ll}
(j,l) & \text{w.p. } \propto h(x_n| \bm{X}_{l|j,n-1}^*)\\
(j,k_j+1) & \text{w.p. } \propto h(x_n| \bm{X}_{k_j+1|j,n-1}^*)
\end{array} \right. \quad \text{for } n \text{ s.t. } z_n=(j,l),
\end{align*}
where $\bm{X}_{l|j,n-1}^*$ denotes the set of $x_{n'}$ such that $z_{n'}^*=(j,l)$ for $n'<n$, and similarly, $\bm{X}_{k_j+1|j,n-1}^*$ denotes the set of $x_{n'}$ such that $z_{n'}^*=(j,k_j+1)$ for $n'<n$. Note that through sequential allocation, there is a positive probability that all points may be allocated to one cluster, 
and in that case the move is accepted with probability one, i.e. we remain at the current allocation. The probability of proposing the smart-split $z^*_{1:N}$ from $z_{1:N}$ is
\begin{align*}
&q_{\text{SS}}(z^*_{1:N}|z_{1:N})\\
&=\frac{h(\bm{X}_{l|j})^{-1}}{\sum_{(j',l'): N_{l'|j'}>1}h(\bm{X}_{l'|j'})^{-1}} \prod_{n:z_n=(j,l)} \frac{2h(x_n| \bm{X}_{z^*_{x,n}|j,n-1}^*)}{h(x_n| \bm{X}_{l|j,n-1}^*)+h(x_n| \bm{X}_{k_j+1|j,n-1}^*)},
\end{align*}
where the factor of $2$ is is needed as the proposed state $z^*_{1:N}$ is equivalent when the labels $l$ and $k_j+1$ are interchanged. 
The acceptance probability is $\min(1,p)$, where
\begin{align*}
p&= \frac{\alpha_j \Gamma(N_{l|j}^*)\Gamma(N_{k_j+1|j}^*)h(\bm{X}^*_{l|j})h(\bm{X}^*_{k_j+1|j})}{\Gamma(N_{l|j})h(\bm{X}_{l|j})} \frac{1}{k_{x,2+}^*k_{j}} \frac{\sum_{(j',l'): N_{l'|j'}>1}h(\bm{X}_{l'|j'})^{-1}}{h(\bm{X}_{l|j})^{-1}}\\
&* \prod_{n:z_n=(j,l)} \frac{h(x_n| \bm{X}_{l|j,n-1}^*)+h(x_n| \bm{X}_{k_j+1|j,n-1}^*)}{h(x_n| \bm{X}_{z^*_{x,n}|j,n-1}^*)} ,
\end{align*}
where $k_{x,2+}^*$ is equal to $k_{x,2+}+2$ if $x$-cluster $l$ was the only cluster within $y$-cluster $j$, i.e. $k_j=1$, and $k_{x,2+}^*$ is equal to $k_{x,2+}+1$ otherwise.

\item \textbf{Dumb-Merge:} $x$-cluster $l$ within $y$-cluster $j$ is selected uniformly among the $k_{x,2+}$ $x$-clusters contained within a $y$-cluster with more than one $x$-cluster with probability $1/k_{x,2+}$ and a second $x$-cluster $l' \neq l$ within $y$-cluster $j$ is selected uniformly among the $k_{j}-1$ remaining $x$-clusters within $y$-cluster $j$ with probability $1/(k_{j}-1)$. The probability of proposing the dumb-merge $z^*_{1:N}$ from $z_{1:N}$ is
\begin{align*}
q_{\text{DM}}(z^*_{1:N}|z_{1:N})=\frac{2}{k_{x,2+}(k_{j}-1)},
\end{align*}
where the factor of $2$ is needed as the proposed state $z^*_{1:N}$ can also be reached by first selecting $x$-cluster $l'$ within $y$-cluster $j$ and then selecting $x$-cluster $l$ within $y$-cluster $j$. 
The acceptance probability is $\min(1,p)$, where
\begin{align*}
p&= \frac{\Gamma(N_{l|j}^*)h(\bm{X}_{l|j}^*)}{\alpha_j \Gamma(N_{l|j})\Gamma(N_{l'|j})h(\bm{X}_{l|j})h(\bm{X}_{l'|j})} k_{x,2+}(k_{j}-1) \frac{h(\bm{X}_{l|j}^*)^{-1}}{\sum_{(j',l'): N_{l'|j'}^*>0}h(\bm{X}^*_{l'|j'})^{-1}}\\
&* \prod_{n:z_n^*=(j,l)} \frac{h(x_n| \bm{X}_{z_{x,n}|j,n-1})}{h(x_n| \bm{X}_{l|j,n-1})+h(x_n| \bm{X}_{l'|j,n-1})} .
\end{align*}
\end{enumerate}

\item \textbf{Dumb-Split/Smart-Merge:} with probability $1/2$, one of the following two moves is proposed.
\begin{enumerate}
\item \textbf{Dumb-Split:} $x$-cluster $l$ within $y$-cluster $j$ is uniformly selected among the $k_{x,1+}$ $x$-clusters containing more than one data point with probability $1/k_{x,1+}$. The data points in $x$-cluster $l$ within $y$-cluster $j$ are then randomly reallocated to $x$-cluster $l$ or a new $x$-cluster $k_j+1$ with probability $1/2$. Note that again there is a positive probability that all points may be allocated to one cluster, 
and in that case the move is accepted with probability one, i.e. we remain at the current allocation. The probability of proposing the dumb-split $z^*_{1:N}$ from $z_{1:N}$ is
\begin{align*}
q_{\text{DS}}(z^*_{1:N}|z_{1:N})=\frac{1}{k_{x,1+}}\frac{2}{2^{N_{l|j}}}=\frac{1}{k_{x,1+}}\frac{1}{ 2^{N_{l|j}-1}},
\end{align*}
where the factor of $2$ is is needed as the proposed state $z^*_{1:N}$ is equivalent when the labels $l$ and $k_j+1$ are interchanged. 
The acceptance probability is $\min(1,p)$, where
\begin{align*}
p&= \frac{\alpha_j \Gamma(N_{l|j}^*)\Gamma(N_{k_j+1|j}^*)h(\bm{X}^*_{l|j})h(\bm{X}^*_{k_j+1|j})}{\Gamma(N_{l|j})h(\bm{X}_{l|j})} k_{x,1+} 2^{N_{l|j}-1} \frac{1}{k_{x,2+}^*}\\
&*\left(\frac{h(\bm{X}_{l|j})}{\sum_{h\neq l} h(\bm{X}^*_{(l,h)|j})} +\frac{h(\bm{X}_{l|j})}{\sum_{h\neq k_j+1} h(\bm{X}^*_{(k_j+1,h)|j})} \right).
\end{align*}

\item \textbf{Smart-Merge:} $x$-cluster $l$ within $y$-cluster $j$ is selected uniformly among the $k_{x,2+}$ $x$-clusters contained within a $y$-cluster with more than one $x$-cluster with probability $1/k_{x,2+}$. A second $x$-cluster $l' \neq l$ within $y$-cluster $j$ is selected among the $k_{j}-1$ remaining $x$-clusters within $y$-cluster $j$ with probability proportional to $h(\bm{X}_{(l,l')|j})$, where $\bm{X}_{(l,l')|j}$ denotes the set of $x_n$ under the merger of $x$-clusters $l$ and $l'$. The probability 
 of proposing the smart-merge $z^*_{1:N}$ from $z_{1:N}$ is
\begin{align*}
q_{\text{SM}}(z^*_{1:N}|z_{1:N})=\frac{1}{k_{x,2+}}\left(\frac{h(\bm{X}_{(l,l')|j})}{\sum_{h\neq l} h(\bm{X}_{(l,h)|j})} +\frac{h(\bm{X}_{(l,l')|j})}{\sum_{h\neq l'} h(\bm{X}_{(l',h)|j})} \right),
\end{align*}
which is the sum of the probability of first selecting $l$ and then $l'$ and vice versa, as the proposed state $z^*_{1:N}$ is equivalent under these proposals. 
The acceptance probability is $\min(1,p)$, where
\begin{align*}
p&= \frac{\Gamma(N_{l|j}^*)h(\bm{X}_{l|j}^*)}{\alpha_j \Gamma(N_{l|j})\Gamma(N_{l'|j})h(\bm{X}_{l|j})h(\bm{X}_{l'|j})} \frac{1}{k_{x,1+}^*}\frac{1}{ 2^{N_{l|j}^*-1}} k_{x,2+}\\
&* \left(\frac{h(\bm{X}_{l|j}^*)}{\sum_{h\neq l} h(\bm{X}_{(l,h)|j})} +\frac{h(\bm{X}^*_{l|j})}{\sum_{h\neq l'} h(\bm{X}_{(l',h)|j})} \right)^{-1},
\end{align*}
where $k_{x,1+}^*$ is equal to $k_{x,1+}+1$ if two singleton clusters are merged; $k_{x,1+}$ if one of merged clusters is a singleton;  and $k_{x,1+}-1$ if neither cluster is a singleton.
\end{enumerate}
\end{itemize}

\paragraph{Cluster parameters.}	 The parameters for each cluster are conditionally independent across $j=1,\ldots,k$ with full conditional 
		\begin{equation*}
		\pi(\sigma^2_j,\beta_{0,j}, \lambda_j | \mathbf{\tilde{Y}}_j ) \propto h(\mathbf{\tilde{Y}}_j |  \sigma^2_j,\beta_{0,j} ,\lambda_j)\pi(\sigma^2_j) \pi(\beta_{0,j}) \pi(\lambda_j),
		\end{equation*}		
which is not available in closed form. We use Hamiltonian Monte Carlo \citep{duane1987hybrid} to sample from the full conditional.

\paragraph{Mass parameters.} 
The concentration parameters $\alpha_\theta$ and  $\alpha_{\psi,1:k}$ are updated using the auxiliary variable technique of \citet{EW95}. For $\alpha_\theta$, sample an auxiliary variable $\xi \sim\Be(\alpha_\theta+1,N)$; set  $\widehat{v}_\theta=v_\theta-\log(\xi)$ and
	\begin{align*}
	 \widehat{u} _\theta= \left\lbrace \begin{array}{cc} u_\theta +k-1 & \text{ w.p. } \frac{N\widehat{v}_\theta}{N\widehat{v}_\theta+u_\theta +k-1} \\
	  u_\theta +k &\text{ w.p. } \frac{u_\theta +k-1}{N\widehat{v}_\theta+u_\theta +k-1}
	\end{array} \right.;\\
	\end{align*}
	and sample $\alpha \sim \Gam(\widehat{u}_\theta, \widehat{v}_\theta).$ 
Similarly, for $\alpha_{\psi,j}$, 
for $j =1,\ldots,k$, sample an auxiliary variable $\xi_j \sim\Be(\alpha_{\psi,j}+1,N_j)$; set $\widehat{v}_{\psi,j}=v_\psi-\log(\xi_j)$ and
		\begin{align*}
		  \widehat{u}_{\psi,j}= \left\lbrace \begin{array}{cc} u_\psi +k_j-1 & \text{ w.p. } \frac{N_j\widehat{v}_{\psi,j}}{N_j\widehat{v}_{\psi,j}+u_\psi +k_j-1} \\
  u_\psi +k_j& \text{ w.p. } \frac{u_\psi +k_j-1}{N_j\widehat{v}_{\psi,j}+u_\psi +k_j-1}
	\end{array} \right.;\\
	\end{align*}
	and sample $\alpha_{\psi,j} \sim \Gam(\widehat{u}_{\psi,j}, \widehat{v}_{\psi,j}).$

\paragraph{Latent outputs.} The latent outputs are independent across cluster $j=1,\ldots,k$, with full conditional 
	\begin{equation*}
		\pi( \mathbf{\tilde{Y}}_j| \mathbf{Y}_j, \sigma^2_j,\beta_{0,j}, \lambda_j ) \propto h(\mathbf{\tilde{Y}}_j |  \sigma^2_j,\beta_{0,j} ,\lambda_j)\prod_{n:z_n=j} p(y_n|\tilde{y}_n).
		\end{equation*}	
In the Gaussian case, $p(y_n|\tilde{y}_n)=\1(y_n=\tilde{y}_n)$, and this step is not needed. For the other probit-type models, the full conditional of the latent outputs in cluster $j$ is a truncated multivariate Gaussian, which is sampled through a Gibbs algorithm combined with cumulative distribution function inversion techniques \citep{kotecha1999gibbs}. 

\section{Predictions for the enriched mixture of generalised GP experts}
\label{app:Predenriched}

Letting $\zeta = (z_{1:N}, \sigma^{2}_{1:k}, \beta_{0,1:k}, \lambda_{1:k}, \alpha_\theta,  \alpha_{\psi,1:k},  \tilde{y}_{1:N})$ denote the model parameters and latent variables, the MCMC algorithm provides samples $\zeta^{(m)}$, for $m=1,\ldots,M$, from the posterior.  
In the Gaussian example, the posterior density at $y_*$ given a new $x_*$ is given by
\begin{align*}
&f(y_*|x_*,y_{1:N},x_{1:N})= \int f(y_*|x_*,y_{1:N},x_{1:N},\zeta ) \frac{\pi(\zeta |y_{1:N},x_{1:N})f(x_*|x_{1:N},\zeta)}{f(x_*|x_{1:N})} d\zeta \nonumber \\
&\approx \frac{1}{f(x_*|x_{1:N})}\sum_{m=1}^M f(y_*|x_*,y_{1:N},x_{1:N},\zeta^{(m)} ) f(x_*|x_{1:N},\zeta^{(m)}) \nonumber \\
&\approx  C^{-1} \left( \sum_{m=1}^M p^{(m)}_{k^{(m)}+1}(x_*) h(y_*)  +\sum_{j=1}^{k^{(m)}}  p^{(m)}_{j}(x_*)  h(y_*|\mathbf{Y}_j^{(m)}, \beta_{0,j}^{(m)},\lambda_j^{(m)}, \sigma_j^{2\,(m)} )  \right),
\end{align*}
with $$f(x_*|x_{1:N}) \approx C :=  \sum_{m=1}^M p^{(m)}_{k^{(m)}+1}(x_*) +\sum_{j=1}^{k^{(m)}}  p^{(m)}_{j}(x_*) .$$ 
In this case, we have a weighted average of the GP predictive densities across clusters and the marginal likelihood $h(y_*)$ for a new cluster. 
Note that the marginal likelihood $h(y_*)$ for a new cluster is unavailable in closed form as it requires integration over the parameters $(\beta_0,\lambda,\sigma^2)$. However, we can compute a simple Monte Carlo estimate of this quantity by sampling from the prior,
$$ h(y_*) \approx \frac{1}{S}\sum_{s=1}^S \Norm(y_* \mid \beta_0^s, \sigma^{2\,s}+K_{\lambda^s}(x_*,x_*)),$$
with $(\sigma^{2\,s}, \beta_0^s,\lambda^s)$ i.i.d. samples from the prior. 

For other types of outputs through probit models, we can similarly use the MCMC output to compute predictive quantities of interest at a test input $x_*$. For example, considering the ordered probit  with  ordered categories $l=0,\ldots, L$ and fixed cutoffs $0=\varepsilon_0<\varepsilon_1 <\ldots < \varepsilon_{L-1} $, 
we first note that we can compute the expectation and density of the latent continuous $\tilde{y}_*$ given the test input $x_*$, as in the Gaussian example. 
The posterior probability that $y_*=l$ given the test input $x_*$ is
\begin{align*}
&\mathbb{P}(y_*=l|x_*,y_{1:N},x_{1:N}) = \int \mathbb{P}(y_*=l|x_*,y_{1:N},x_{1:N},\zeta )\frac{\pi(\zeta |y_{1:N},x_{1:N})f(x_*|x_{1:N},\zeta)}{f(x_*|x_{1:N})} d\zeta \nonumber \\
&\approx  C^{-1} \left( \sum_{m=1}^M p^{(m)}_{k^{(m)}+1}(x_*) \mathbb{P}(y_*=l|x_*)+\sum_{j=1}^{k^{(m)}}  p^{(m)}_{j}(x_*) \mathbb{P}(y_*=l|x_*,\mathbf{\tilde{Y}}_j^{(m)}, \sigma^{2\,(m)}_j,\beta^{(m)}_{0,j},\lambda^{(m)}_j)  \right).
\end{align*}
For cluster $j$ of sample $m$, the probability that $y_*=l$ is
\begin{align*}
 &\mathbb{P}(y_*=l|x_*,\mathbf{\tilde{Y}}_j^{(m)}, \sigma^{2\,(m)}_j,\beta^{(m)}_{0,j},\lambda^{(m)}_j) = \mathbb{P}(\varepsilon_{l-1} <\tilde{y}_* \leq \varepsilon_{l}  |x_*,\mathbf{\tilde{Y}}_j^{(m)}, \sigma^{2\,(m)}_j,\beta^{(m)}_{0,j},\lambda^{(m)}_j)\\
  &= \Phi \left( \frac{\varepsilon_{l} - \widehat{m}_j^{(m)}(x_*)}{ \sqrt{\widehat{K}_j^{(m)}(x_*,x_*)+\sigma^{2\,(m)}_j}} \right)  - \Phi \left( \frac{\varepsilon_{l-1} - \widehat{m}_j^{(m)}(x_*)}{ \sqrt{\widehat{K}_j^{(m)}(x_*,x_*)+\sigma^{2\,(m)}_j}} \right),
\end{align*}
with $\varepsilon_{-1}=-\infty$, $\varepsilon_{L}=\infty$ and $\widehat{m}_j^{(m)}(x_*)$ and $\widehat{K}_j^{(m)}(x_*,x_*)$ denoting the GP predictive mean and kernel functions  in cluster $j$ of sample $m$. 
For a new cluster,  the marginal probability $\mathbb{P}(y_*=l|x_*)$ is unavailable in closed form as it requires integration over the parameters $(\beta_0,\lambda,\sigma^2)$. We can again employ a Monte Carlo approach to estimate this quantity,
\begin{align*}
\mathbb{P}(y_*=l|x_*) \approx \frac{1}{S}\sum_{s=1}^S  &\Phi \left( \frac{\varepsilon_{l} -  \beta_0^s}{ \sqrt{K_{\lambda^s}(x_*,x_*))+\sigma^{2\,s}}} \right) - \Phi \left( \frac{\varepsilon_{l-1} -  \beta_0^s}{ \sqrt{K_{\lambda^s}(x_*,x_*))+\sigma^{2\,s}}} \right),
\end{align*}
with $(\sigma^{2\,s},{\beta_0^s},\lambda^s)$ i.i.d. samples from the prior. 

An advantage of jointly modelling the outputs and inputs includes the possibility to compute the predictive distribution of $y_*$ based only on a subset of inputs, say only based on a single input $x_{*d}$. 
In this case, the weights would only involve the local predictive marginal likelihood of $x_{*d}$ for each cluster $h(x_{*d}|\mathbf{X}_{l|j,d}^{(m)})$, $j=1,\ldots,k^{(m)}$ and $l=1,\ldots,k_j^{(m)}$, and for a new cluster $h(x_{*d})$.
 However, the local expectation would need to be integrated with respect to predictive marginal likelihood of $x_{*-d}=(x_{*1},\ldots,x_{*d-1},x_{*d+1},\ldots,x_{*D})$ in each nested clustering. 
 For example, in the Gaussian case,
\begin{align*}
\mathbb{E}[y_*|x_{*d},y_{1:N},x_{1:N}]\approx C_d^{-1} &\left(\sum_{m=1}^M  p^{(m)}_{k^{(m)}+1}(x_{*d}) \mu_\beta + \sum_{j=1}^{k^{(m)}} p^{(m)}_{j,1}(x_{*d})\mathbb{E}_{x_{*-d}}[ \widehat{m}_j^{(m)}(x_*)] \right.\\
&\left.+ \sum_{j=1}^{k^{(m)}} \sum_{l=1}^{k_j^{(m)}} p^{(m)}_{j,l}(x_{*d}) \mathbb{E}_{x_{*-d}}[ \widehat{m}_j^{(m)}(x_*)|\mathbf{X}_{l|j,-d}^{(m)} ]
 \right),
\end{align*}
where expectations are taken with respect to  $h(x_{*-d})$ and $h(x_{*-d}|\mathbf{X}_{l|j,-d}^{(m)})$, i.e.  
\begin{align*}
 \mathbb{E}_{x_{*-d}}[ \widehat{m}_j^{(m)}(x_*)] &= \int  \widehat{m}_j^{(m)}(x_*) \prod_{d' \neq d } h(x_{*d'}) dx_{*-d}, \\
   \mathbb{E}_{x_{*-d}}[ \widehat{m}_j^{(m)}(x_*)|\mathbf{X}_{l|j,-d}^{(m)}] &= \int  \widehat{m}_j^{(m)}(x_*)    \prod_{d' \neq d } h(x_{*d'}|\mathbf{X}_{l|j,d'}^{(m)}) dx_{*-d},
 \end{align*}
with
\begin{align*}
\begin{split}
p^{(m)}_{k^{(m)}+1}(x_{*d})=\frac{\alpha^{(m)}_\theta}{\alpha^{(m)}_\theta+N} h(x_{*d}); &\quad \quad p^{(m)}_{j,1}(x_{*d})= \frac{ N_j^{(m)}}{\alpha^{(m)}_\theta+N} \frac{\alpha_{\psi,j}^{(m)}}{\alpha_{\psi,j}^{(m)}+N_j^{(m)}} h(x_{*d})\\
p^{(m)}_{j,l}(x_{*d})&= \frac{ N_j^{(m)}}{\alpha^{(m)}_\theta+N}  \frac{N_{l|j}^{(m)}}{\alpha_{\psi,j}^{(m)}+N_j^{(m)}} h(x_{*d}|\bm{X}_{l|j,d}^{(m)}),
\end{split} 
\end{align*}
with $C_d= \sum_{m=1}^M p^{(m)}_{k^{(m)}+1}(x_{*d})+\sum_{j=1}^{k^{(m)}} p^{(m)}_{j,1}(x_{*d})+\sum_{j=1}^{k^{(m)}} \sum_{l=1}^{k_j^{(m)}} p^{(m)}_{j,l}(x_{*d})$.

%

\begin{figure}[th]
\begin{center}
	\subfigure[$D=1$: $y$-cluster 1]{\includegraphics[width=0.3\textwidth]{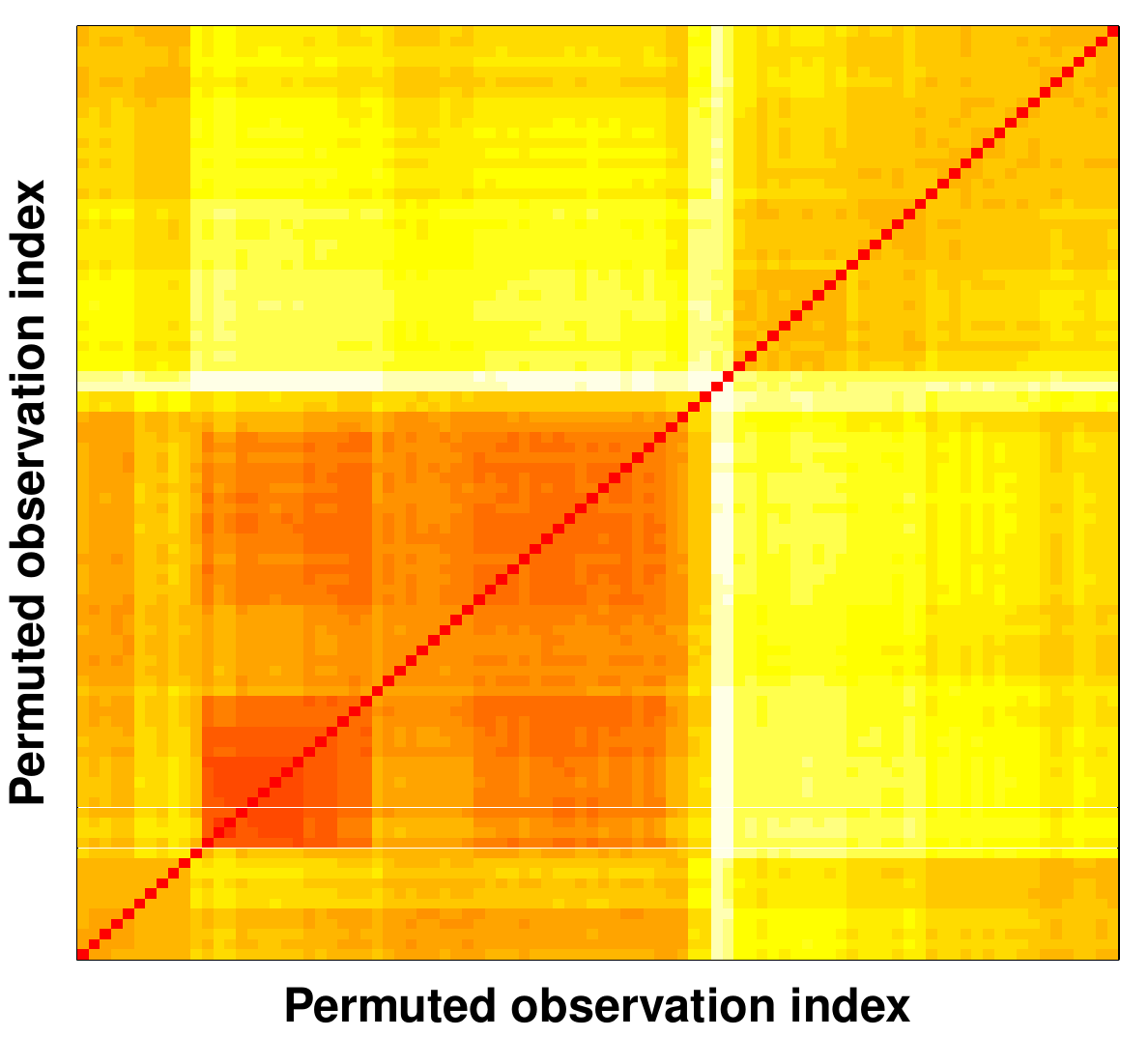}\label{fig:adni_EmE_psm1}}
	\subfigure[$D=5$: $y$-cluster 1]{\includegraphics[width=0.3\textwidth]{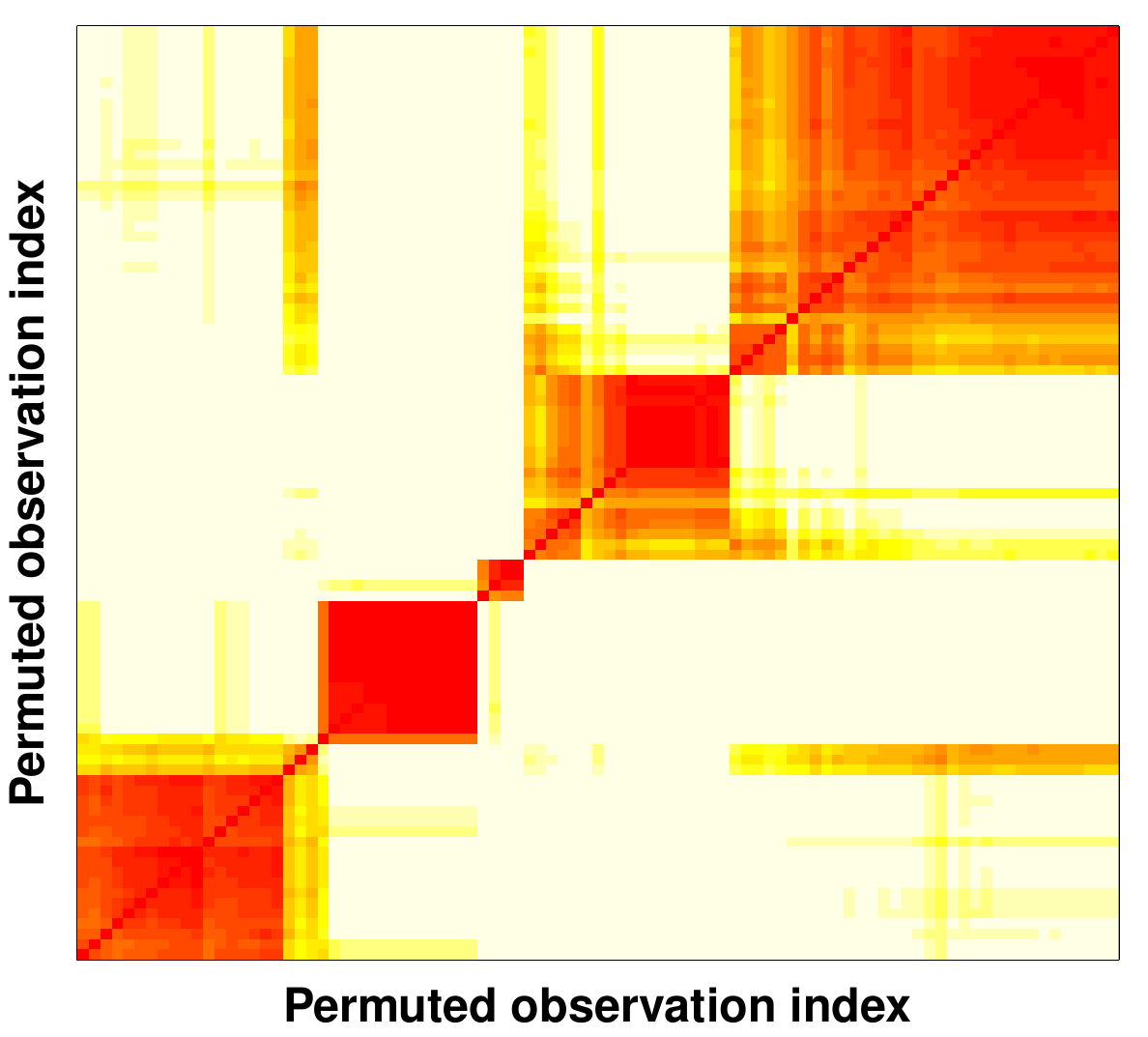}\label{fig:adni_EmE_psm1}}
	\subfigure[$D=10$: $y$-cluster 1]{\includegraphics[width=0.3\textwidth]{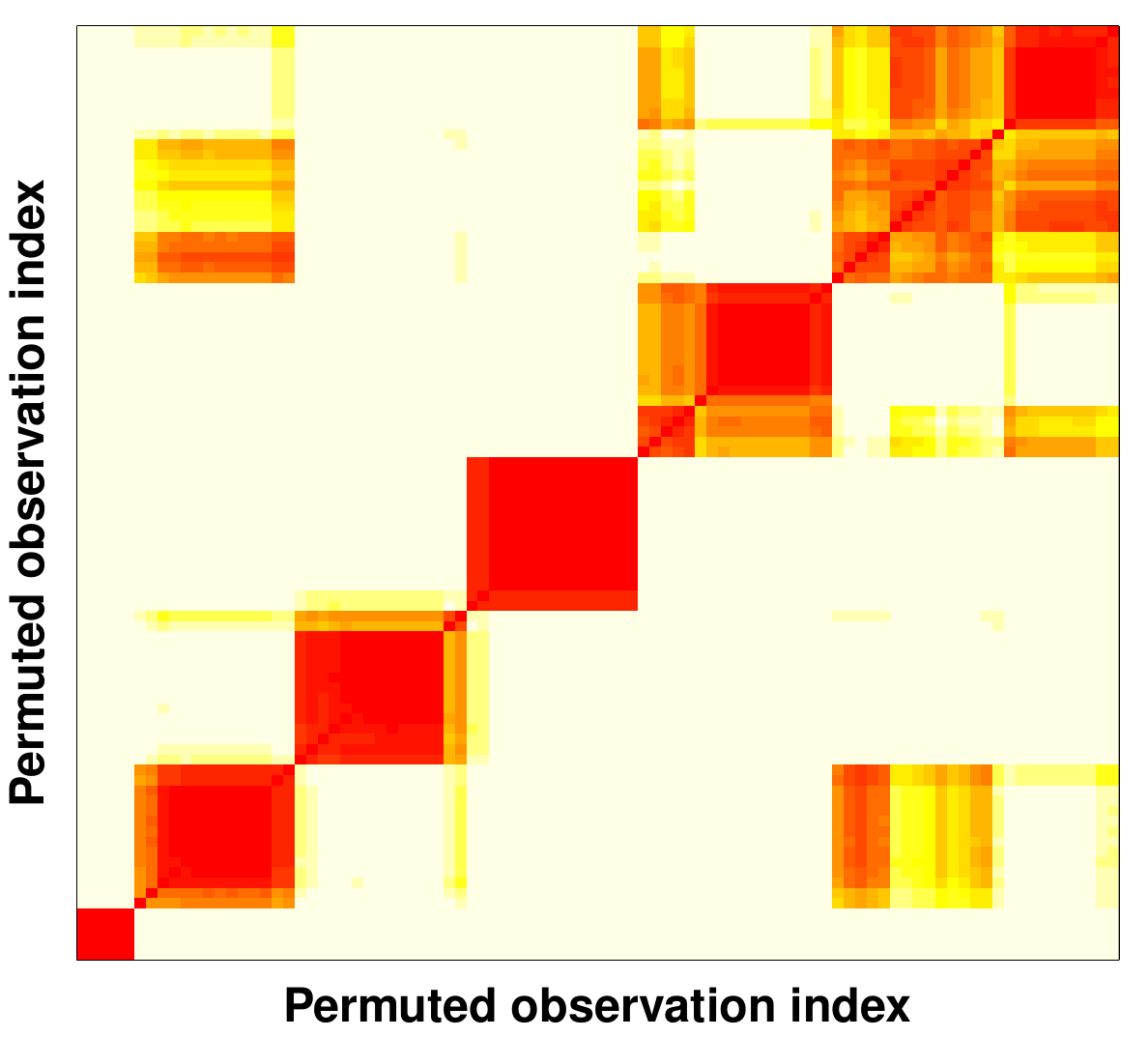}\label{fig:adni_EmE_psm1}}
\\
		\subfigure[$D=1$: $y$-cluster 2]{\includegraphics[width=0.3\textwidth]{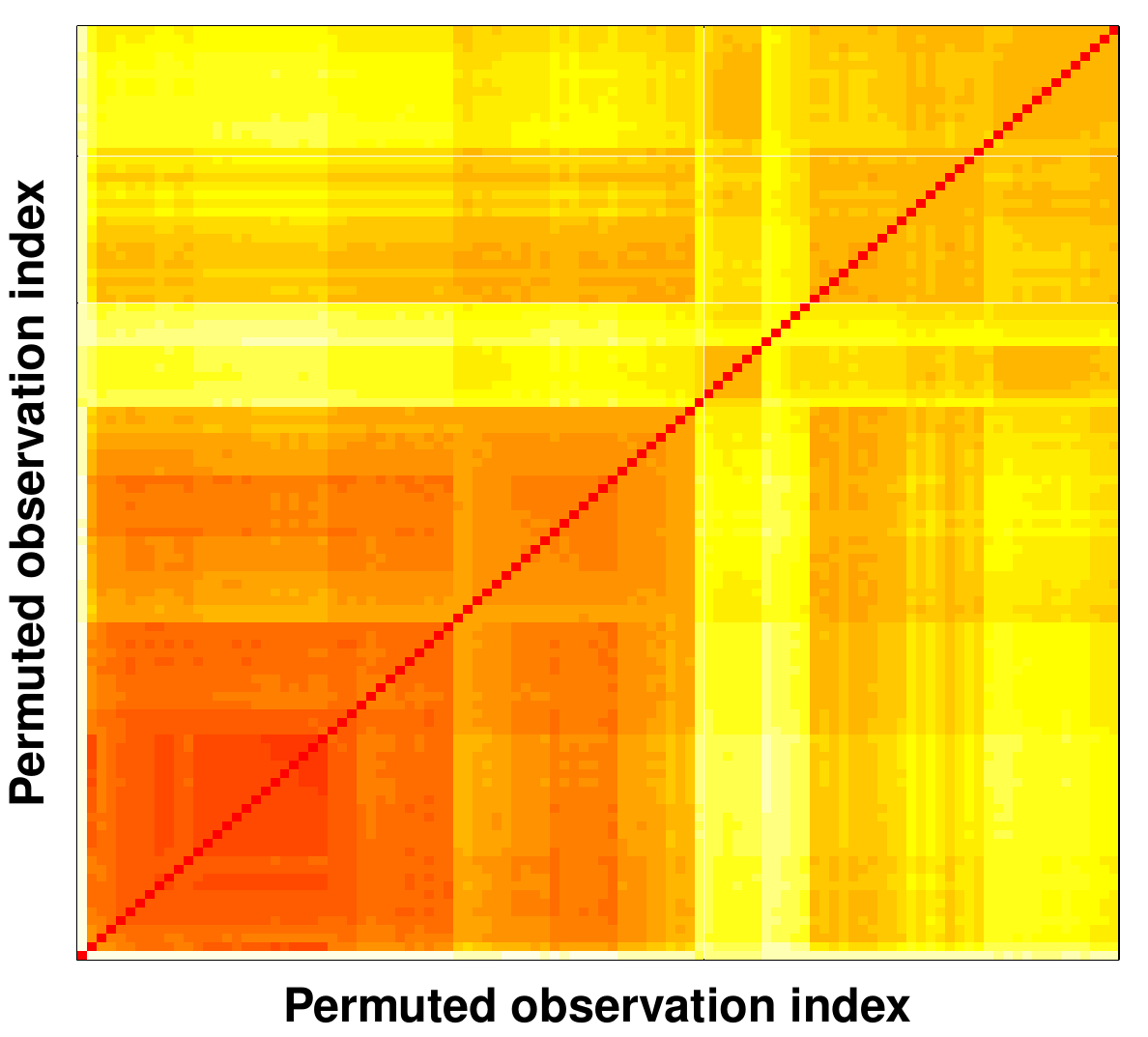}\label{fig:adni_EmE_psm1}}
	\subfigure[$D=5$: $y$-cluster 2]{\includegraphics[width=0.3\textwidth]{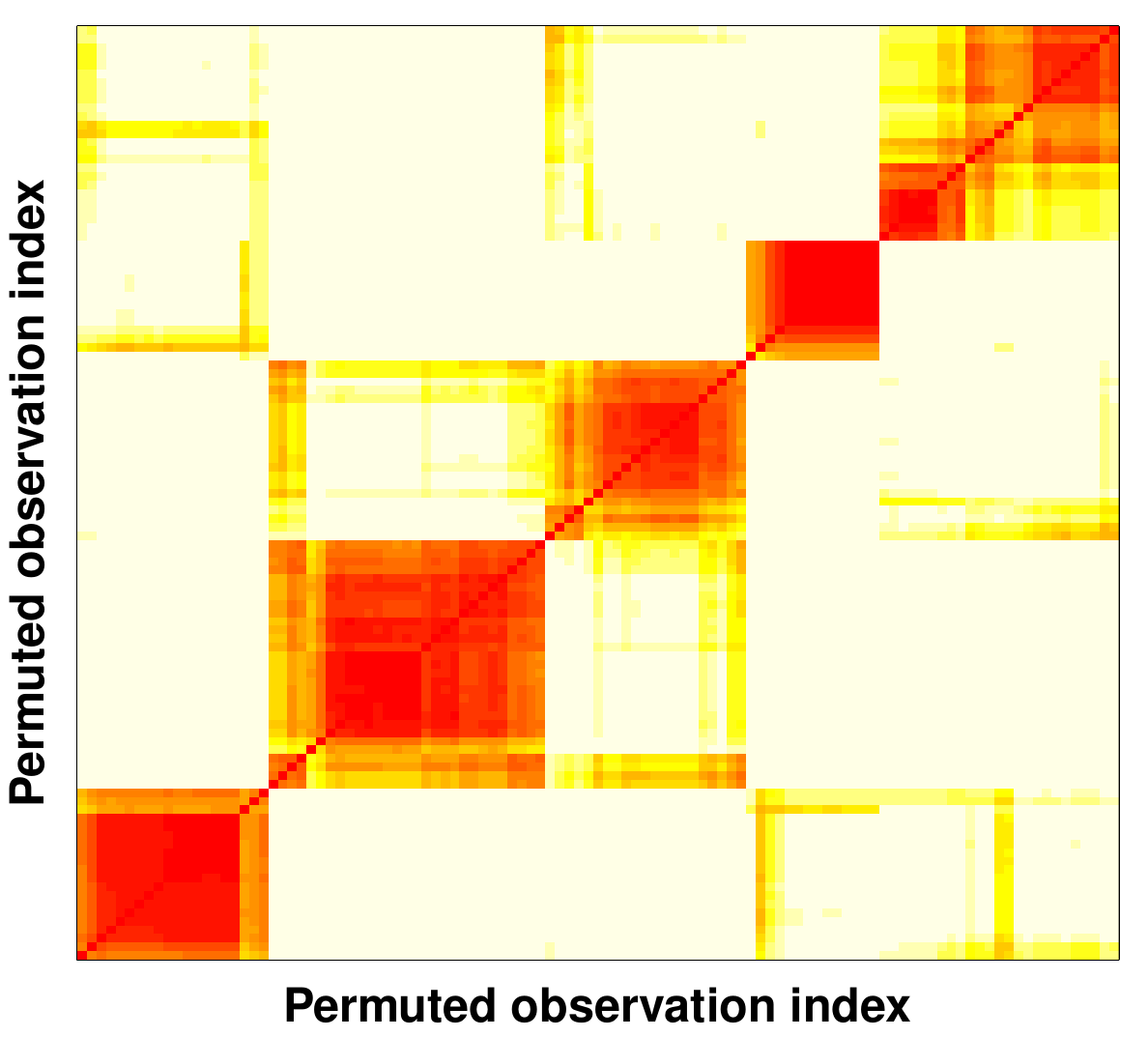}\label{fig:adni_EmE_psm1}}
	\subfigure[$D=10$: $y$-cluster 2]{\includegraphics[width=0.3\textwidth]{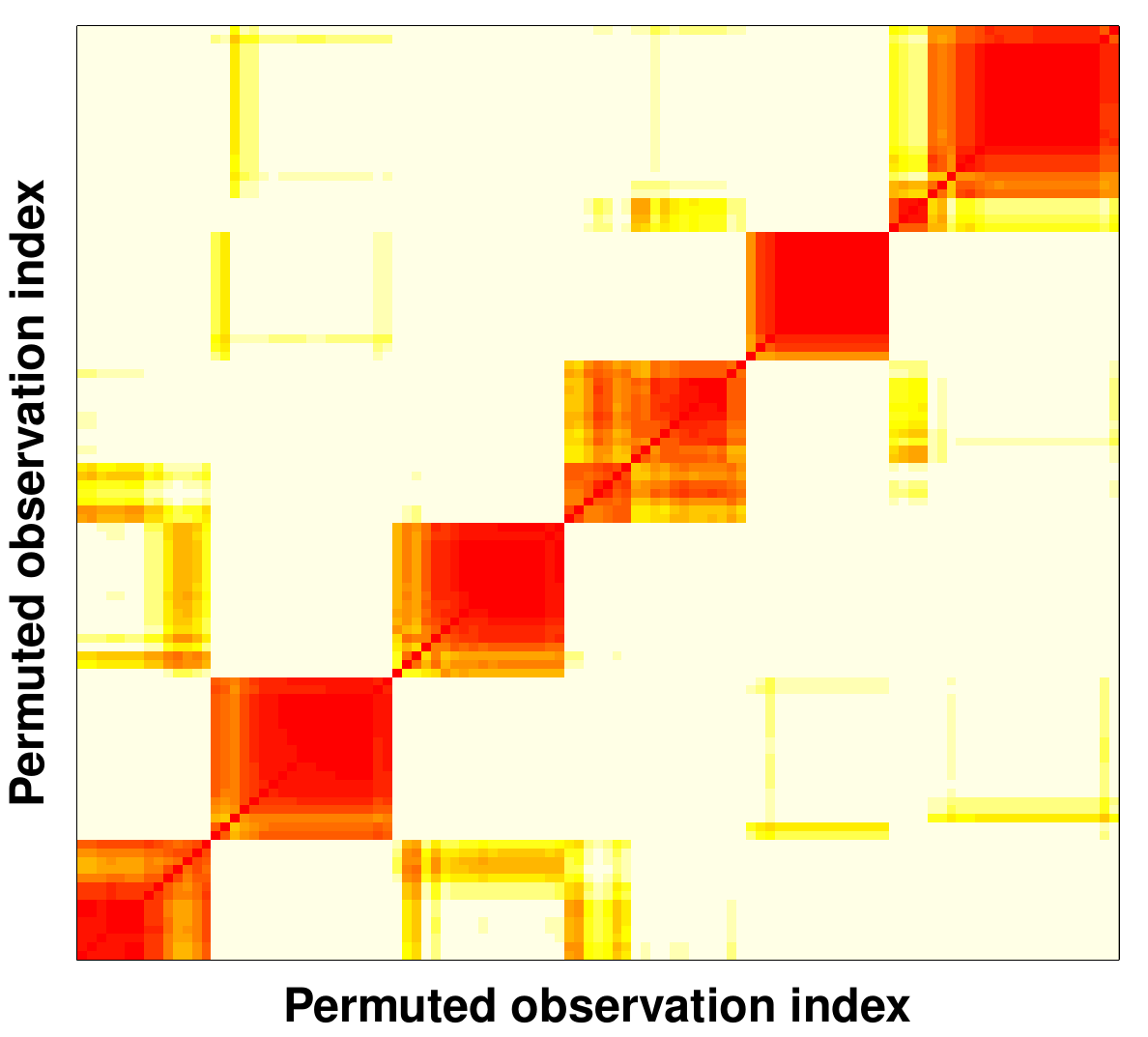}\label{fig:adni_EmE_psm1}}
	\caption{Simulated example. Heat map of the posterior similarity matrix for the $x$-clustering within the two estimated $y$-clusters for the enriched MoE. Rows correspond to $y$-cluster, whilst columns correspond to increasing $D=1,5,10$. To improve visualisation, observations are permuted based on hierarchical clustering.}
	\label{fig:santer_psm}
	\end{center}
\vspace{-5mm}
\end{figure}

\section{Examples}\label{sec:examples}

In the following subsections, we provide further details and insights on the results of the corresponding examples presented in the manuscript.

	\subsection{Simulated Mixture of Damped Cosine Functions}\label{sec:simulated}

In the first example, a data set of $200$ points was generated from a mixture of two  damped cosine functions  by:
\begin{align}
\label{eq:santner_true_pred}
\begin{split}
y_n | x_n \stackrel{ind}{\sim} &p\left(x_{n,1}\right) \Norm \left(\exp\left\{\beta_{1,0} x_{n,1} \right\} \cos \left( \beta_{1,1} \pi x_{n,1}\right), \sigma_1^2 \right)\\
& + (1-p\left(x_{n,1}\right)) \Norm\left(\exp\left\{\beta_{2,0} x_{n,1} \right\} \cos \left( \beta_{2,1} \pi x_{n,1}\right), \sigma_2^2 \right),
\end{split}
\end{align}
with mixture weights, $p\left(x_{n,1}\right)$, equal to
\begin{equation*}
 \frac{ \tau_1 \exp\left\{-\frac{\tau_1}{2}\left(x_{n,1} - \mu_1\right)^2\right\}}{\tau_1 \exp\left\{-\frac{\tau_1}{2}\left(x_{n,1} - \mu_1\right)^2\right\}  +   \tau_2 \exp\left\{-\frac{\tau_2}{2}\left(x_{n,1} - \mu_2\right)^2\right\}}.
\end{equation*}
The damped cosines are parametrised by $\beta_1 = \left(0.1,0.6\right)'$, $\beta_2 = \left(-0.1, 0.4 \right)'$ with $\sigma_1=0.15, \, \sigma_2=0.05$. The mixture model is parametrised by $\tau_1 = \tau_2 = 0.8$, $\mu_1 = 3$ and $\mu_2=5$.  The inputs are independently sampled from a multivariate normal $x_n \sim \Norm(\mu,\Sigma)$,
centred at $\mu=\left(4, \dots, 4\right)$, with standard deviation of $2$ along each dimension, that is $\Sigma_{h,h}=4$. 
The covariance matrix $\Sigma$ assumes with the additional inputs positively correlated among each other, with $\Sigma_{h,l}=3.5$ for $h \neq l$, $h >1$ and $l >1$ , but independent of the first input, with $\Sigma_{1,l}=0$ for $l >1$.

\begin{figure}[!t]
\begin{center}	
\subfigure[VI distance]{\includegraphics[width=0.32\textwidth]{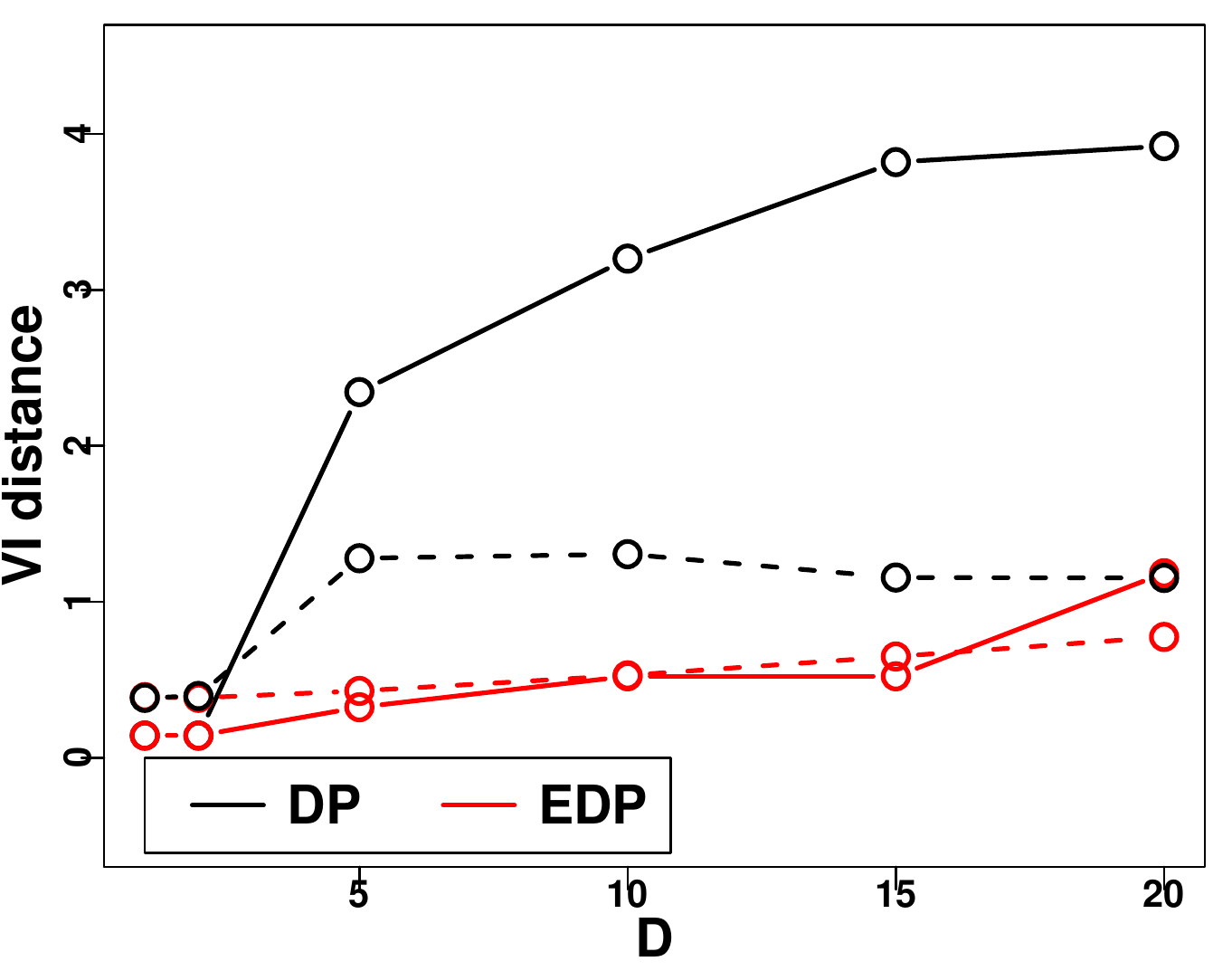}\label{fig:clusters}}
\subfigure[Epoch Time]{\includegraphics[width=0.32\textwidth]{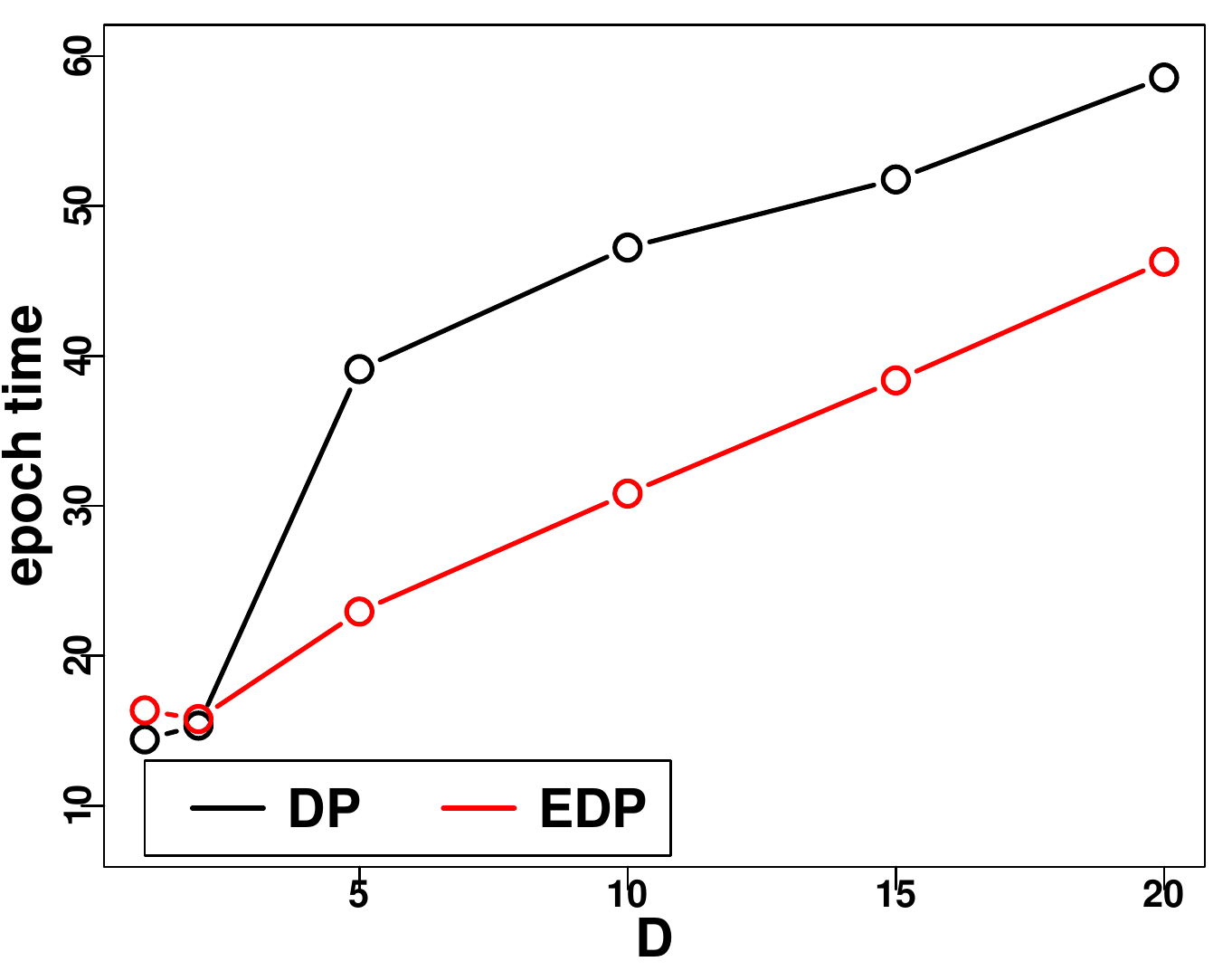}\label{fig:epoch}}
\subfigure[Coverage of CI]{\includegraphics[width=0.32\textwidth]{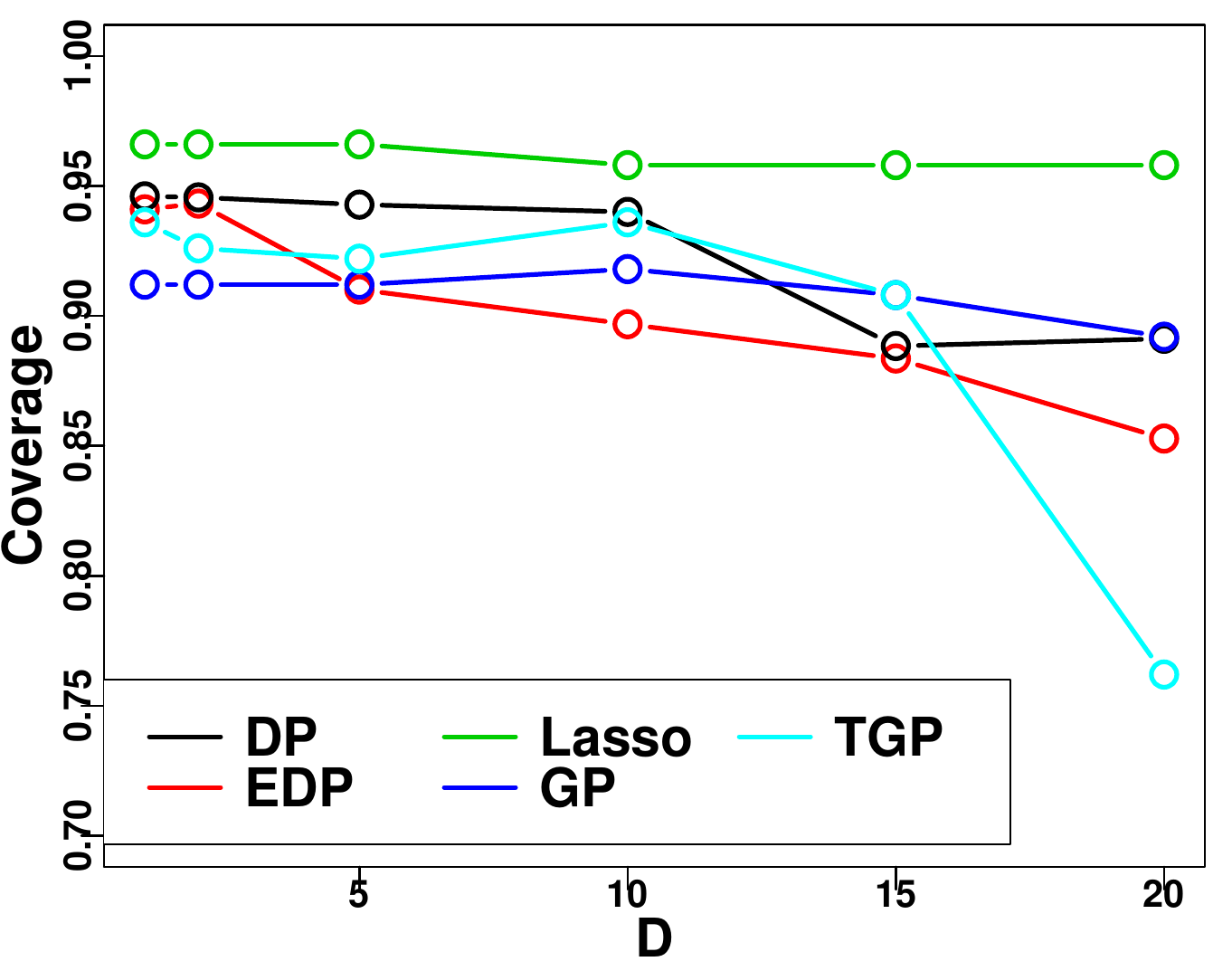}\label{fig:cicoverage}}
\caption{Simulated example. Comparison of the EDP MoE with the DP MoE, Lasso, GP, and TGP in terms of the VI distance between the true and estimated clustering (with dashed lines for the size of the credible ball), average epoch time (in seconds), and the coverage of the 95\% credible intervals (CI). }
\label{fig:simcomparison}
	\end{center}
	\vspace{-5mm}
\end{figure}

For both the DP and EDP mixtures of GP experts, we employ the same prior choices, based on identified reasonable ranges for the parameters. For the ARD squared exponential kernels of the GPs, we utilise a Gamma($3, 1$) prior on the first input dimension length-scale, Gamma($10, 1/2$) prior on the other input dimension length-scales  
and a Gamma($2, 1.5$) prior on the magnitude.  
The constant means $\beta_0$ of the GPs have a $\Norm(0,0.5^2)$ prior. The variance $\sigma^2_y$ has a log-\Norm($\log\left(0.01\right),0.5^2$). 
For the DP, the mass parameter has hyper-parameters $(u_a =1,v_a=1)$,  and for the EDP, the mass parameters have hyper-parameters $(u_\theta =1,v_\theta=1)$ and $(u_\psi =1,v_\psi=1)$.  A Gaussian input model is used with hyperparameters of the conjugate normal-inverse gamma set to $u_{0,d}=\bar{x}_d$, $c_d=1/4$, $b_{x,d}=1$, and $a_{x,d}=2$.

Posterior inference for both models is performed with $5000$ total iterations and a burn-in of $1000$. Average epoch times (in seconds) after burn-in are reported in Figure \ref{fig:epoch}. When fewer experts are identified through the nested clustering of the EDP (e.g. $D>2$ in our example), average epoch time is reduced for the EDP compared with the DP. Each run was performed independently and in parallel using the high performance computing resources provided by \textit{commented for blind review}.

The VI distance between the true and estimated ($y$-level) clustering is depicted in Figure \ref{fig:clusters}, with dashed lines representing the size of the 95\% VI credible balls. For the DP, the distance increases greatly with $D$, and the true clustering is far from the credible ball. The behaviour of the $y$-level clustering of the EDP is more robust to increasing $D$, while the $x$-level clustering requires an increasing number of clusters. Figure \ref{fig:santer_psm} depicts the heat map of the posterior similarity matrix for the $x$-clustering within the two estimated $y$-clusters, and Table \ref{tab:santer_kj} reports the number of $x$-clusters in the VI estimated $x$-clustering within the two estimated $y$-clusters. 

\begin{table}[t]
	\caption{Simulated example. The number of clusters in the VI estimated $x$-clustering within the two estimated $y$-clusters for the enriched model, as $D$ increases. }
	\label{tab:santer_kj}
\begin{center}
	\begin{tabular}{c|ccccc}
		$y$-cluster &  $D=1$ &  $D=5$ & $D=10$ & $D=15$ & $D=20$\\ \hline
		1 & 1 & 5 & 6 & 7 & 7\\ 
		2 &  1 & 6& 7 & 8 & 7\\ 
	\end{tabular}
	\end{center}
\vskip -0.1in
\end{table}

\begin{figure}
\begin{center}
	\subfigure[True]{\includegraphics[width=0.32\textwidth]{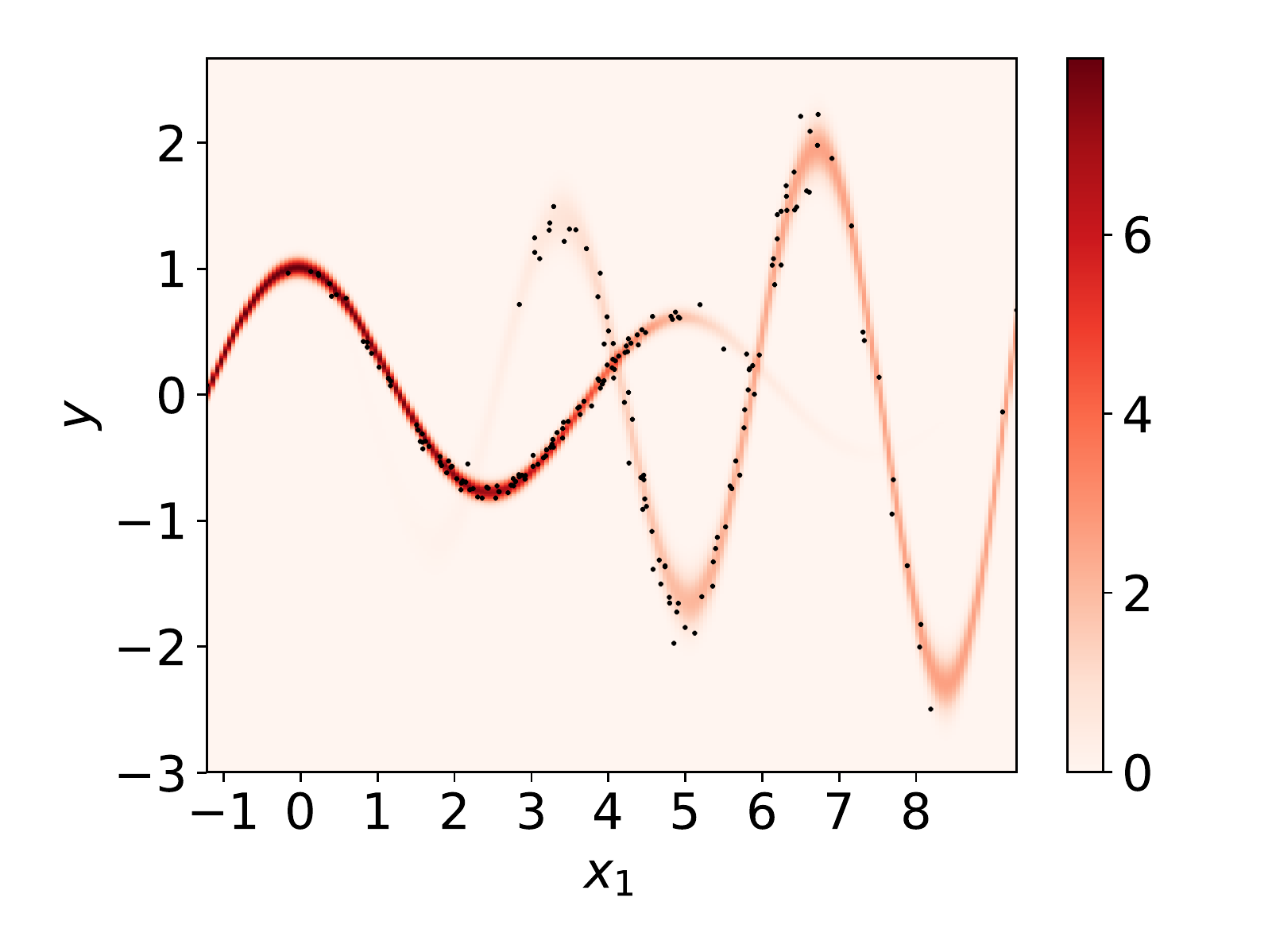}\label{fig:santner_truth}} \\ \vspace{-4mm}
	\subfigure[DP conditional, $D=1$]{\includegraphics[width=0.32\textwidth]{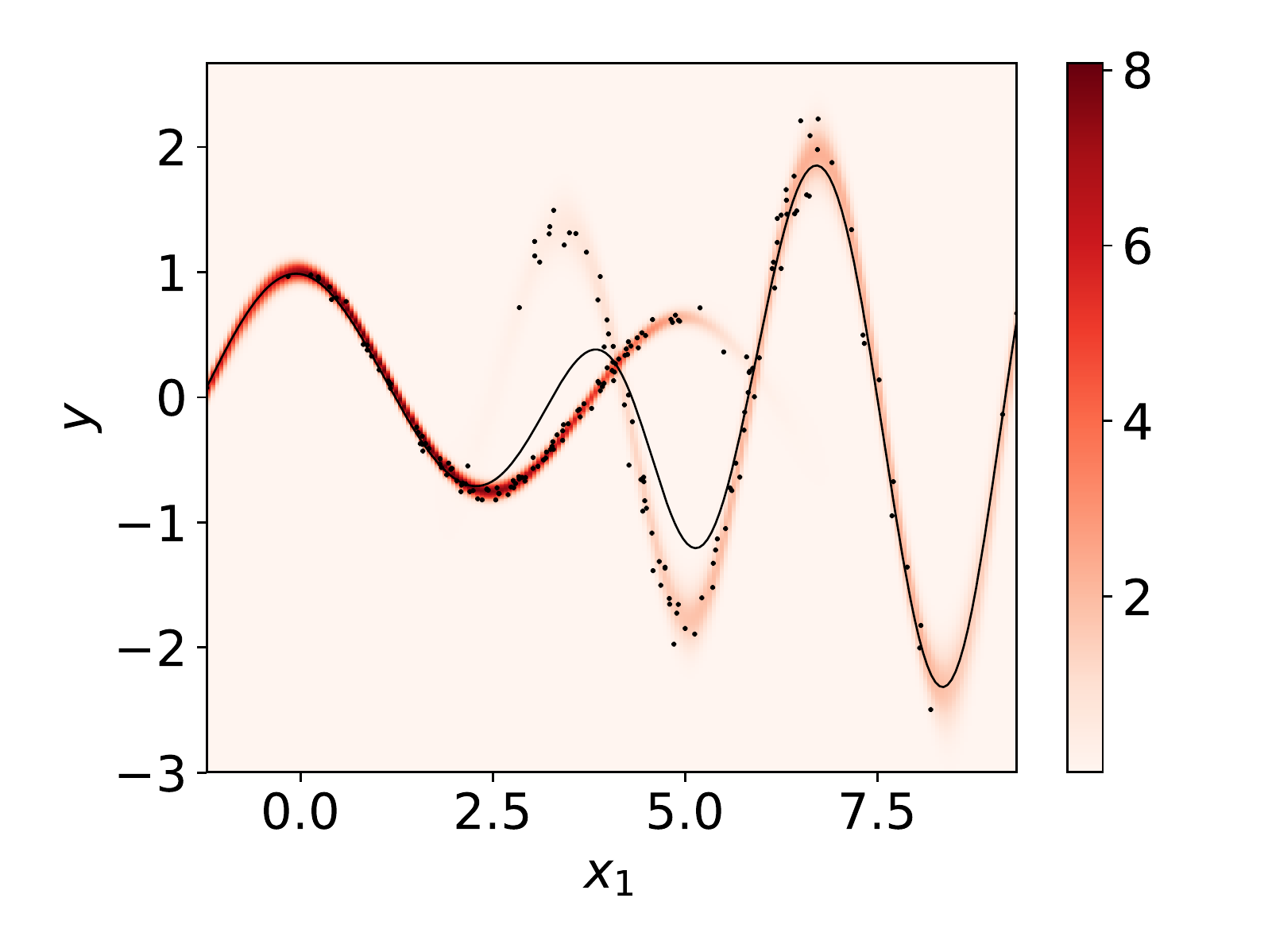}\label{fig:santner_JmE1_conditional}}
	\subfigure[DP conditional, $D=5$]{\includegraphics[width=0.32\textwidth]{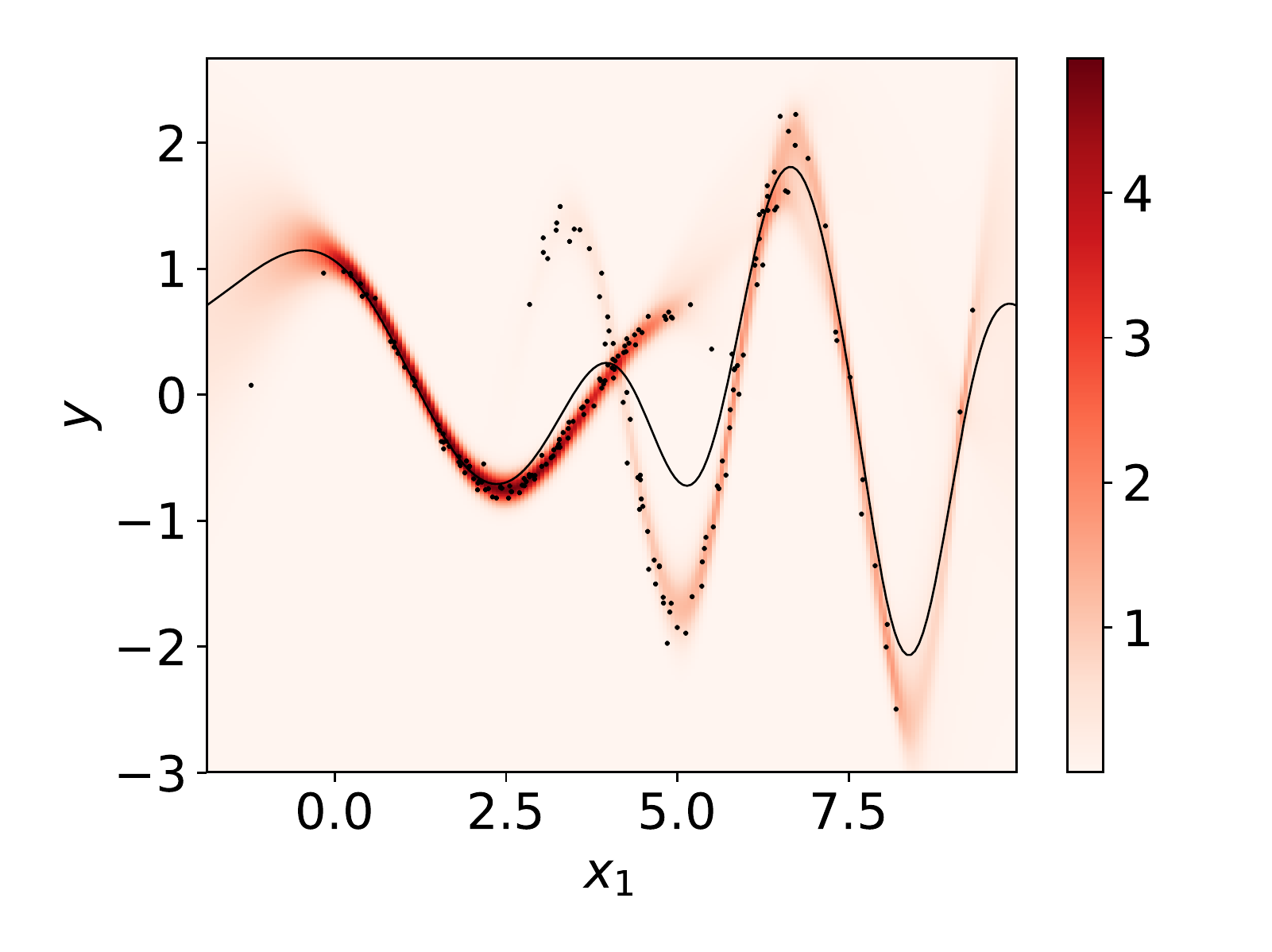}\label{fig:santner_JmE2_conditional}}
	\subfigure[DP conditional, $D=10$]{\includegraphics[width=0.32\textwidth]{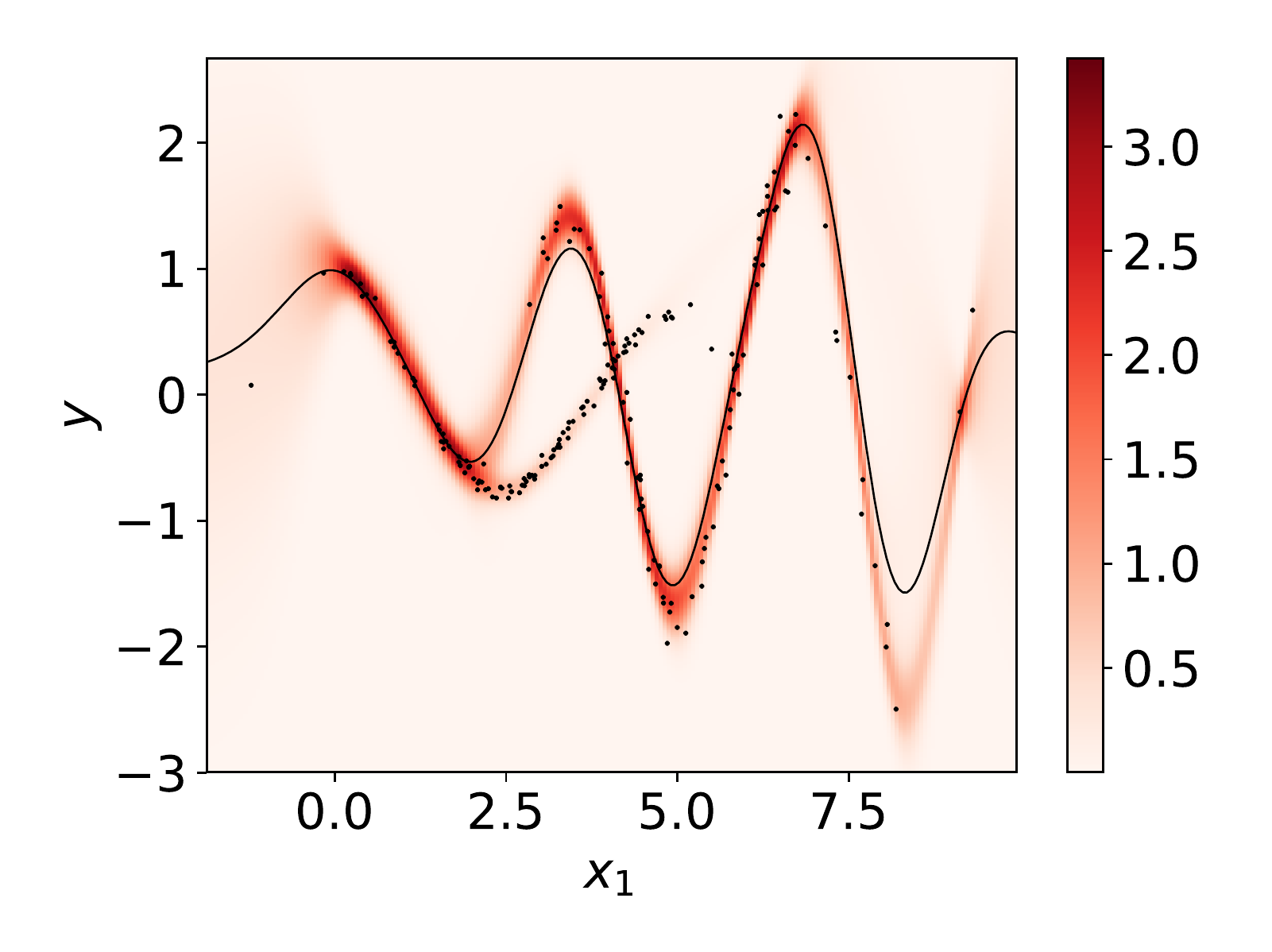}\label{fig:santner_JmE5_conditional}} \\  \vspace{-4mm}
	\subfigure[DP marginal, $D=1$]{\includegraphics[width=0.32\textwidth]{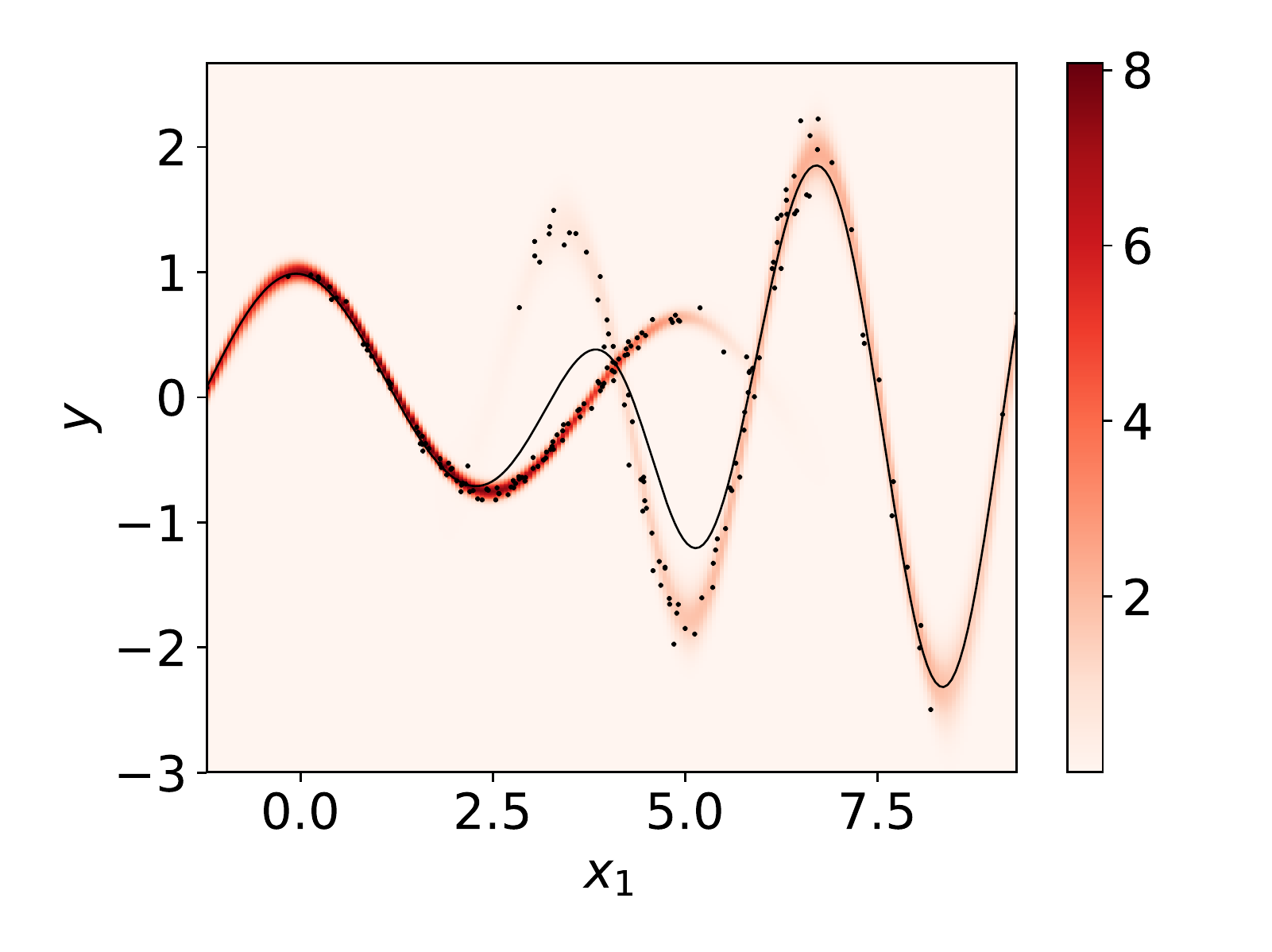}\label{fig:santner_JmE1_marginal}}
	\subfigure[DP marginal, $D=5$]{\includegraphics[width=0.32\textwidth]{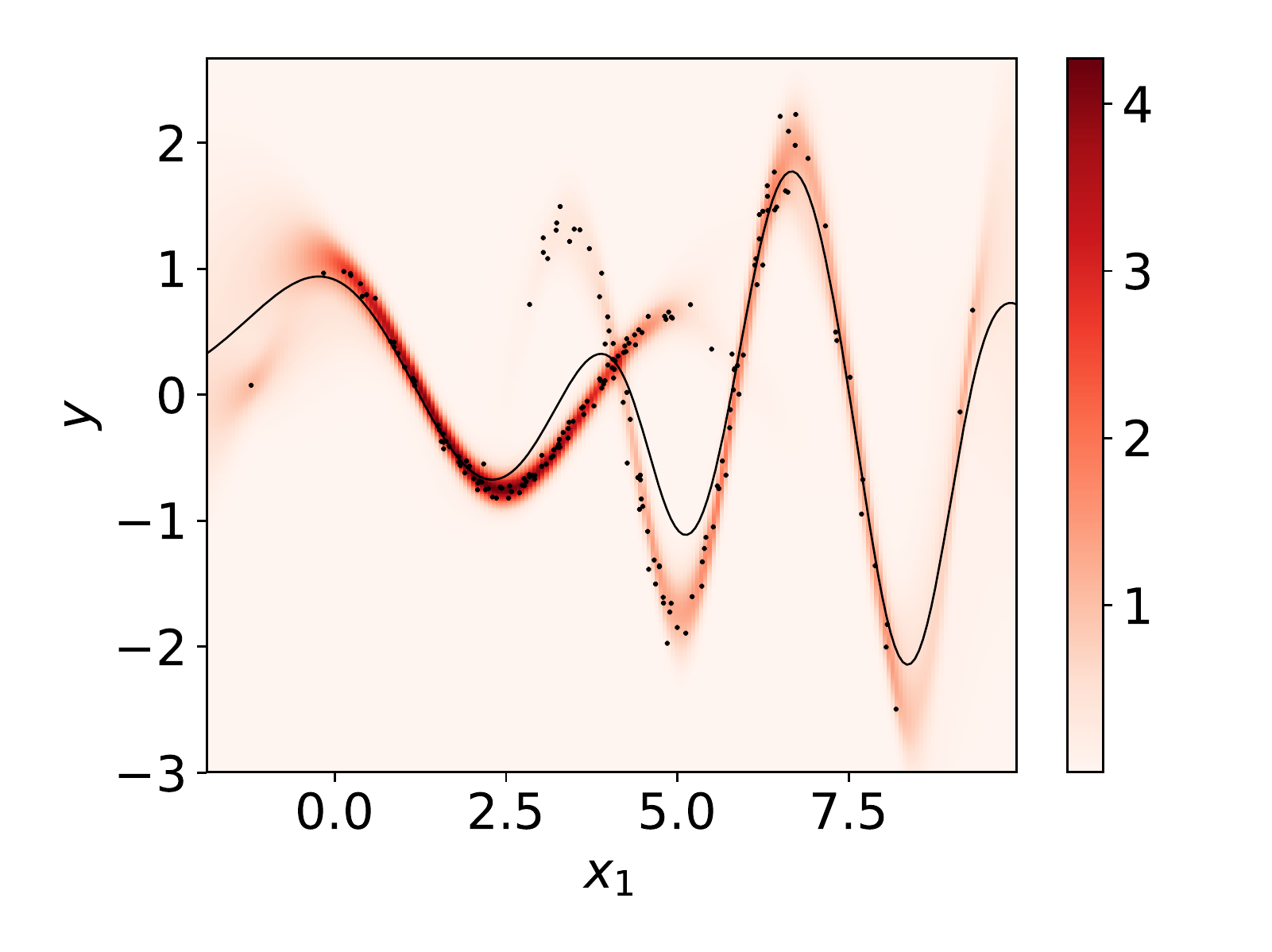}\label{fig:santner_JmE2_marginal}}
	\subfigure[DP marginal, $D=10$]{\includegraphics[width=0.32\textwidth]{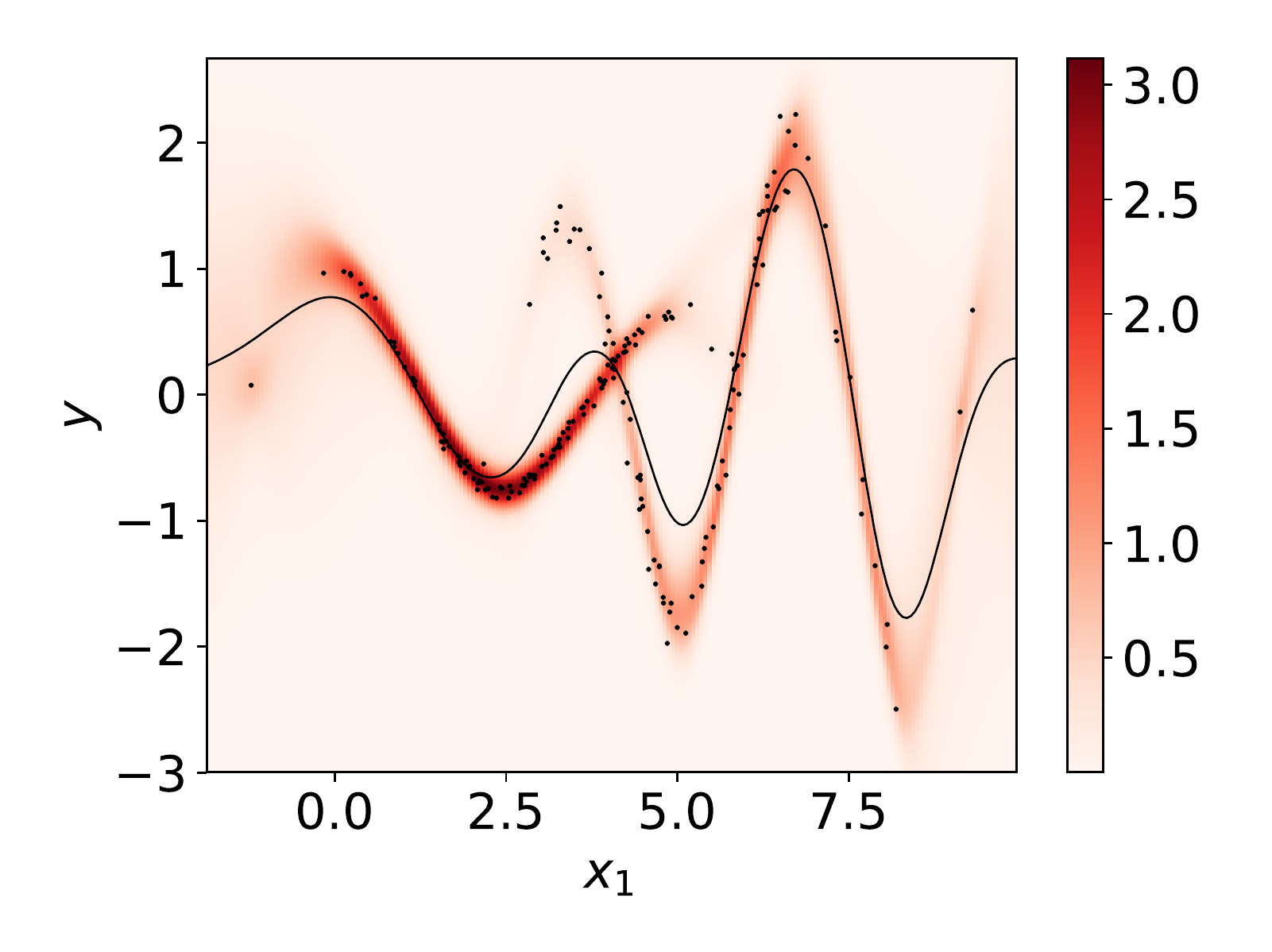}\label{fig:santner_JmE5_marginal}} \\  \vspace{-4mm}
	\subfigure[EDP conditional, $D=1$]{\includegraphics[width=0.32\textwidth]{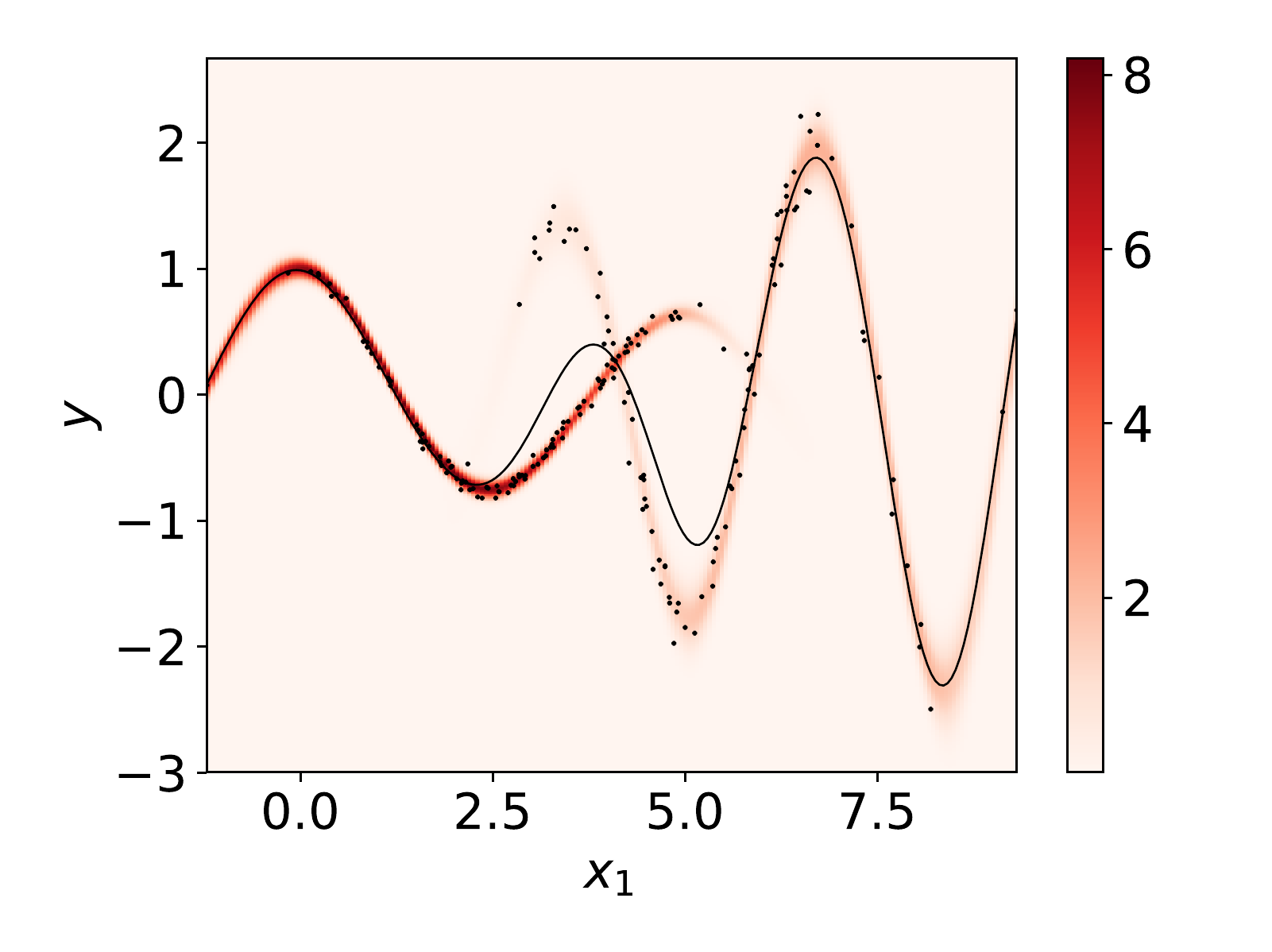}\label{fig:santner_EmE1_conditional}}
	\subfigure[EDP conditional, $D=5$]{\includegraphics[width=0.32\textwidth]{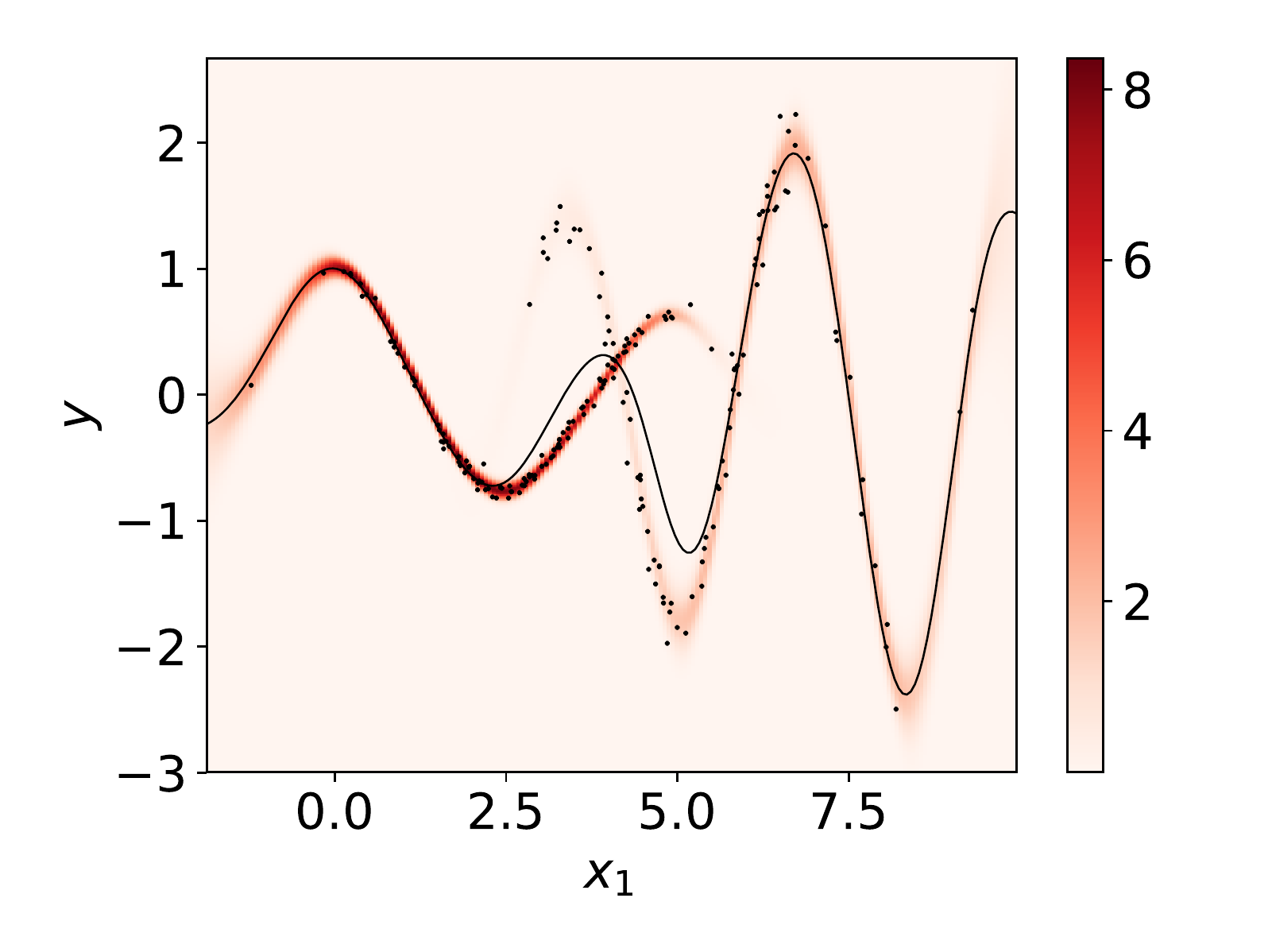}\label{fig:santner_EmE2_conditional}}
	\subfigure[EDP conditional, $D=10$]{\includegraphics[width=0.32\textwidth]{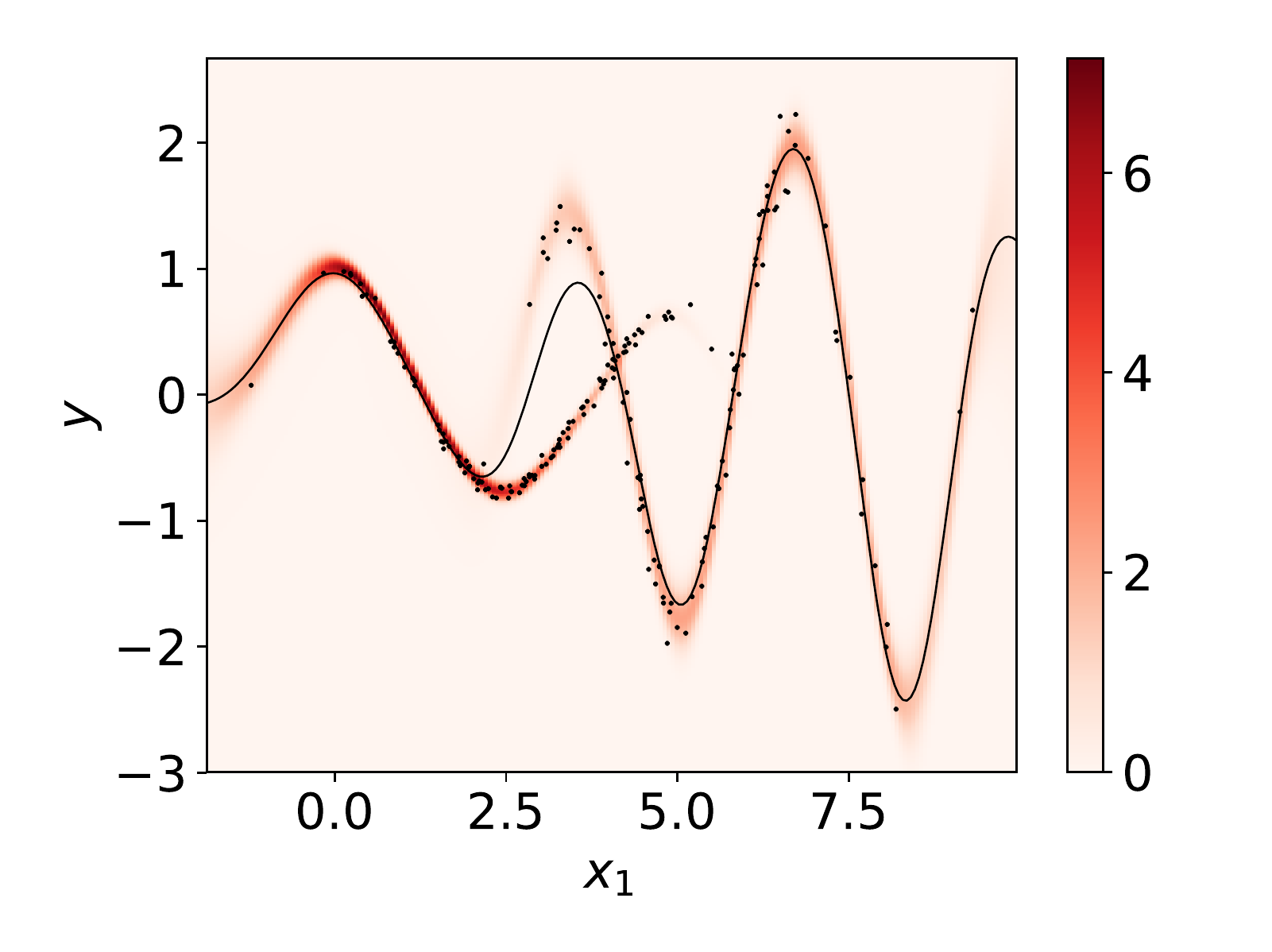}\label{fig:santner_EmE5_conditional}}
	 \\  \vspace{-4mm}
	\subfigure[EDP marginal, $D=1$]{\includegraphics[width=0.32\textwidth]{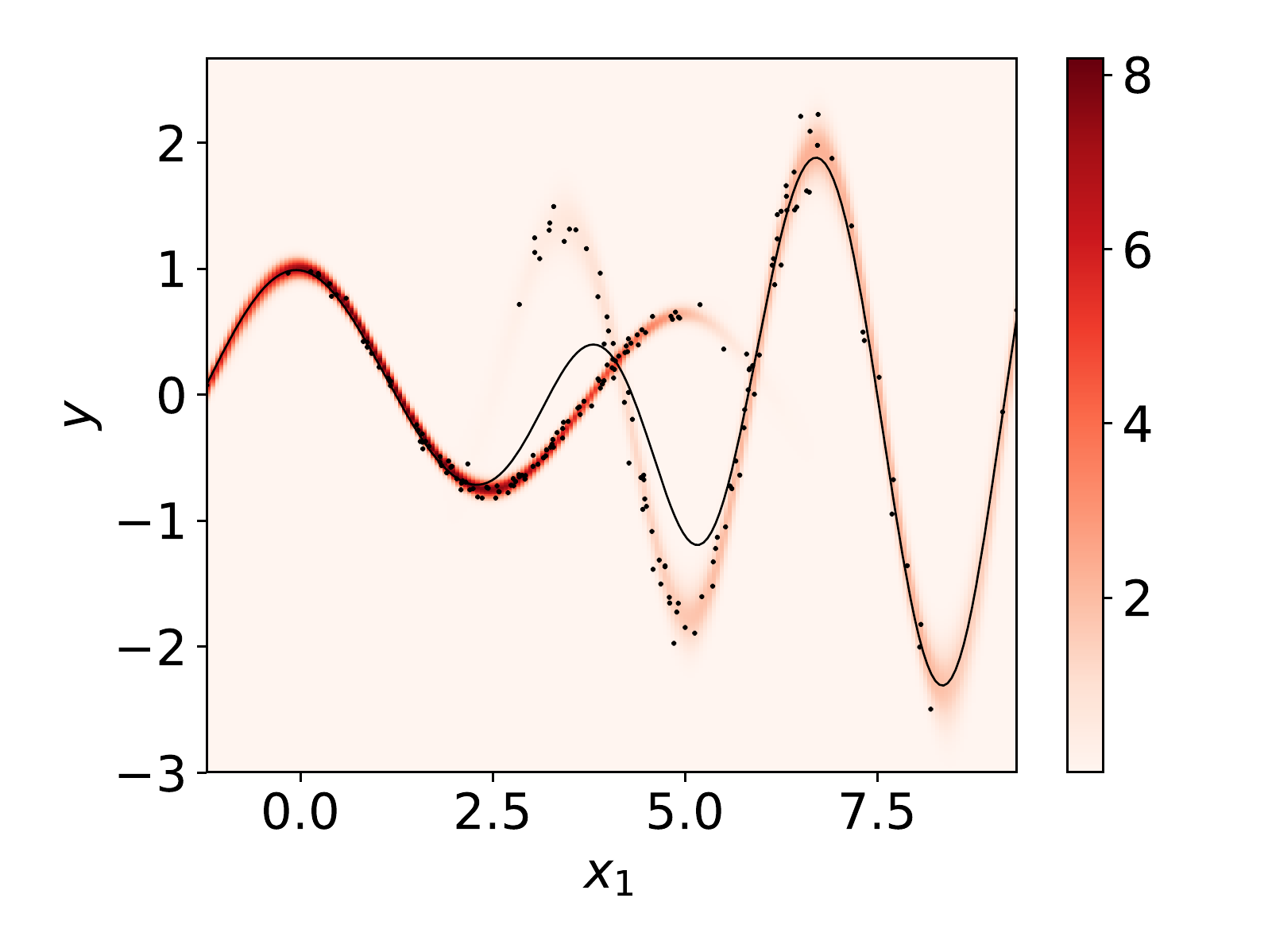}\label{fig:santner_EmE1_marginal}}
	\subfigure[EDP marginal, $D=5$]{\includegraphics[width=0.32\textwidth]{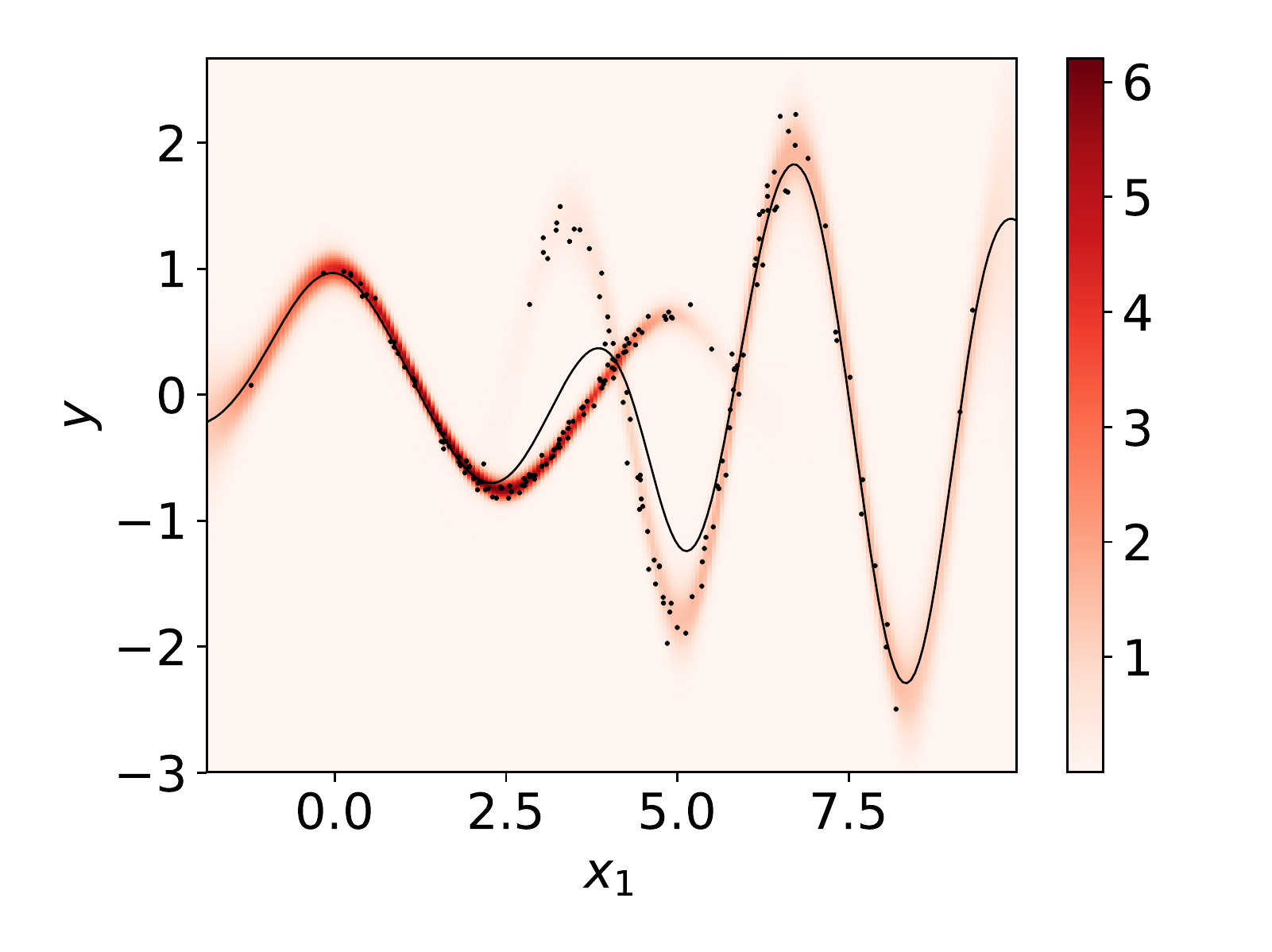}\label{fig:santner_EmE2_marginal}}
	\subfigure[EDP marginal, $D=10$]{\includegraphics[width=0.32\textwidth]{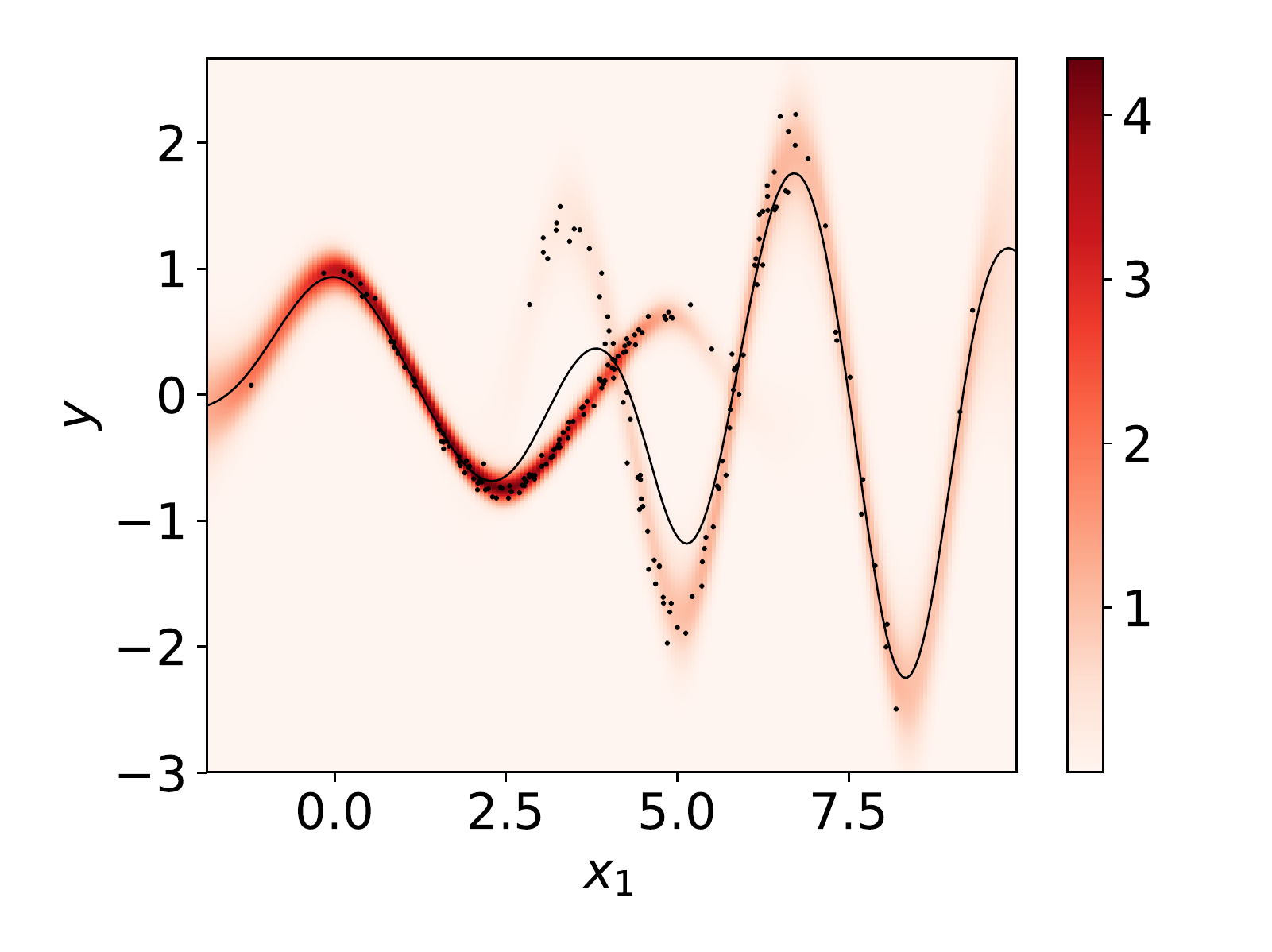}\label{fig:santner_EmE5_marginal}}
	\caption{Simulated example. Predictive density plots for the DP and EDP for a grid of $x_{*, 1}$ values, with additional inputs conditioned on their sample means (second and fourth rows) or marginalised (third and fifth rows), with increasing $D=1,5,10$ (columns).}
	\label{fig:santner_predictive}
	\end{center}
\vskip -0.2in
\end{figure}

We plot the estimates for the predictive response density and mean against the first input over a dense grid. These are presented in Figure~\ref{fig:santner_predictive}, for different choices of $D$. In the second and fourth rows the additional inputs are fixed to their sample means (approximately $4$) for the DP and EDP models, respectively. Further, in the third and fifth rows, the additional inputs are marginalised. 

\begin{figure}
\begin{center}
	\subfigure[DP, $D=1$]{\includegraphics[width=0.32\textwidth]{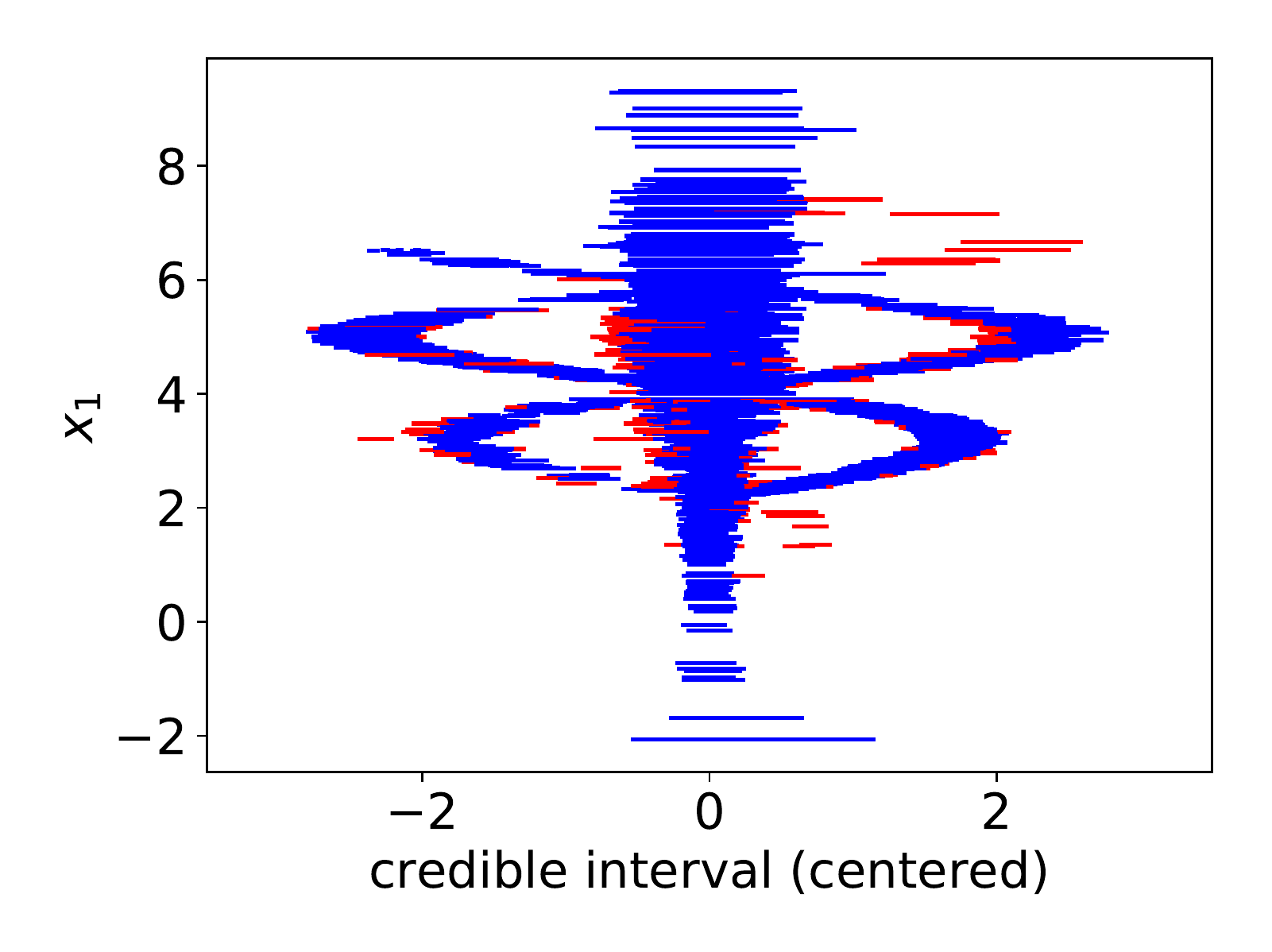}\label{fig:JmE1_CI}}
	\subfigure[DP, $D=5$]{\includegraphics[width=0.32\textwidth]{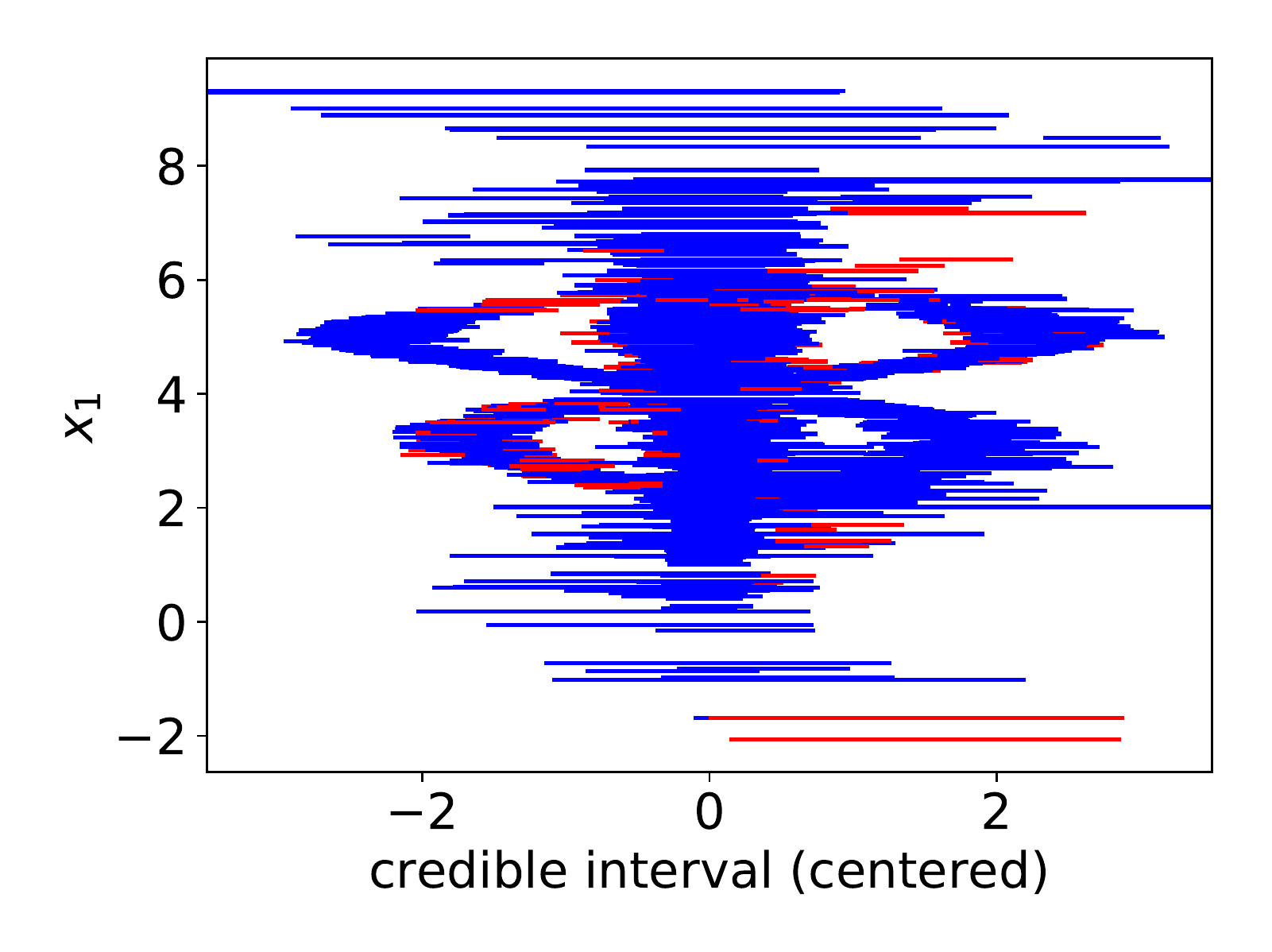}\label{fig:JmE2_CI}}
	\subfigure[DP, $D=10$]{\includegraphics[width=0.32\textwidth]{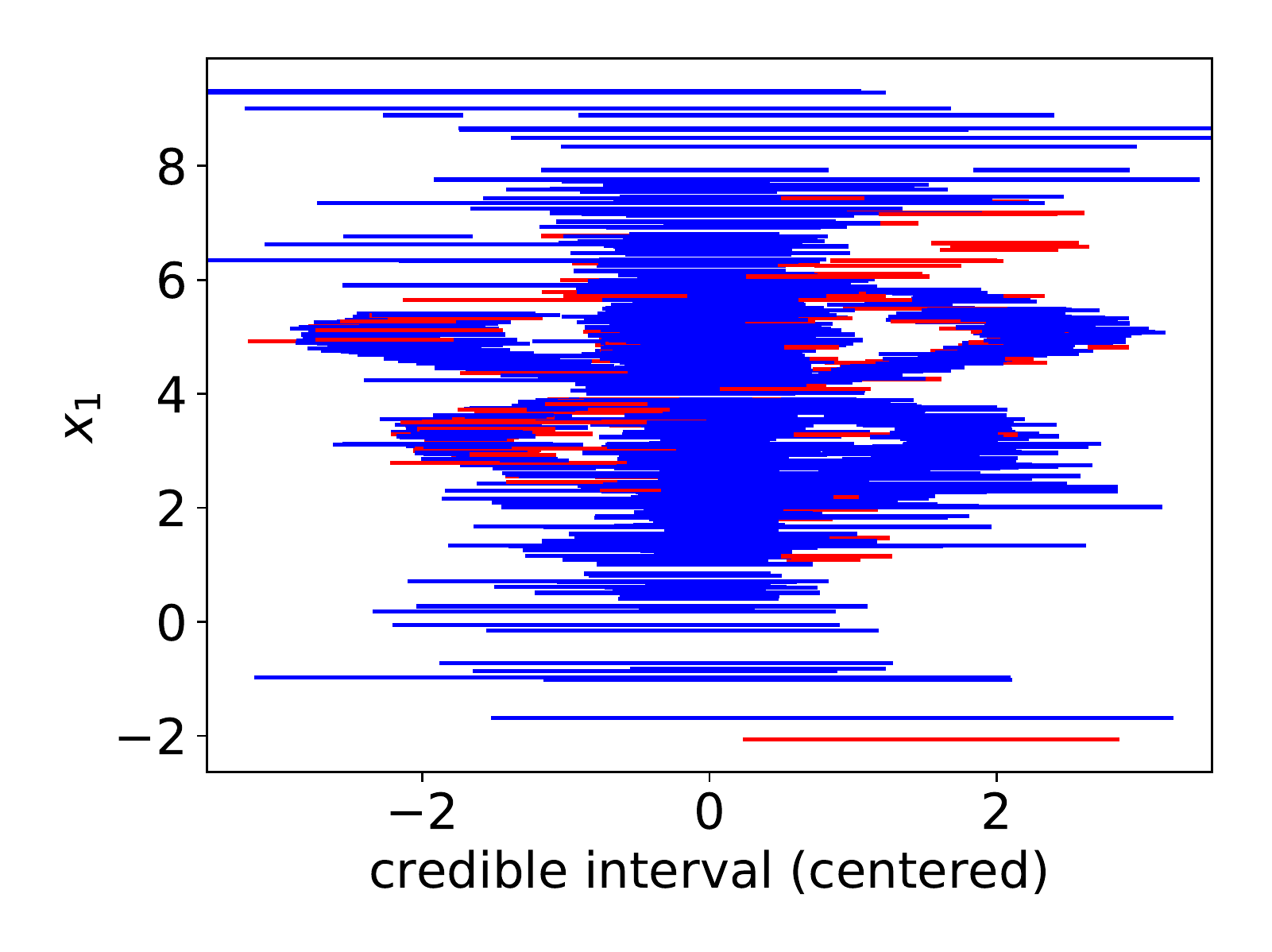}\label{fig:JmE5_CI}} \\
	\subfigure[EDP, $D=1$]{\includegraphics[width=0.32\textwidth]{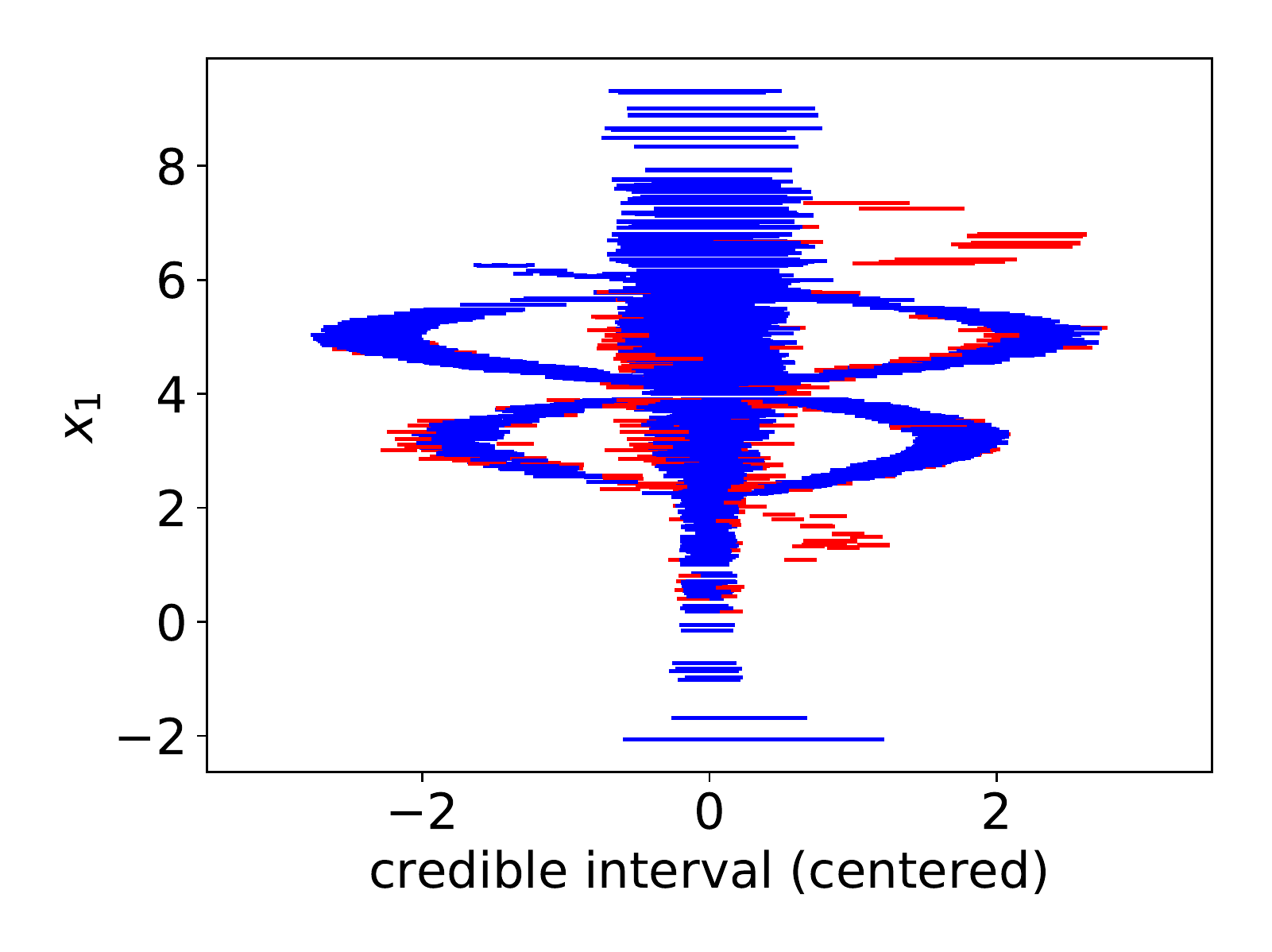}\label{fig:EmE1_CI}}
	\subfigure[EDP, $D=5$]{\includegraphics[width=0.32\textwidth]{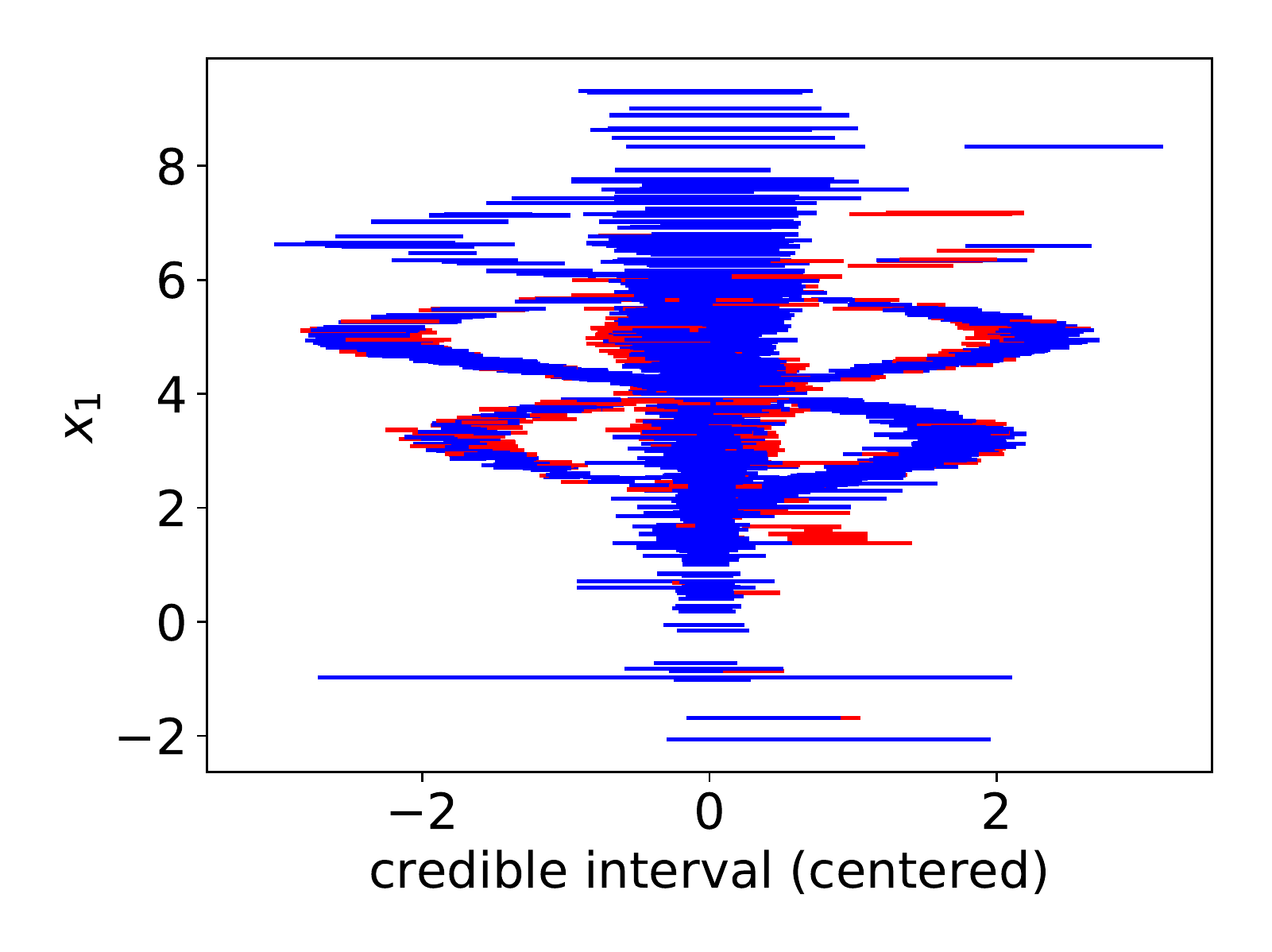}\label{fig:EmE2_CI}}
	\subfigure[EDP, $D=10$]{\includegraphics[width=0.32\textwidth]{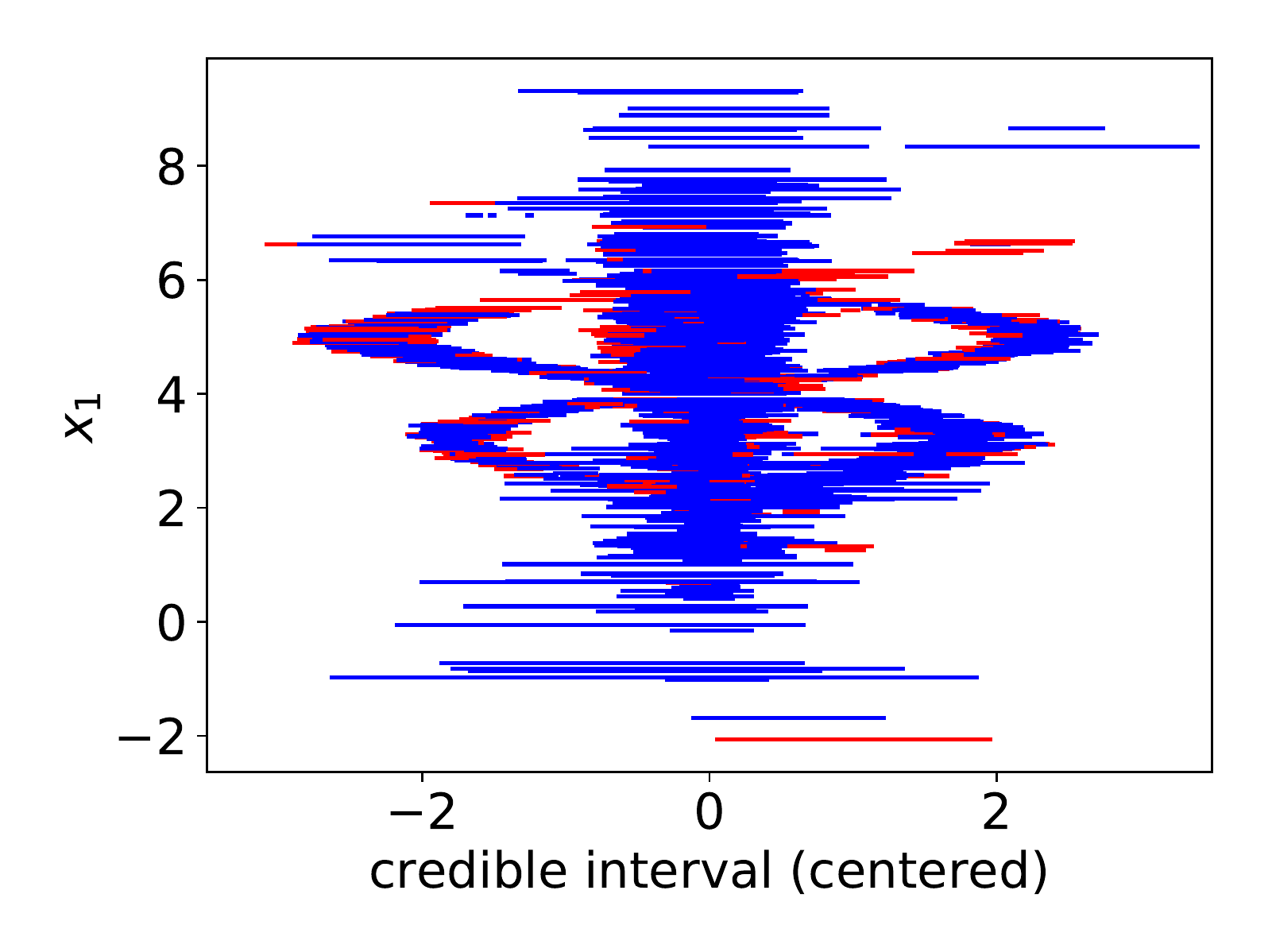}\label{fig:EmE5_CI}}
	\caption{Simulated example. Coverage for the DP and EDP mixture of experts with increasing $D=1,5,10$. Each horizontal line depicts the 95\% credible interval based on highest posterior density and is blue if the sampled truth lies inside, and red otherwise. The percentage of samples lying inside the interval is the empirical coverage.}
	\label{fig:santner_coverage}
	\end{center}
\vskip -0.2in
\end{figure}

Finally, coverage plots are presented in Figure~\ref{fig:santner_coverage}. 
Centered around the true values (sampled from the data generating distribution of equation~\eqref{eq:santner_true_pred}), these plots show the 95\% highest posterior density credible intervals for randomly sampled inputs (in some cases this may be a union of intervals). When the sample of the truth lies within our credible interval the line is blue, otherwise it is red. The increasing uncertainty  of the DP for increasing $D$ is clearly visible from Figure~\ref{fig:santner_coverage}, while the EDP retains smaller credible intervals, with similar coverage.  Figure \ref{fig:cicoverage} summarises the coverage across the competing models. All GP-based methods show a decrease in coverage with increasing $D$. In order to cope with the additional noisy inputs, length-scale priors with heavier tails may be required to effectively identify the relevant inputs.

\subsection{Alzheimer's Challenge}\label{sec:ADNI}

Training data for the challenge is extracted from the Alzheimer's Disease Neuro-Initiative (ADNI) database\footnote{
The ADNI was launched in 2003 by the National Institute on Aging (NIA), the National Institute of Biomedical Imaging and Bioengineering (NIBIB), the Food and Drug Administration (FDA), private pharmaceutical companies and non-profit organizations, as a \$ 60 million, 5-year public- private partnership. The primary goal of ADNI has been to test whether serial magnetic resonance imaging (MRI), positron emission tomography (PET), other biological markers, and clinical and neuropsychological assessment can be combined to measure the progression of mild cognitive impairment (MCI) and early Alzheimer's disease (AD). Determination of sensitive and specific markers of very early AD progression is intended to aid researchers and clinicians to develop new treatments and monitor their effectiveness, as well as lessen the time and cost of clinical trials.
The Principal Investigator of this initiative is Michael W. Weiner, MD, VA Medical Center and University of California-San Francisco. ADNI is the result of efforts of many co-investigators from a broad range of academic institutions and private corporations, and subjects have been recruited from over 50 sites across the U.S. and Canada. The initial goal of ADNI was to recruit 800 adults, ages 55 to 90, to participate in the research, approximately 200 cognitively normal older individuals to be followed for 3 years, 400 people with MCI to be followed for 3 years and 200 people with early AD to be followed for 2 years. For up-to-date information, see \adniinfo.}. This data set consists of six inputs: age (in fraction of years), gender, the baseline mini mental-state exam (MMSE) score, the number of years an individual has spent in education, APOE genotype (recoded to reflect the number of copies of the type 4 allele), and the clinical diagnosis assessed at the baseline. The output is the MMSE score taken at a $24$ month follow-up visit, and the task is to predict the cognitive decline in a patient over this period.

We again employ the same prior choices for both mixtures of GP experts models, based on identified reasonable ranges for the parameters. We consider an ARD squared exponential kernel for the GP with Gamma($a_{l,d},b_{l,d}$) priors on the length-scales with $a_{l}=(3,2,3,5,3,2)$, and $b_{l}=(3/20,5,1,1,5,4)$, in order of the inputs listed above. Additionally, we specify a Gamma($a_m,b_m$) prior on the magnitude with $a_m=2$ and $b_m=1$. 
These parameters were selected to reflect our prior knowledge on the relationship between follow-up MMSE and the inputs and based on the range of the inputs. The GP is assumed to have a prior constant mean with a $\Norm(20,7.5^2)$ prior. The variance $\sigma^2_y$ has a Gamma($a_y,b_y)$ prior with $a_y=1.5$ and $b_y=0.5$.

For the DP, the mass parameter has hyper-parameters $(u_a =1,v_a=1)$,  and for the EDP, the mass parameters have hyper-parameters $(u_\theta =1,v_\theta=1)$ and $(u_\psi =1,v_\psi=1)$. The parametric local model for $x_{n}$ is the product of a normal density for age, a categorical density for gender, and four binomial densities for baseline MMSE, education, APOE4, and diagnosis. The input hyperparameters are $u_0=72$, $c=2$, $b_x=10$, and $a_x=2$ for age; $\gamma_2=(1,1)$ for gender; $\gamma_3=(5,1)$ for MMSE; $\gamma_4=(3,2)$ for education; $\gamma_5=(1,3)$ for APOE4; $\gamma_6=(1,1)$ for diagnosis. Posterior inference for both models is performed with $5000$ total iterations and a burn-in of $1000$.

\begin{figure}[!h]
\begin{center}
	\subfigure[DP]{\includegraphics[width=0.35\textwidth]{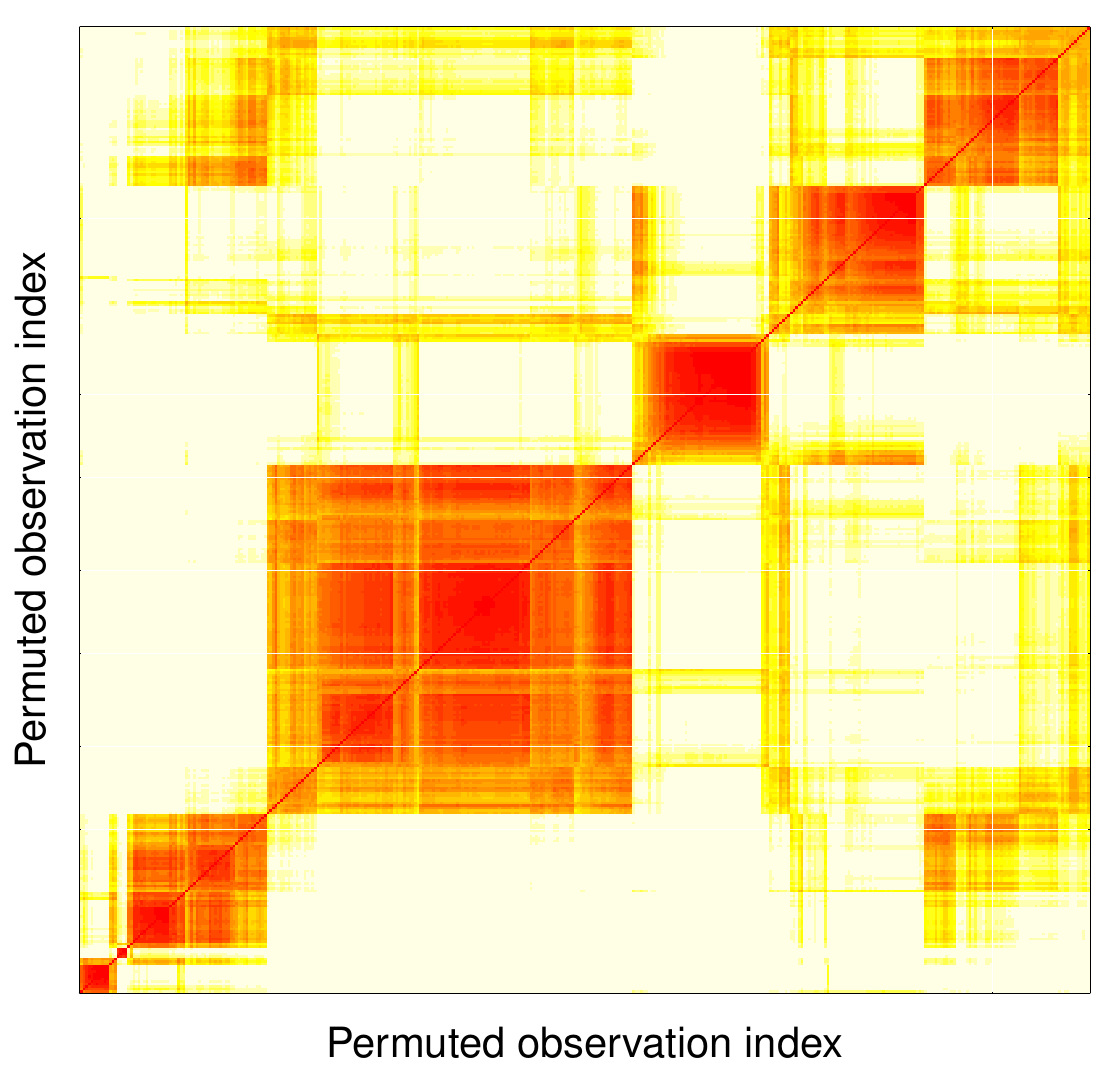}\label{fig:adni_JmE_psm}} \hspace{5mm}
	\subfigure[EDP]{\includegraphics[width=0.35\textwidth]{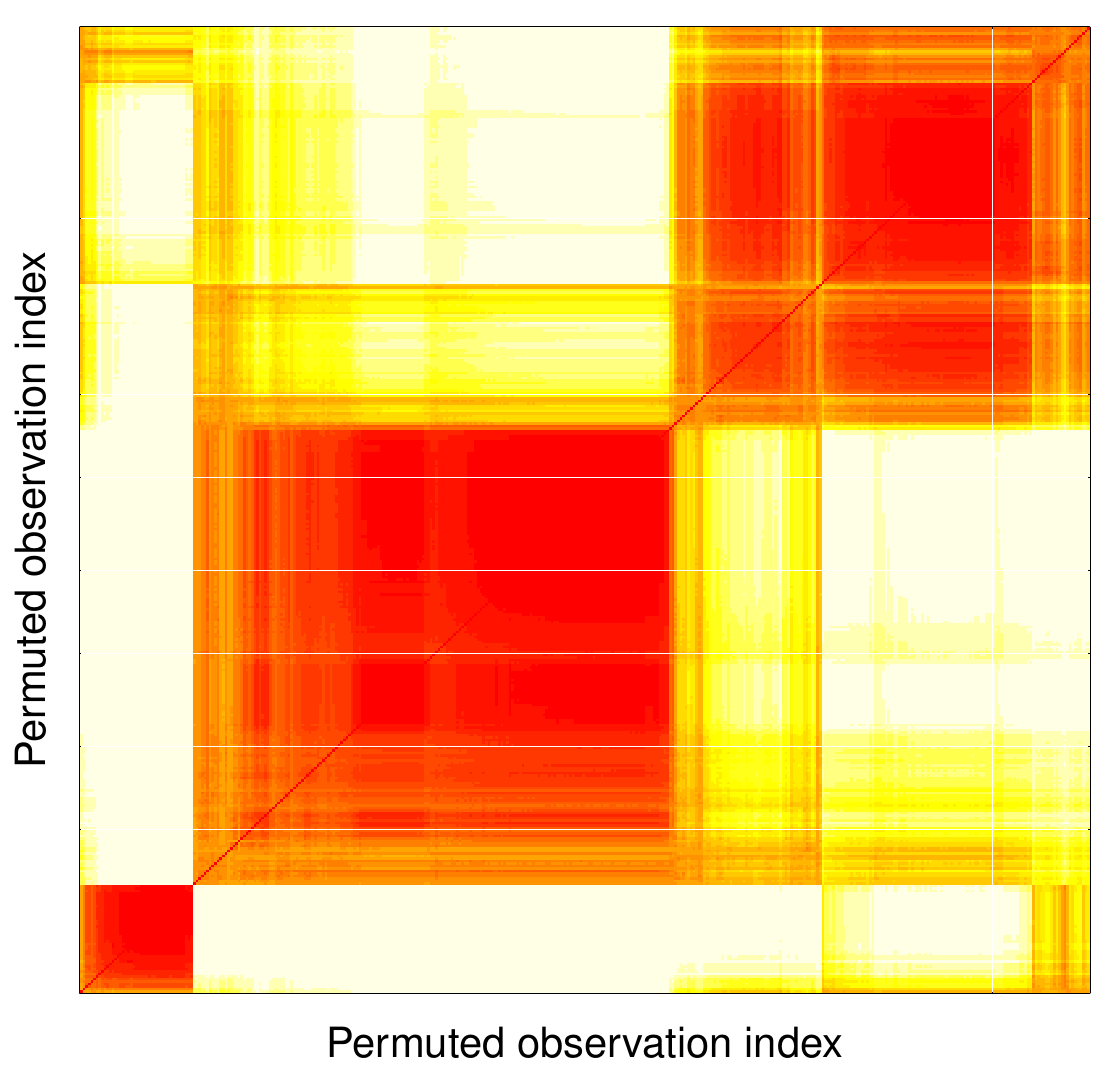}\label{fig:adni_EmE_psm}}
	\caption{Alzheimer's challenge. Heat map of the posterior similarity matrix for the DP and EDP MoE. To improve visualisation, observations are permuted based on hierarchical clustering.}
	\label{fig:adni_psm}
	\end{center}
\vskip -0.2in
\end{figure}

\begin{figure}[!h]
\begin{center}
	\subfigure[Cogntively normal]{\includegraphics[width=0.35\textwidth]{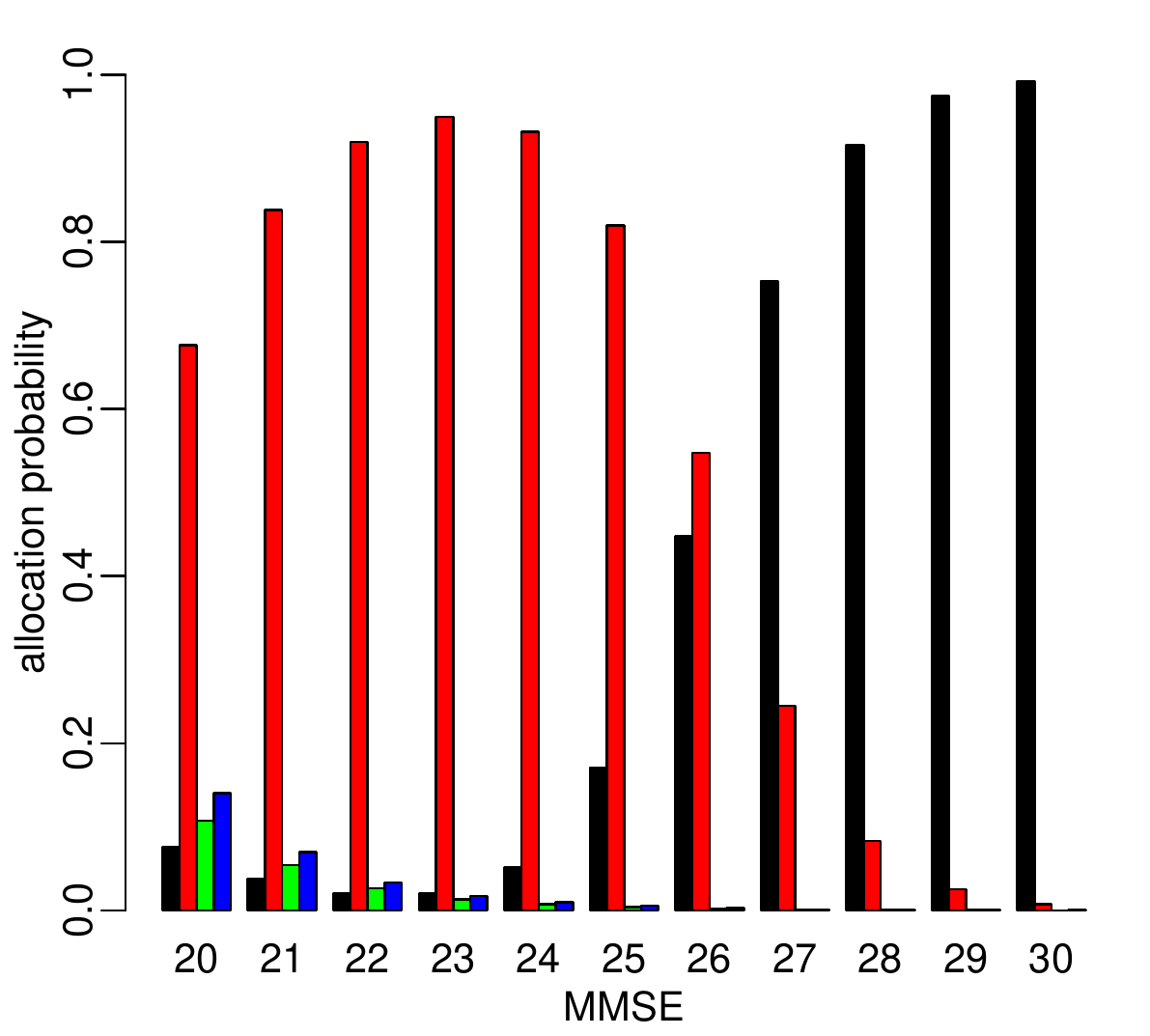}\label{fig:EmE_clusterprob_CN}}
	\subfigure[early MCI]{\includegraphics[width=0.35\textwidth]{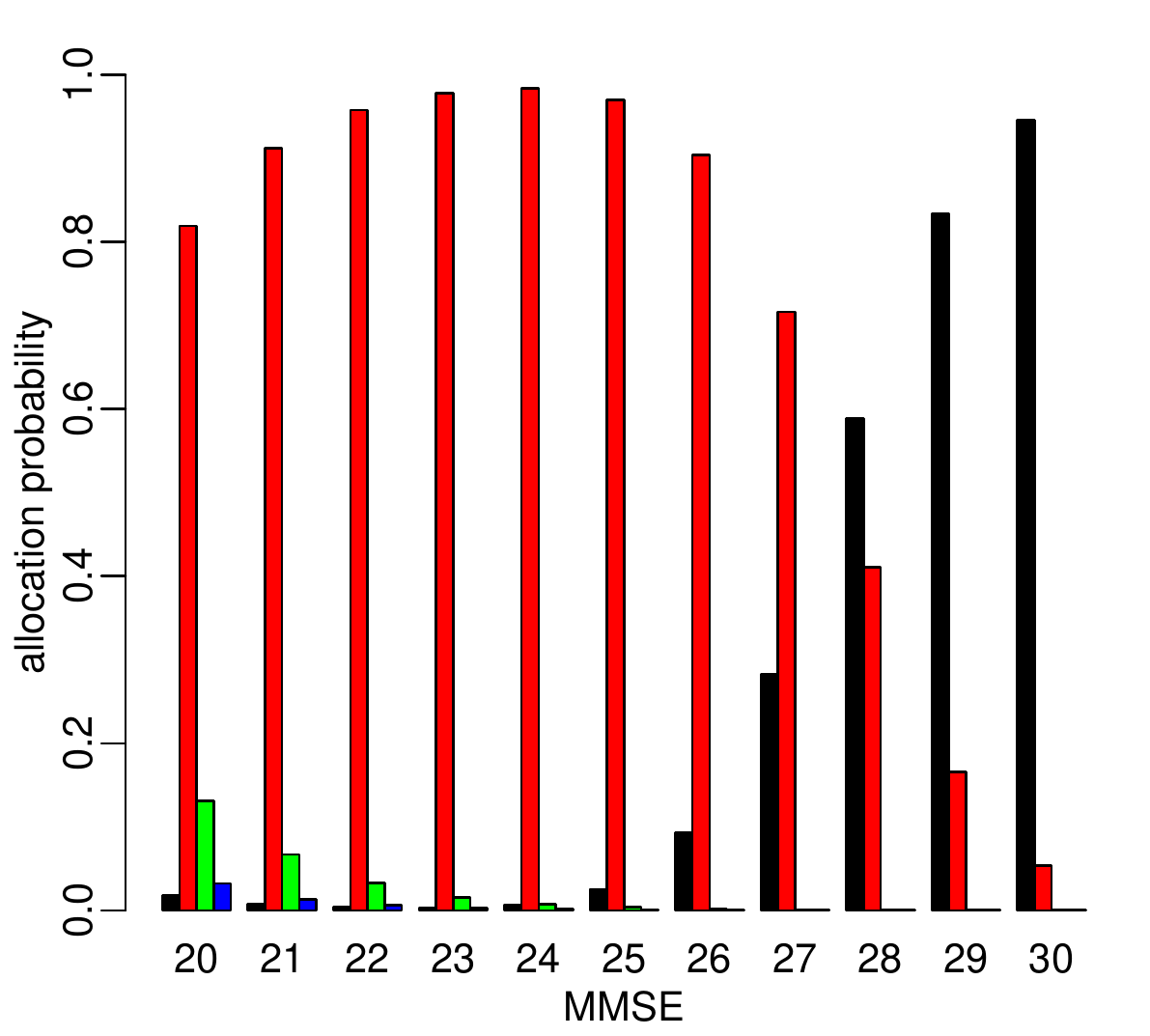}\label{fig:EmE_clusterprob_eMCI}}
\subfigure[late MCI]{\includegraphics[width=0.35\textwidth]{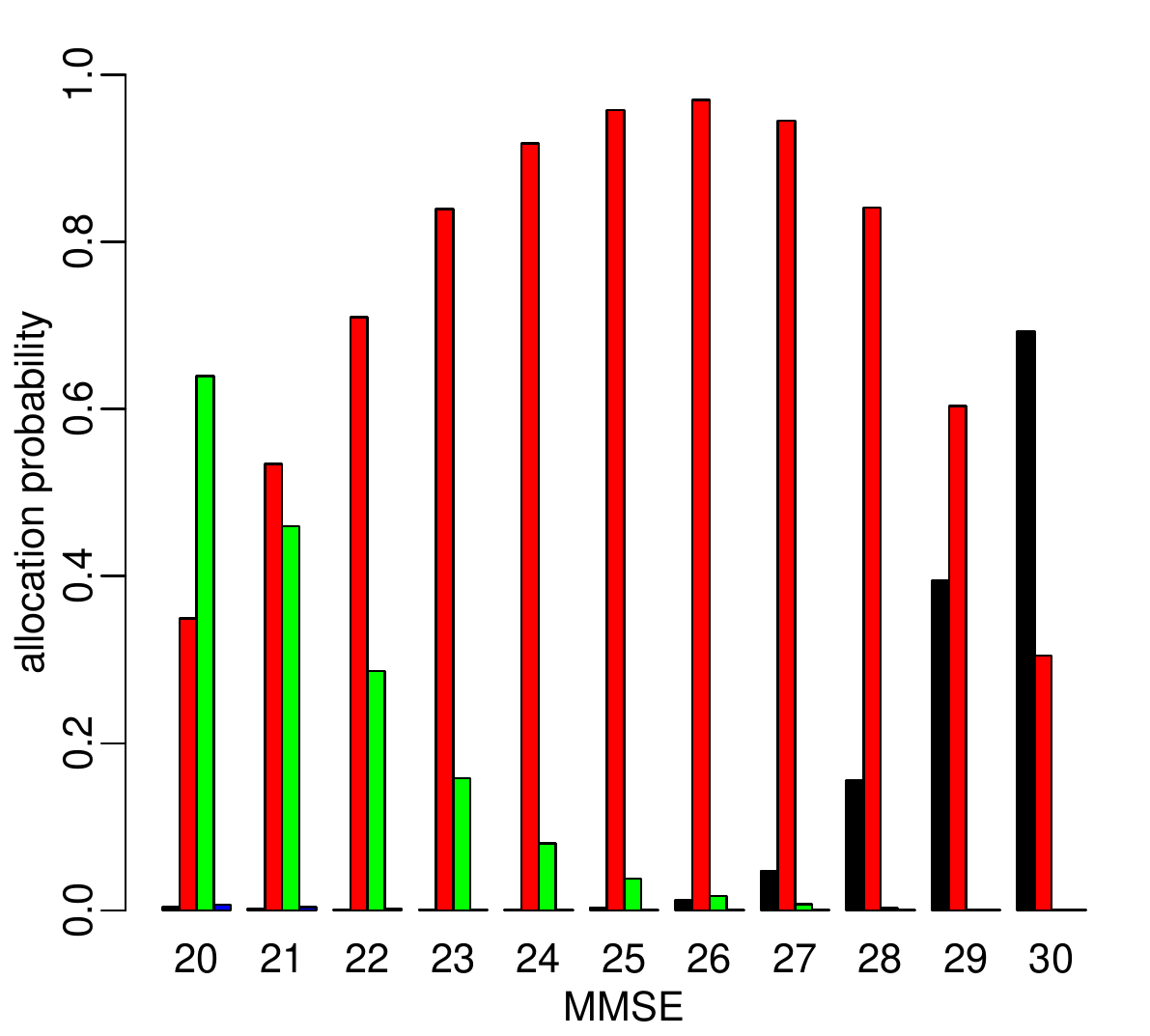}\label{fig:EmE_clusterprob_lMCI}}
\subfigure[Alzheimer's disease]{\includegraphics[width=0.35\textwidth]{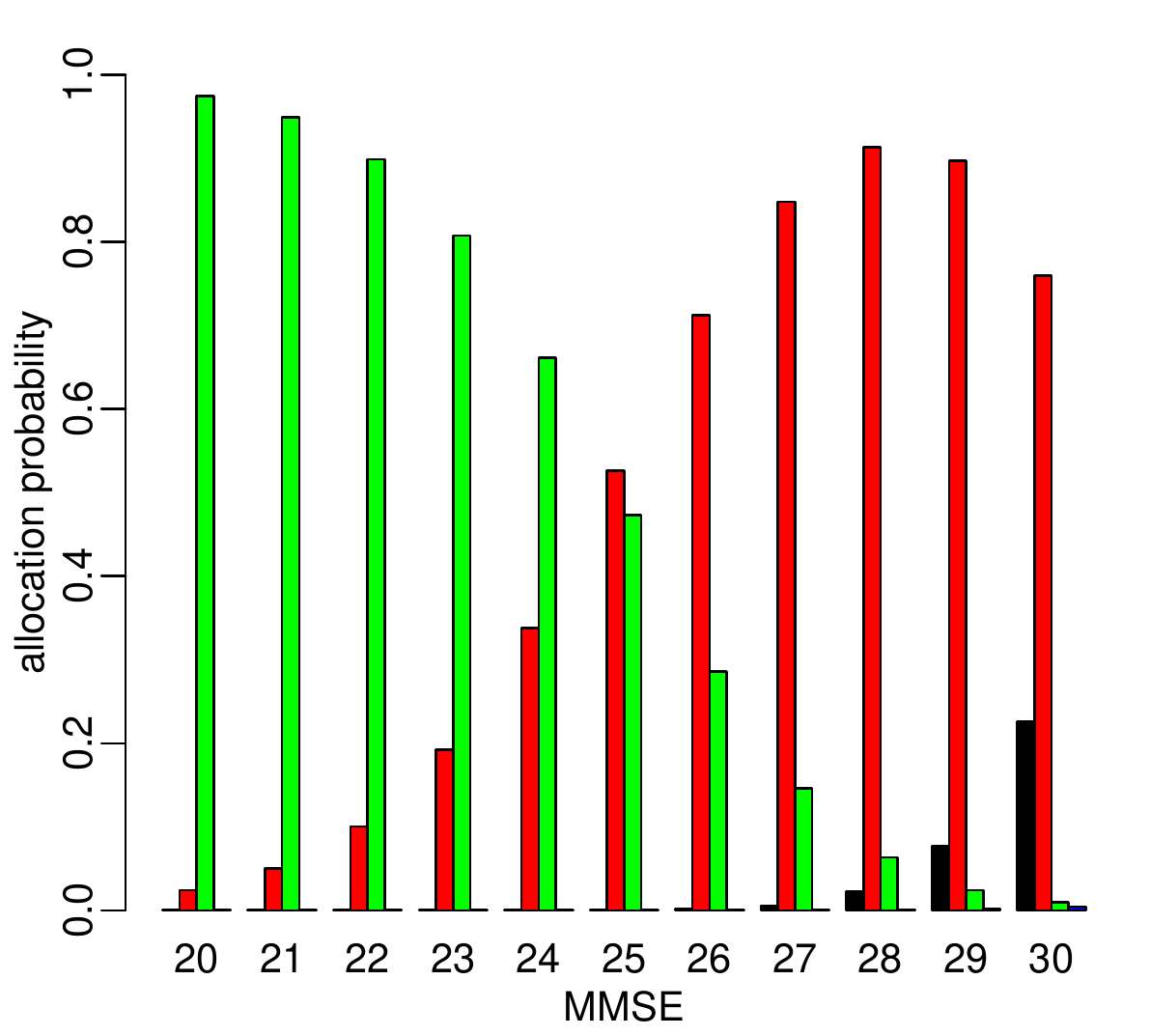}\label{fig:EmE_clusterprob_AD}}	
	\caption{Alzheimer's challenge. The allocation probabilities for a new test point as a function of baseline MMSE and diagnosis of CN in (a), eMCI in (b), lMCI in (c), and AD in (d), with other inputs marginalised. Allocation probabilities are based on the estimated VI clustering and coloured by cluster membership for each of the estimated VI clusters from the enriched model in Figure \ref{fig:adni_VIcluster}.}
	\label{fig:adni_VIcluster_prob}
		\end{center}
\vskip -0.2in
\end{figure}

\begin{figure}[!h]
\begin{center}
\subfigure[Enriched: MMSE]{\includegraphics[width=0.3\textwidth]{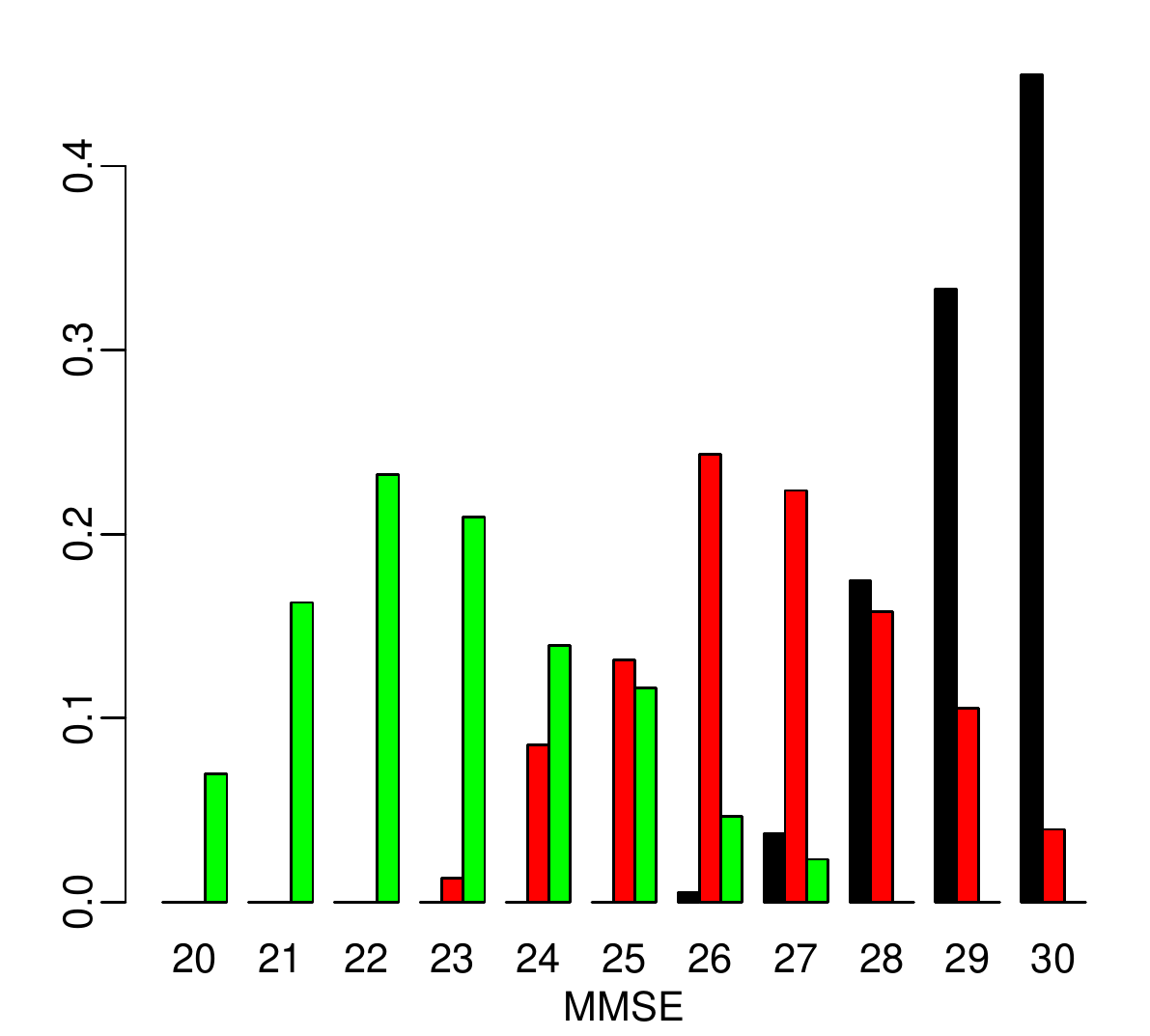}\label{fig:EmE_MMSE_VI}} 
	\subfigure[Enriched: MMSE 24]{\includegraphics[width=0.3\textwidth]{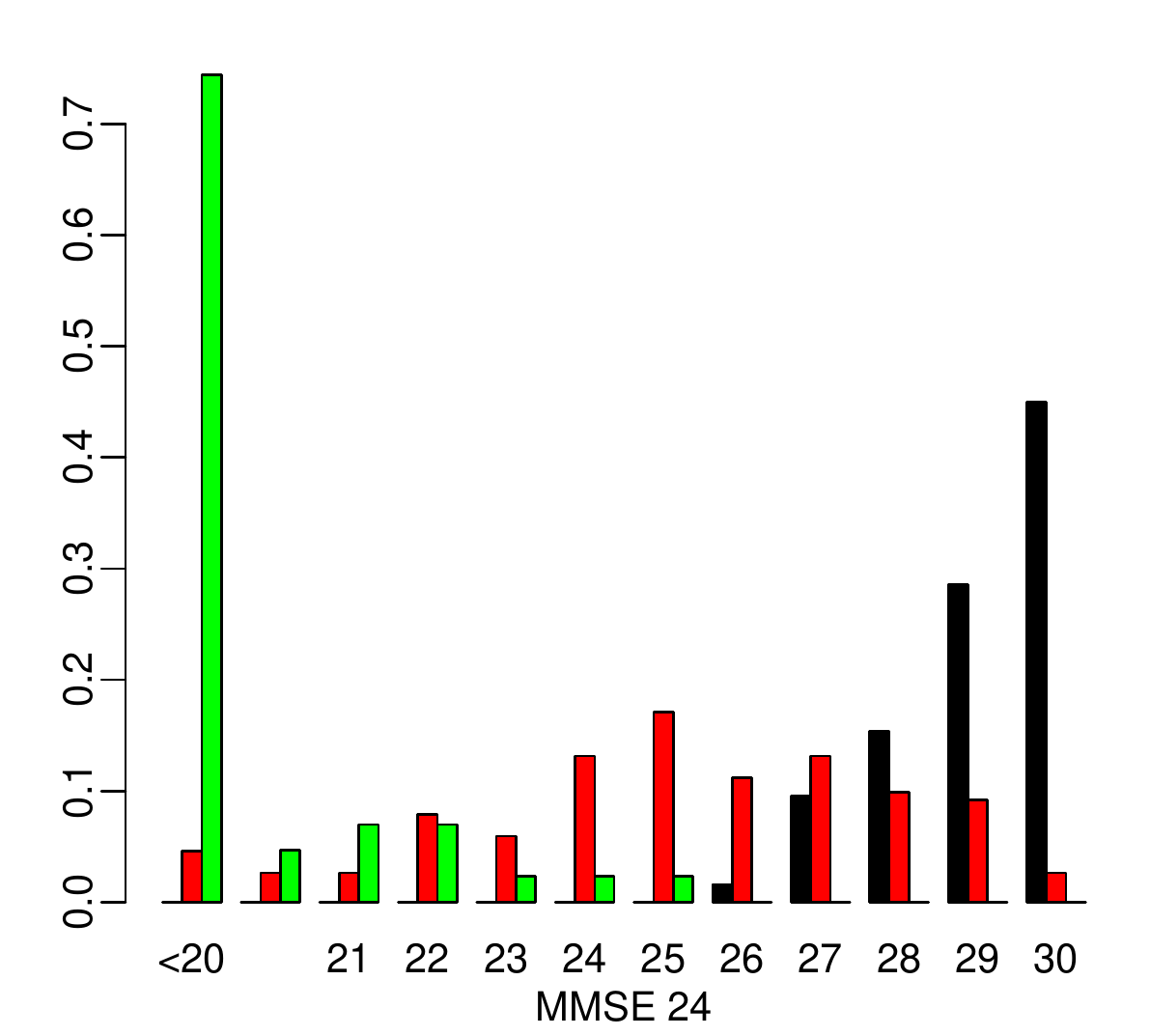}\label{fig:EmE_MMSE24_VI}} 
		\subfigure[Enriched: Education]{\includegraphics[width=0.3\textwidth]{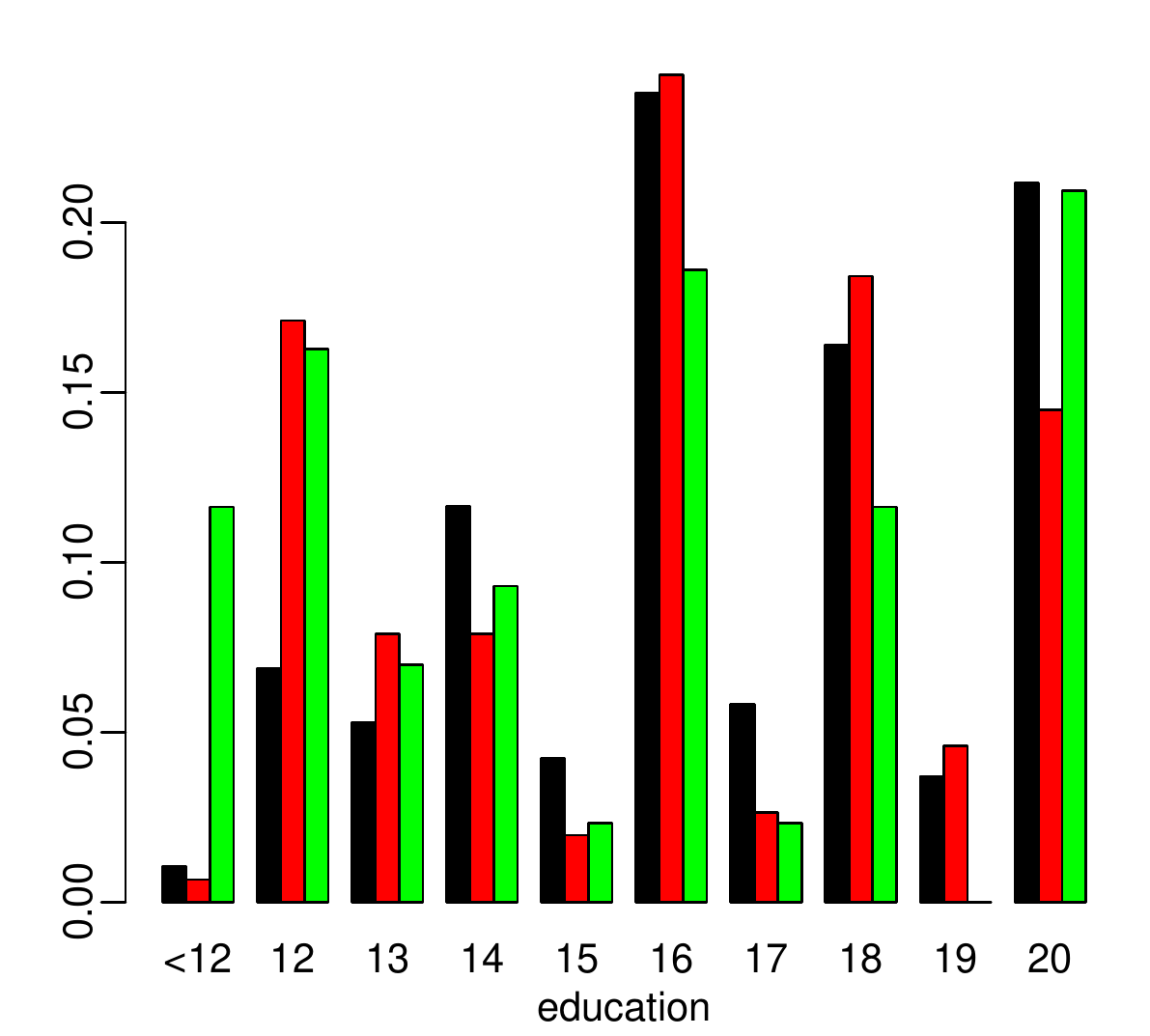}\label{fig:EmE_edu_VI}} 
 \\
 \subfigure[Enriched: Diagnosis]{\includegraphics[width=0.24\textwidth]{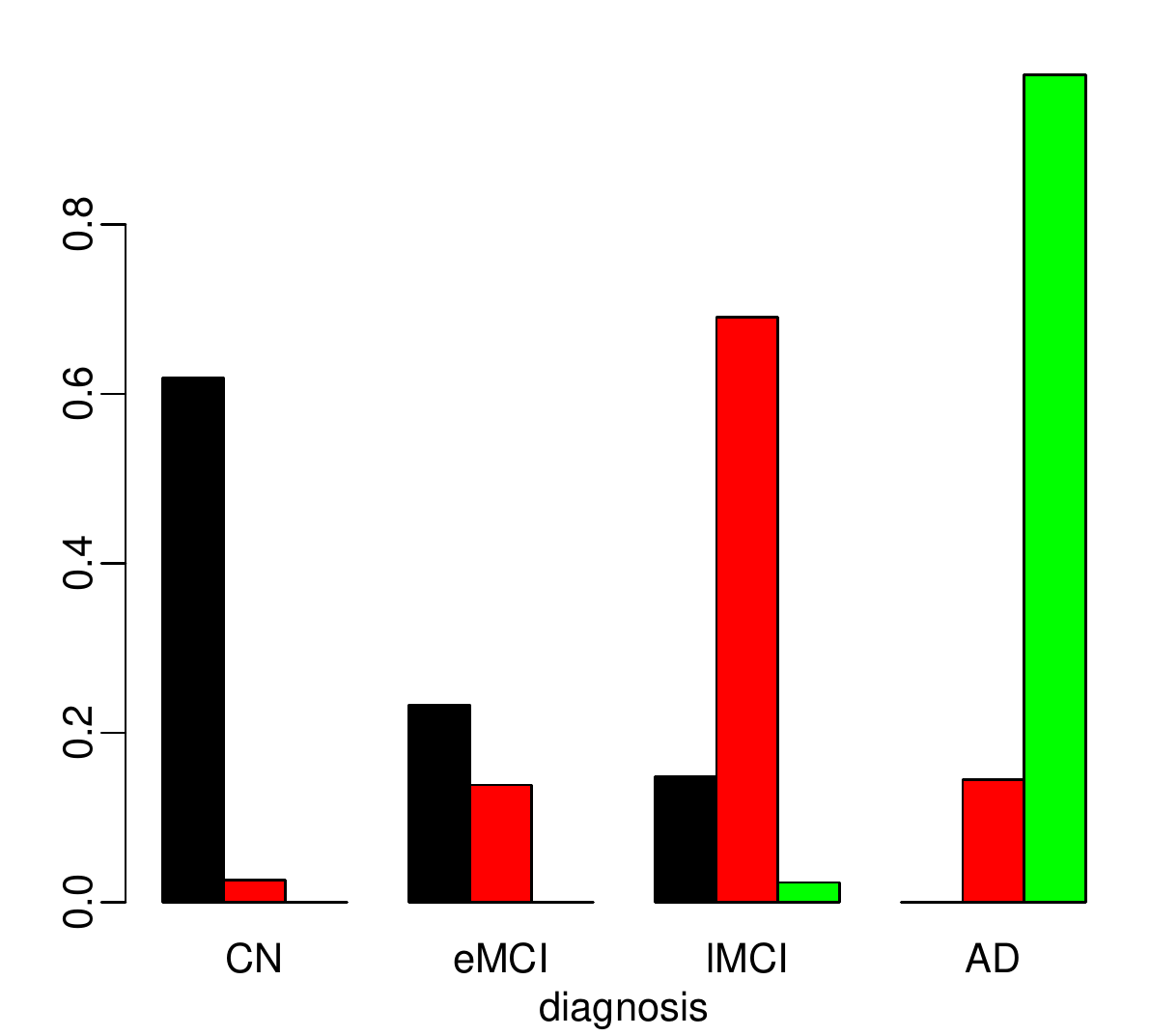}\label{fig:EmE_DX_VI}}
 \subfigure[Enriched: APOE4]{\includegraphics[width=0.24\textwidth]{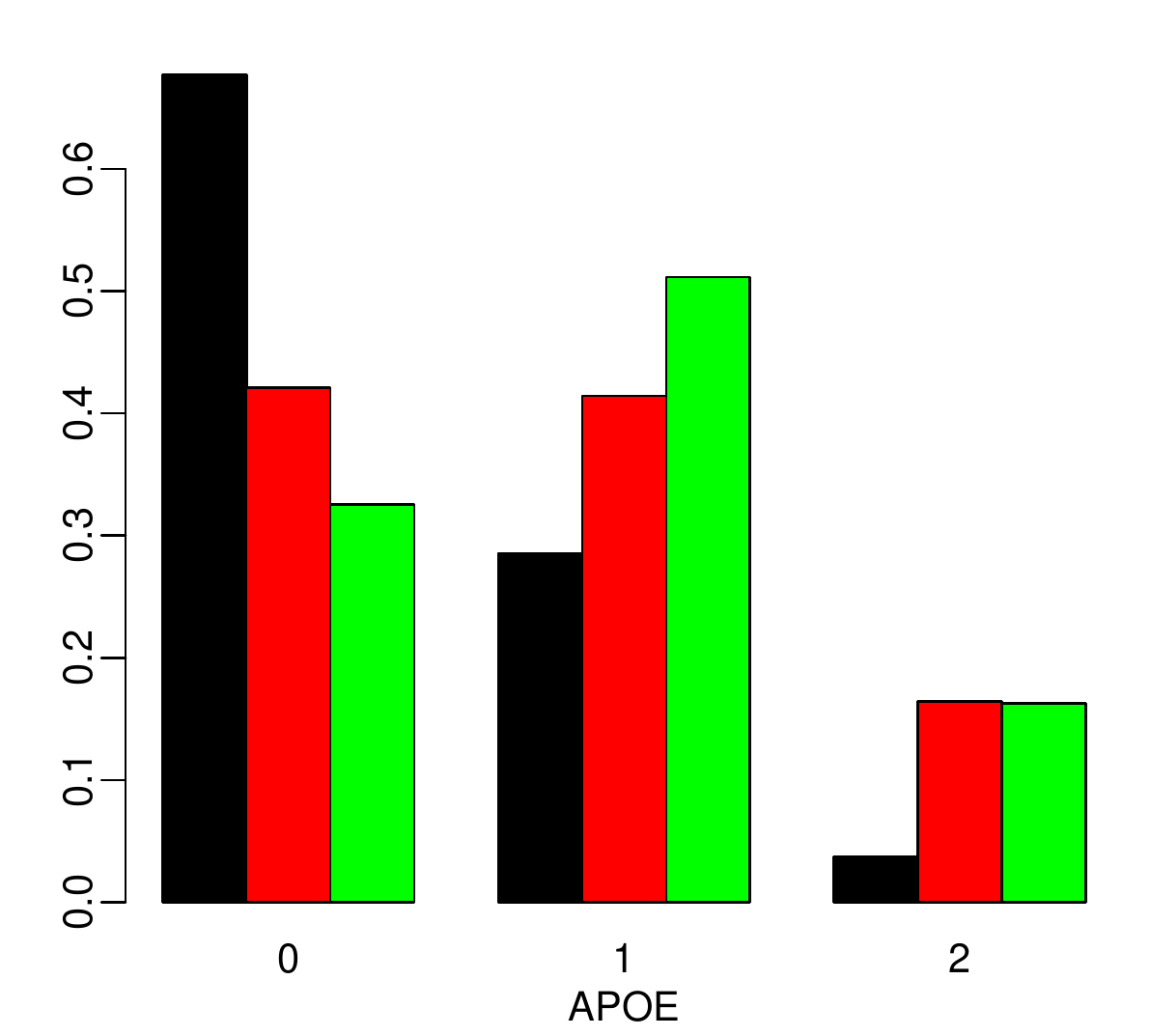}\label{fig:EmE_APOE_VI}} 
	\subfigure[Enriched: Gender]{\includegraphics[width=0.24\textwidth]{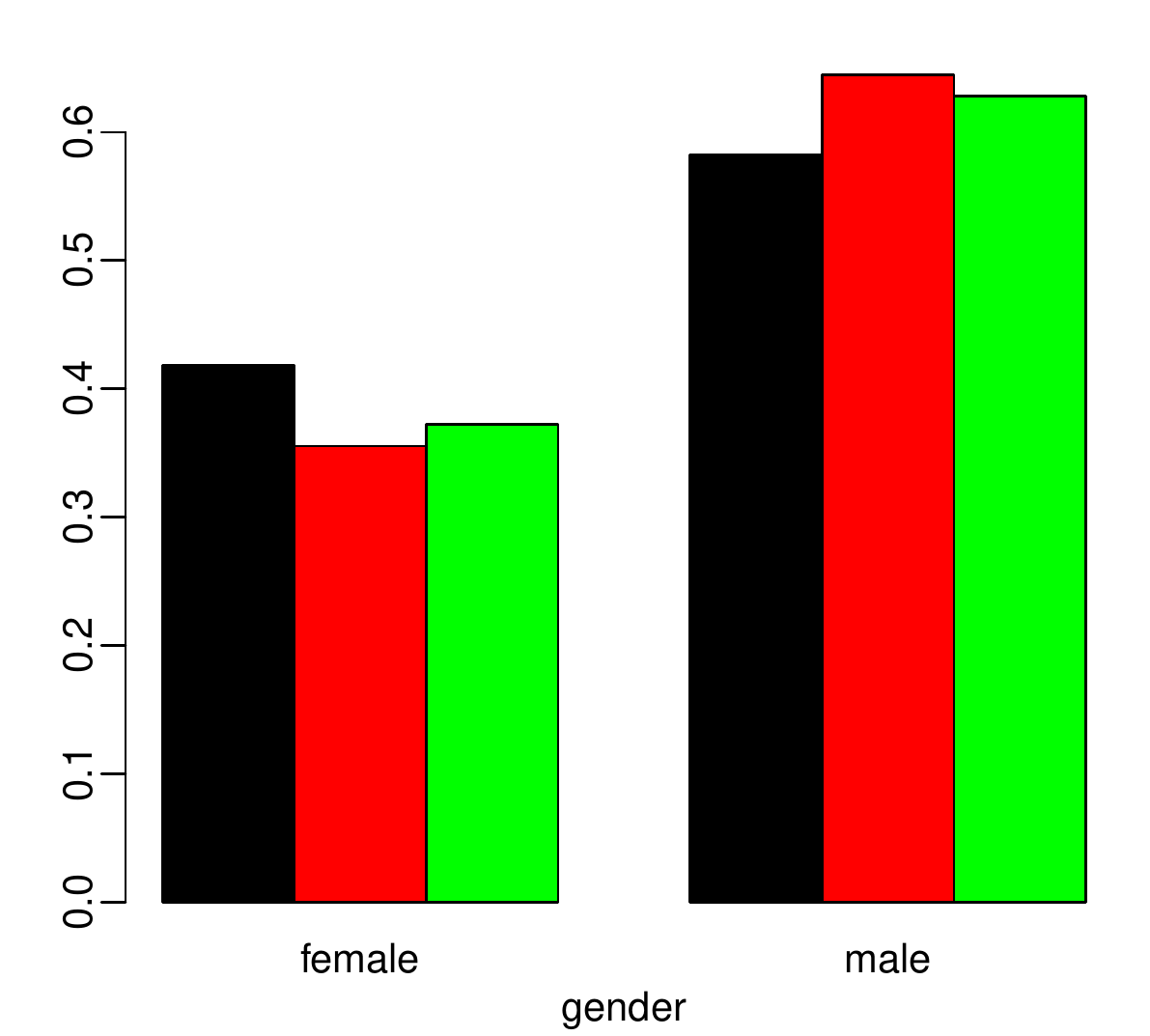}\label{fig:EmE_gender_VI}} 
	\subfigure[Enriched: Age]{\includegraphics[width=0.24\textwidth]{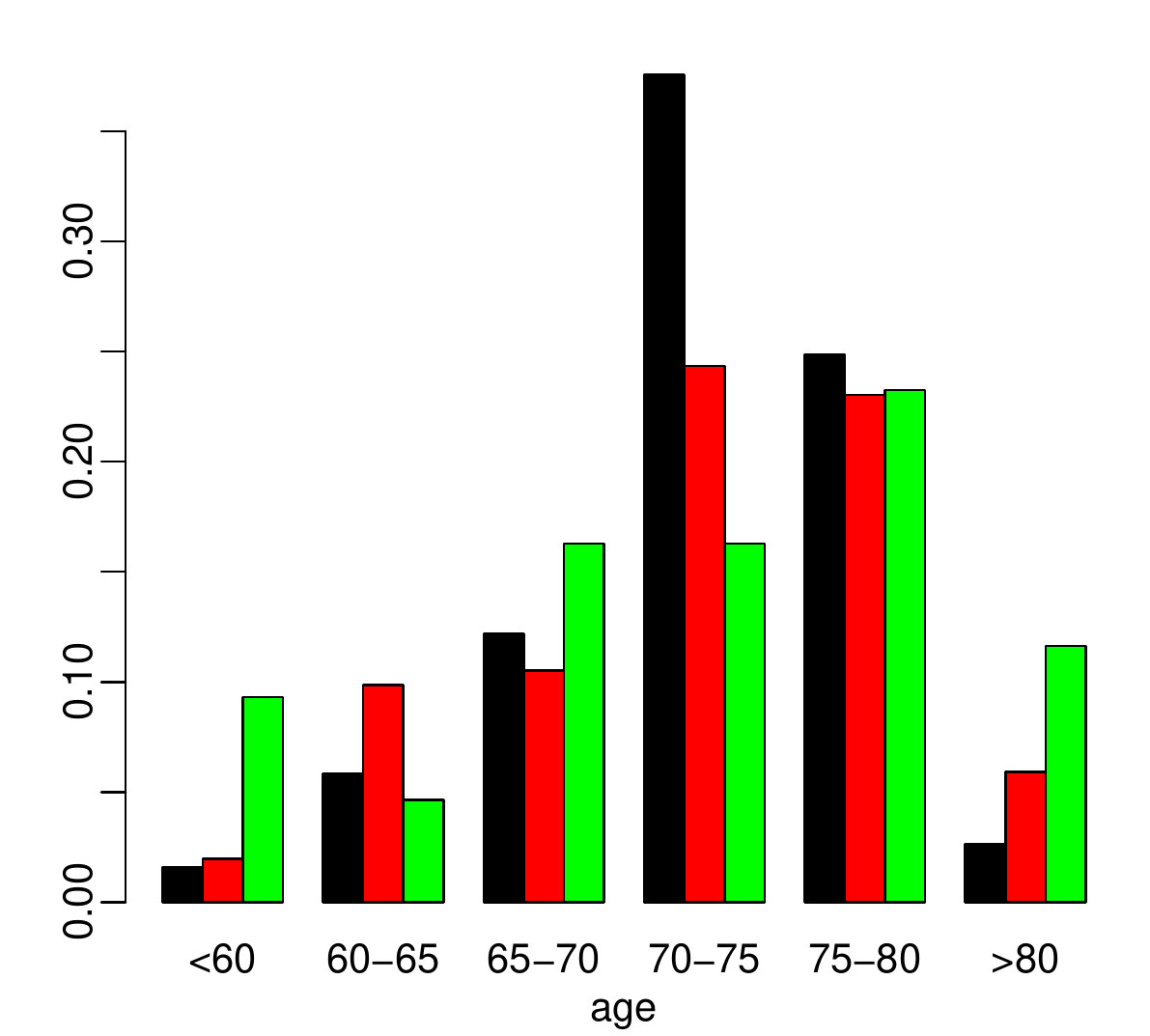}\label{fig:EmE_age_VI}}\\
 \subfigure[Joint: MMSE]{\includegraphics[width=0.3\textwidth]{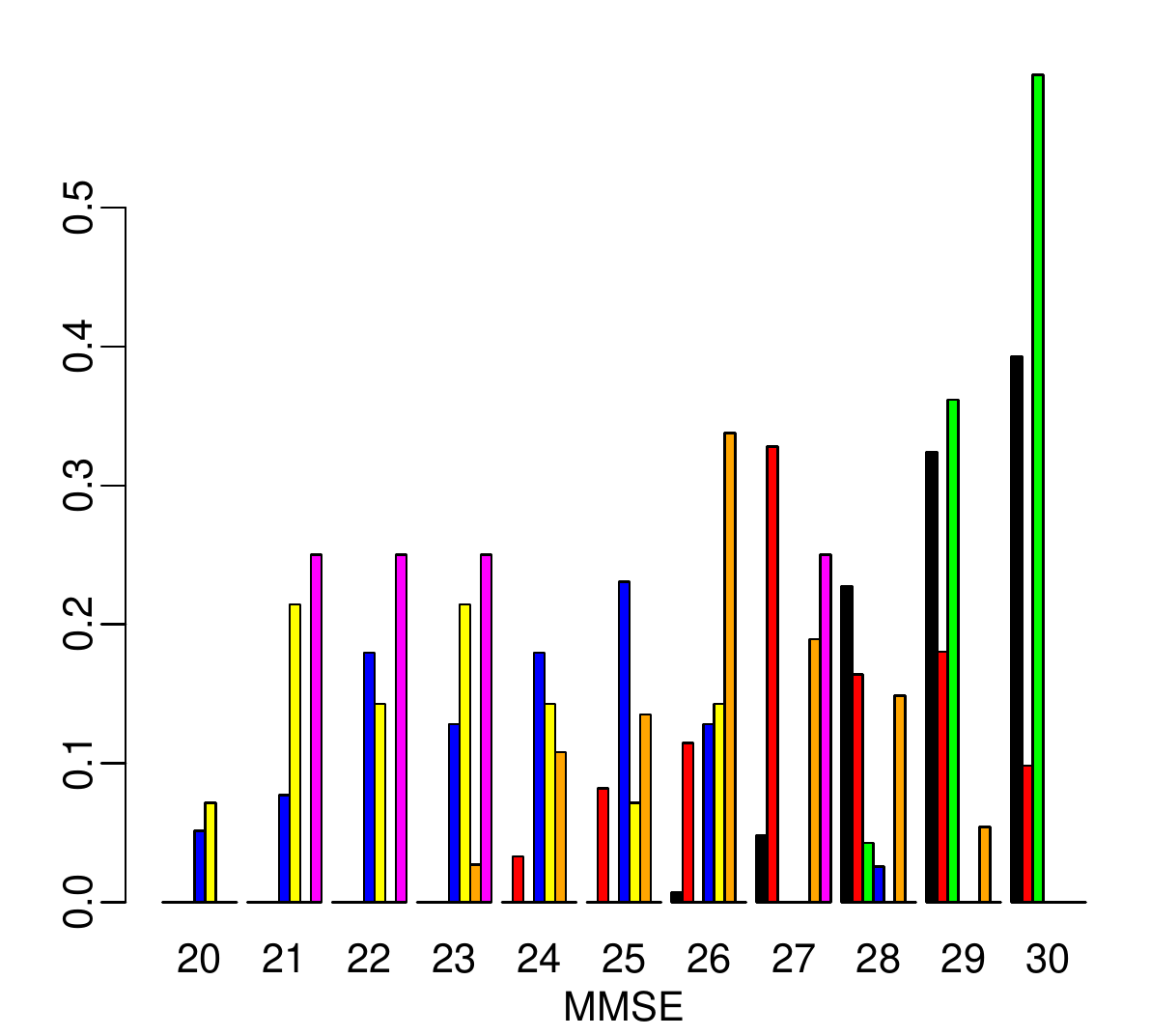}\label{fig:JmE_MMSE_VI}}
	\subfigure[Joint: MMSE 24]{\includegraphics[width=0.3\textwidth]{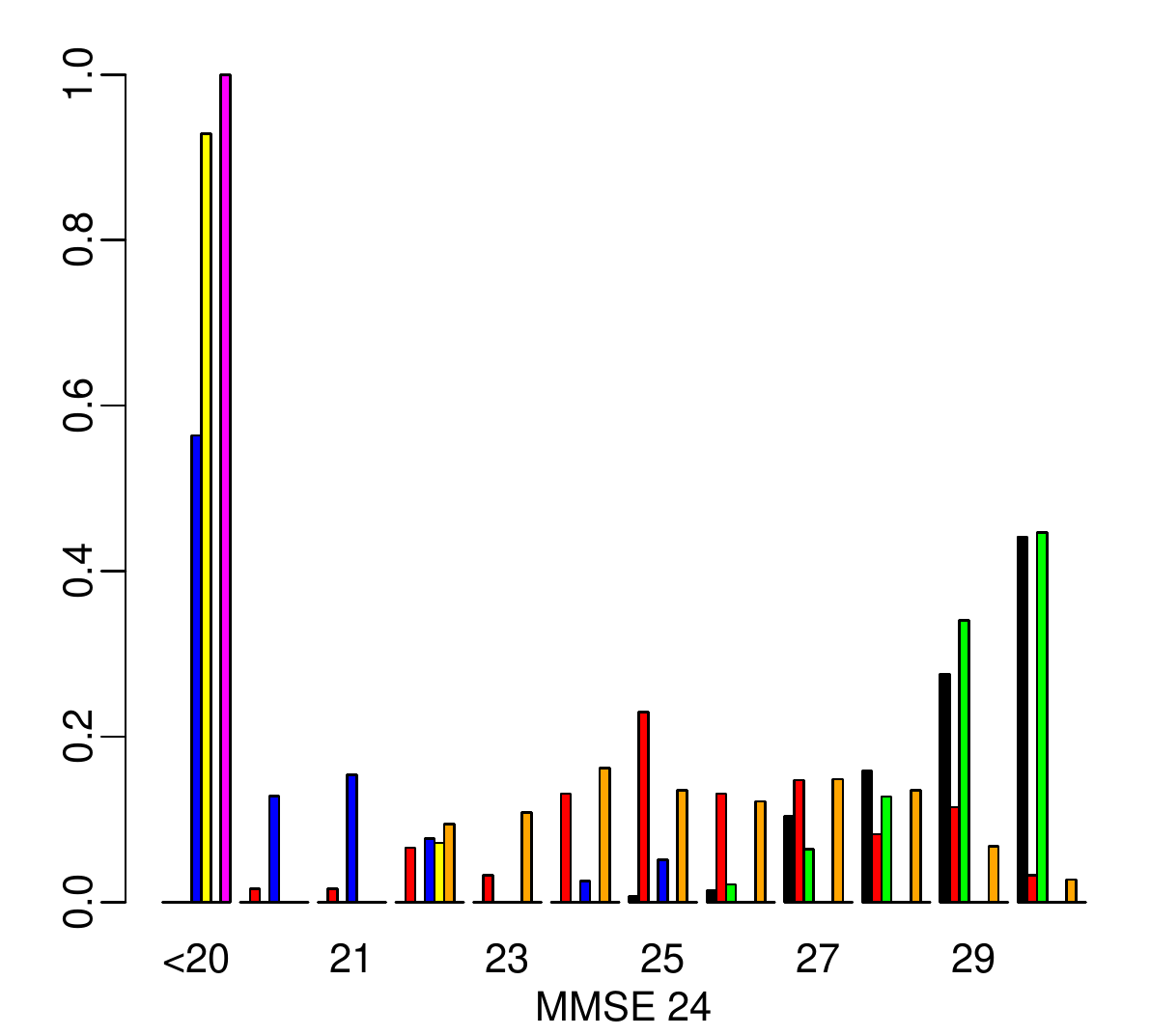}\label{fig:JmE_MMSE24_VI}}
	\subfigure[Joint: Education]{\includegraphics[width=0.3\textwidth]{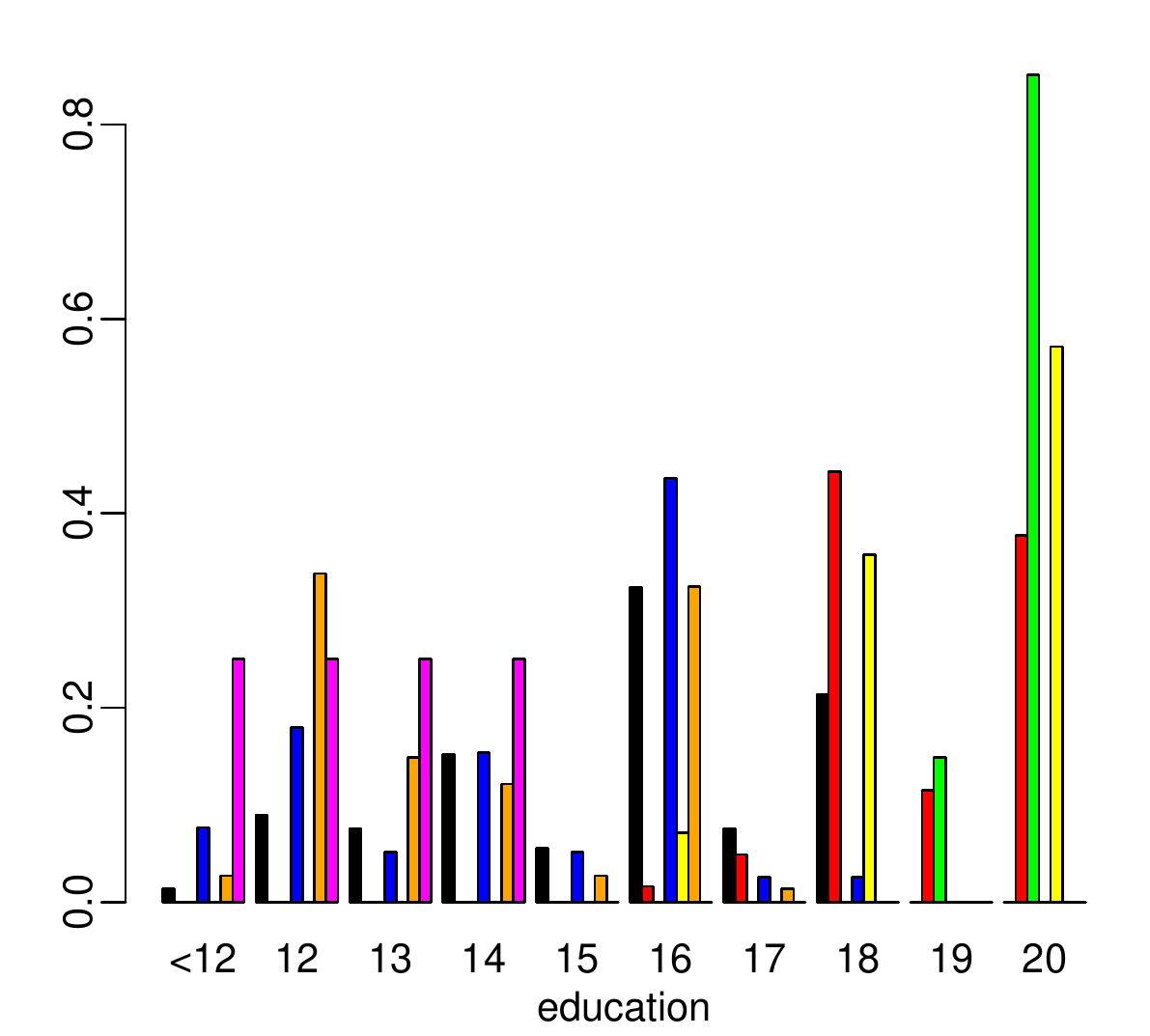}\label{fig:JmE_edu_VI}}\\
		\subfigure[Joint: Diagnosis]{\includegraphics[width=0.24\textwidth]{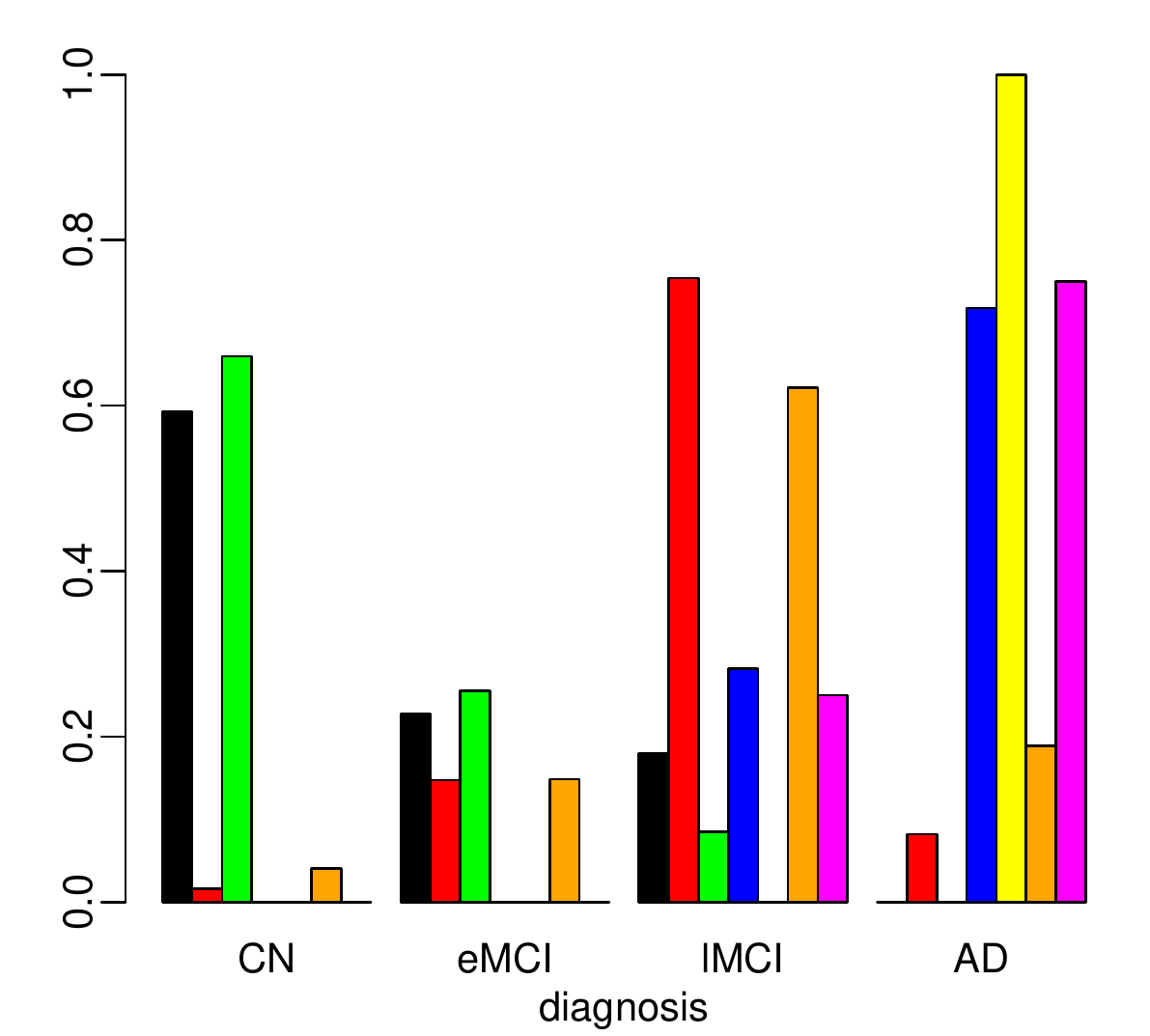}\label{fig:JmE_DX_VIdiag}}
		 \subfigure[Joint: APOE4]{\includegraphics[width=0.24\textwidth]{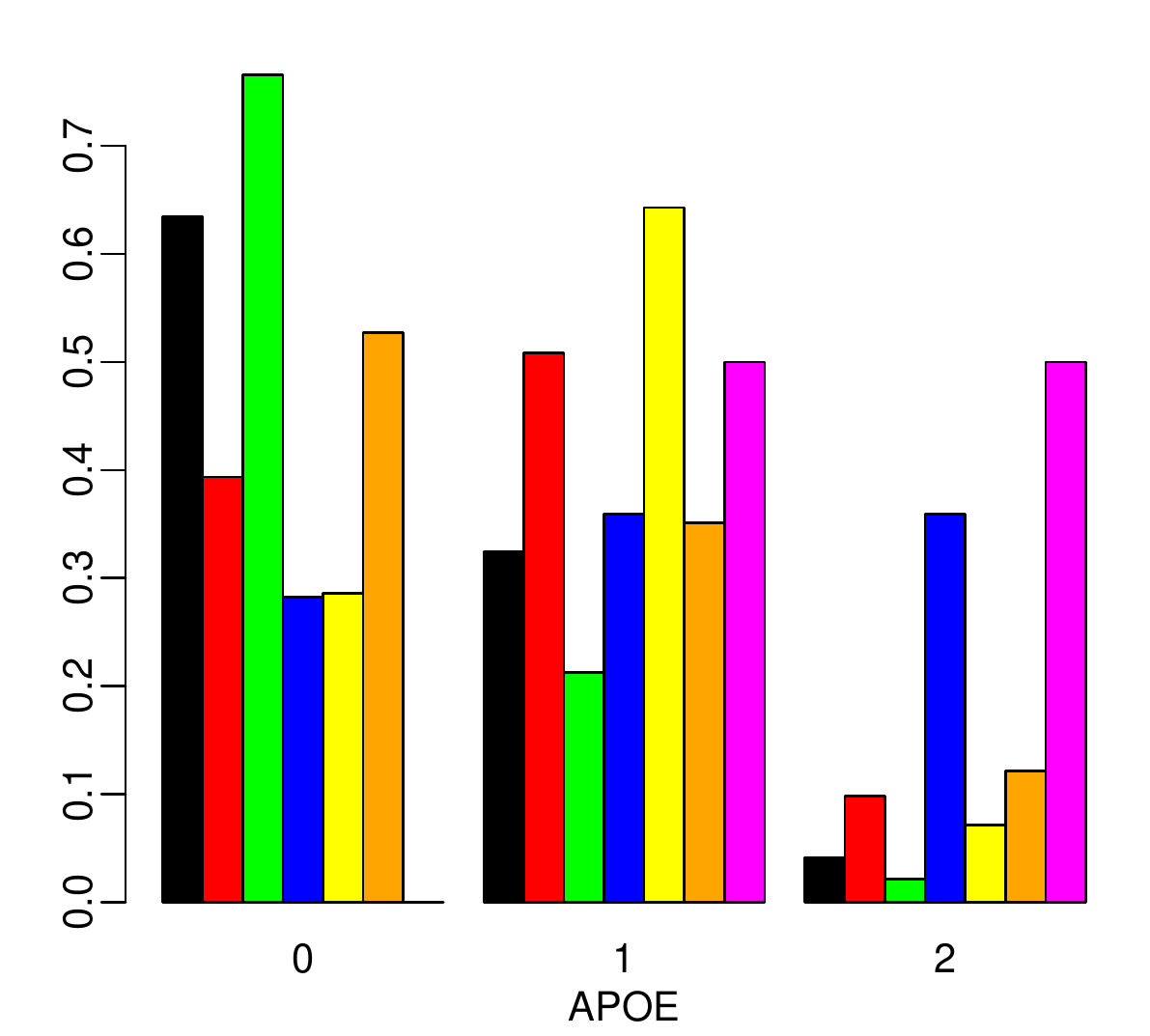}\label{fig:EmE_APOE_VI}} 
	\subfigure[Joint: Gender]{\includegraphics[width=0.24\textwidth]{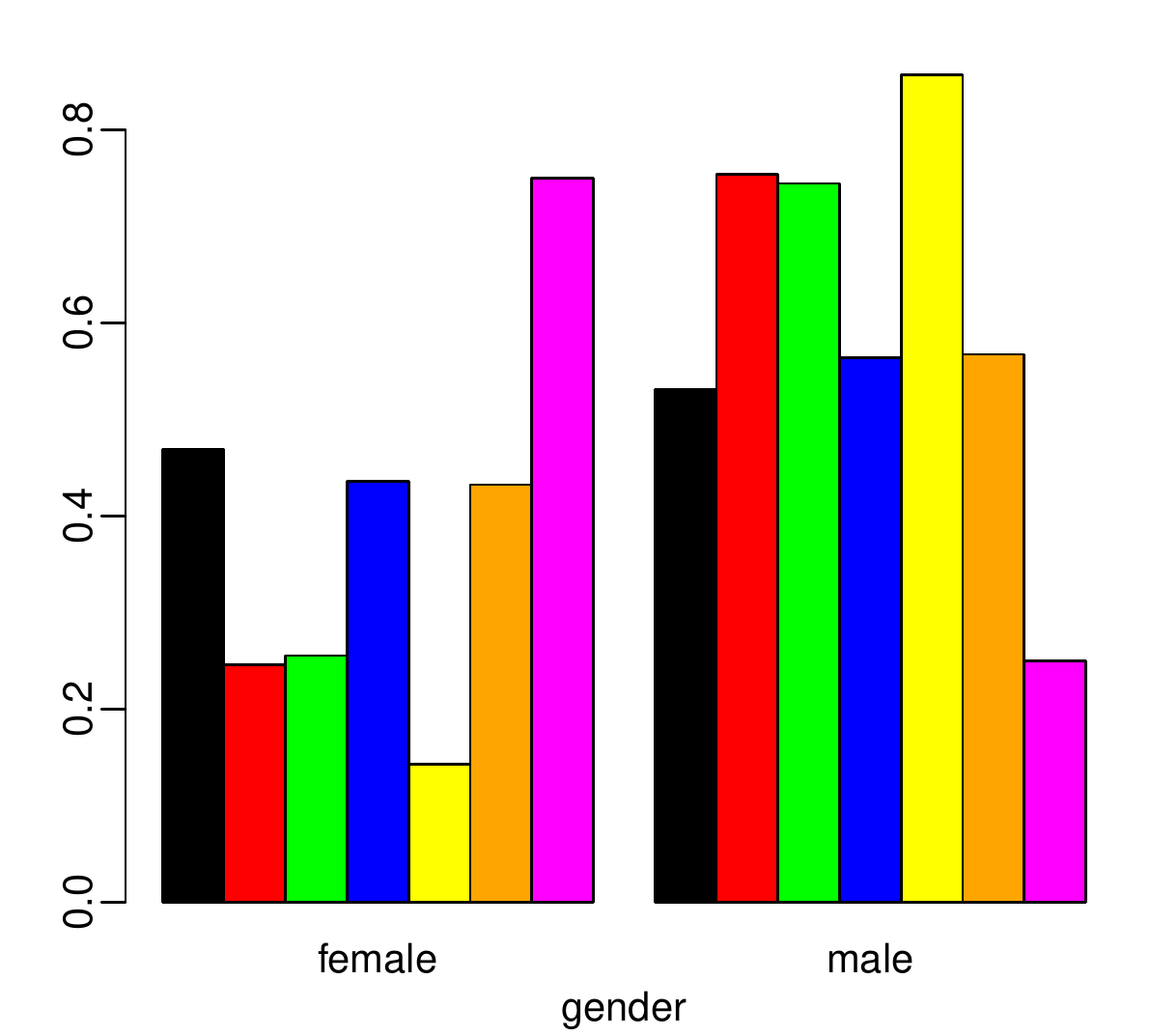}\label{fig:EmE_gender_VI}} 
	\subfigure[Joint: Age]{\includegraphics[width=0.24\textwidth]{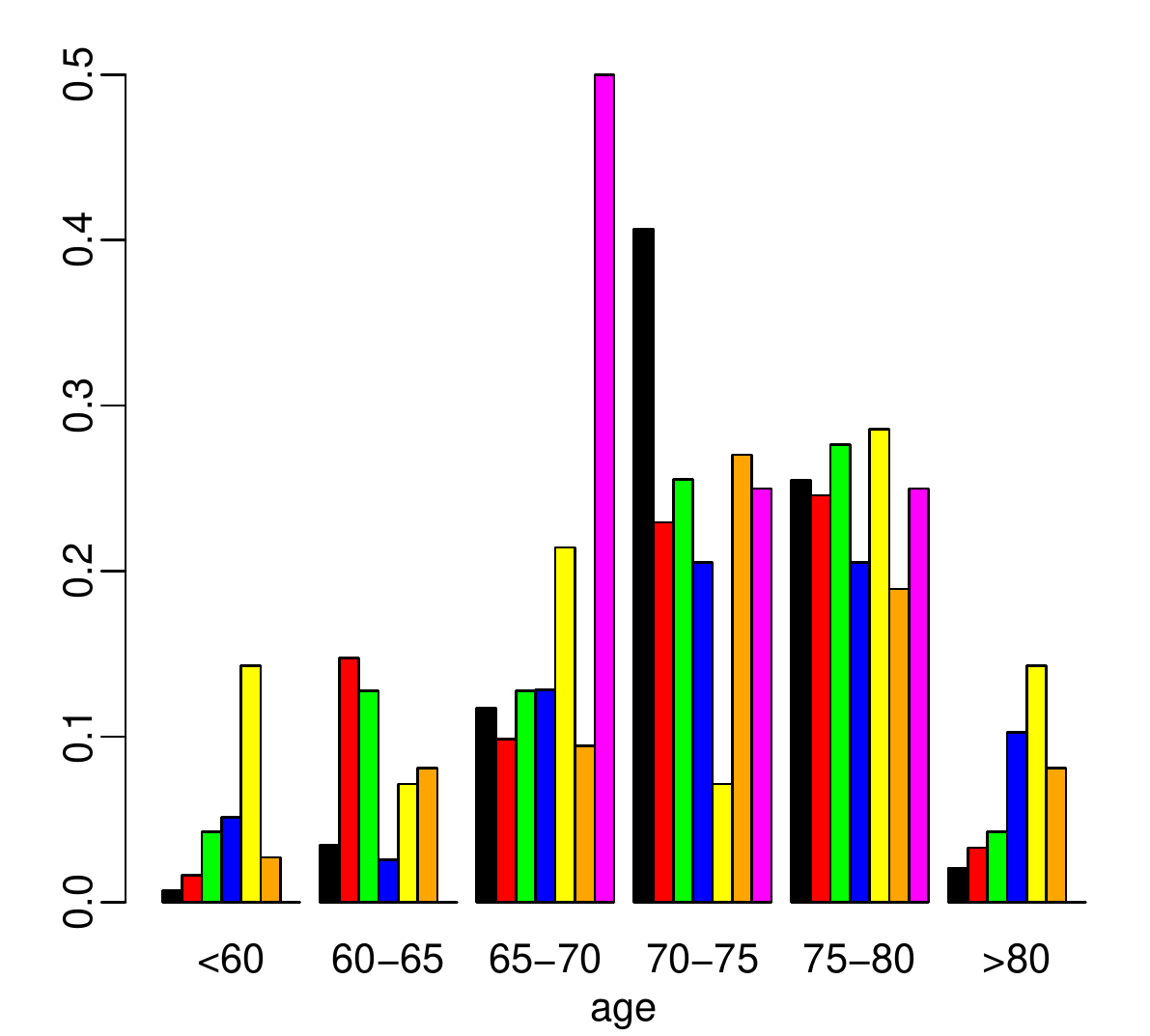}\label{fig:EmE_age_VI}}
		\caption{Alzheimer's challenge. A visualization of the VI clustering estimate through side-by-side bar plots coloured by cluster membership. The first two rows correspond to enriched model and the second two correspond to the joint model.}
	\label{fig:adni_VIcluster}
		\end{center}
\end{figure}

\begin{figure}[!h]
\begin{center}
	\subfigure[$y$-cluster 1]{\includegraphics[width=0.32\textwidth]{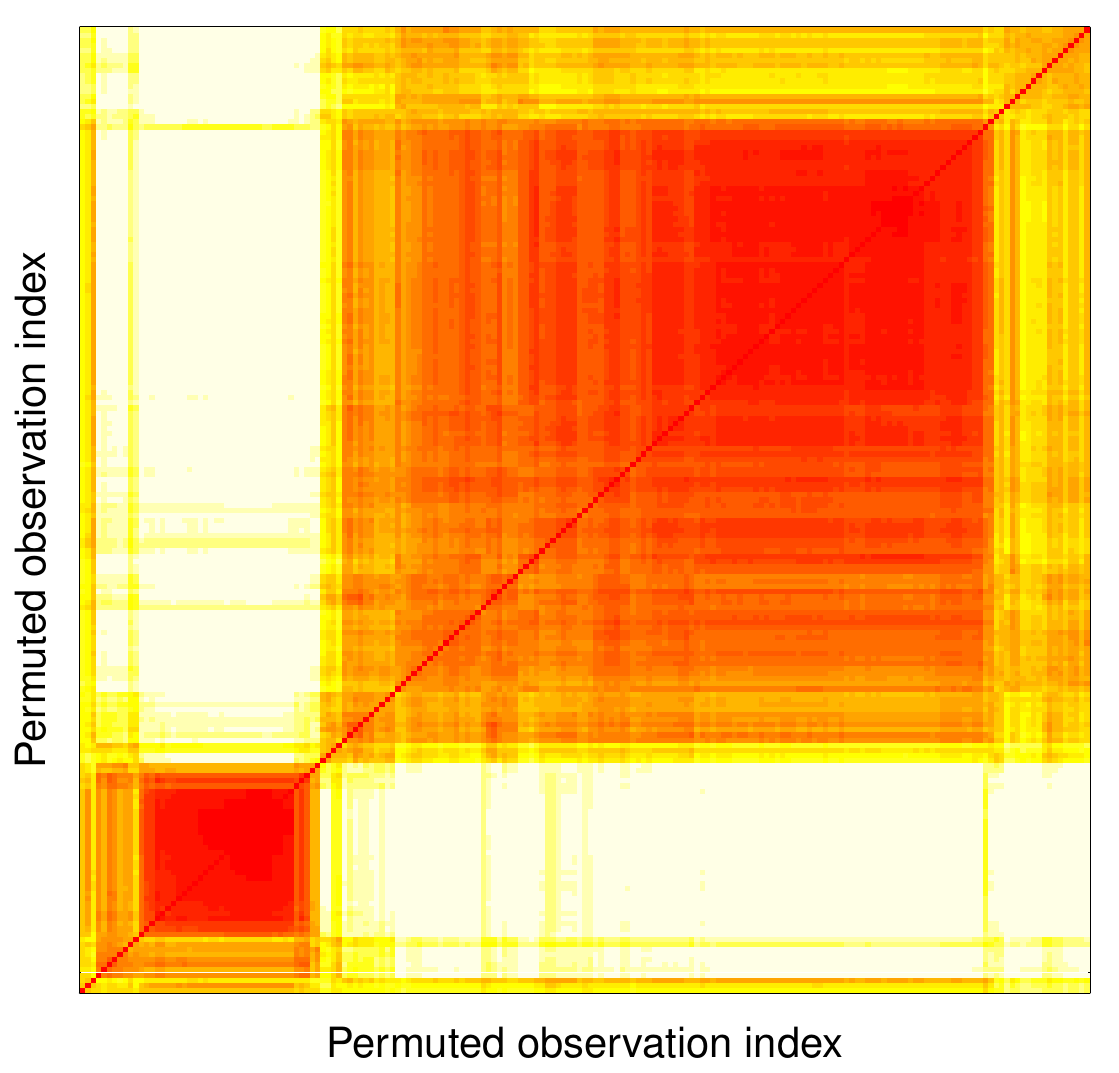}\label{fig:adni_EmE_psm1}}
	\subfigure[$y$-cluster 2]{\includegraphics[width=0.32\textwidth]{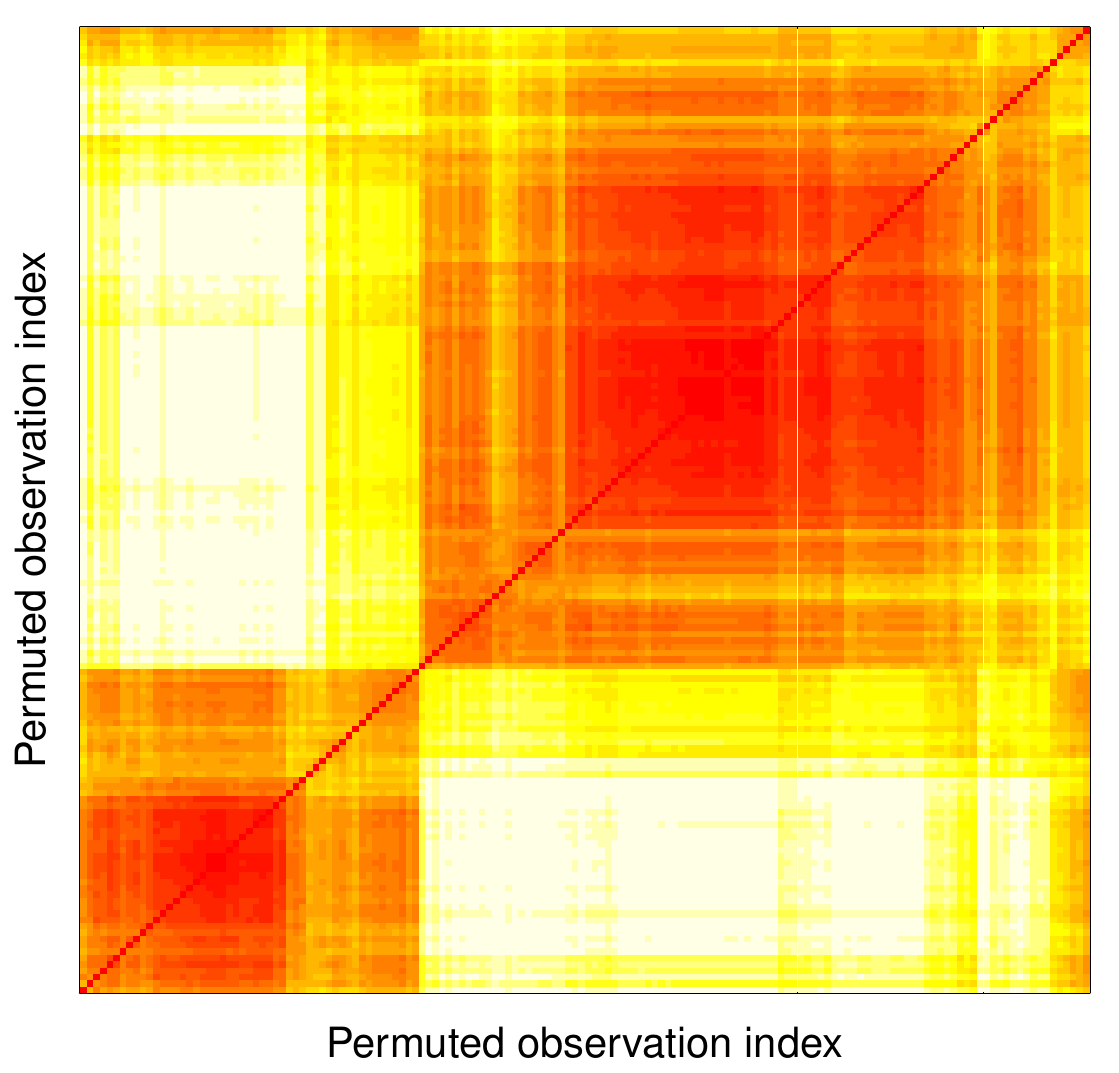}\label{fig:adni_EmE_psm2}}
	\subfigure[$y$-cluster 3]{\includegraphics[width=0.32\textwidth]{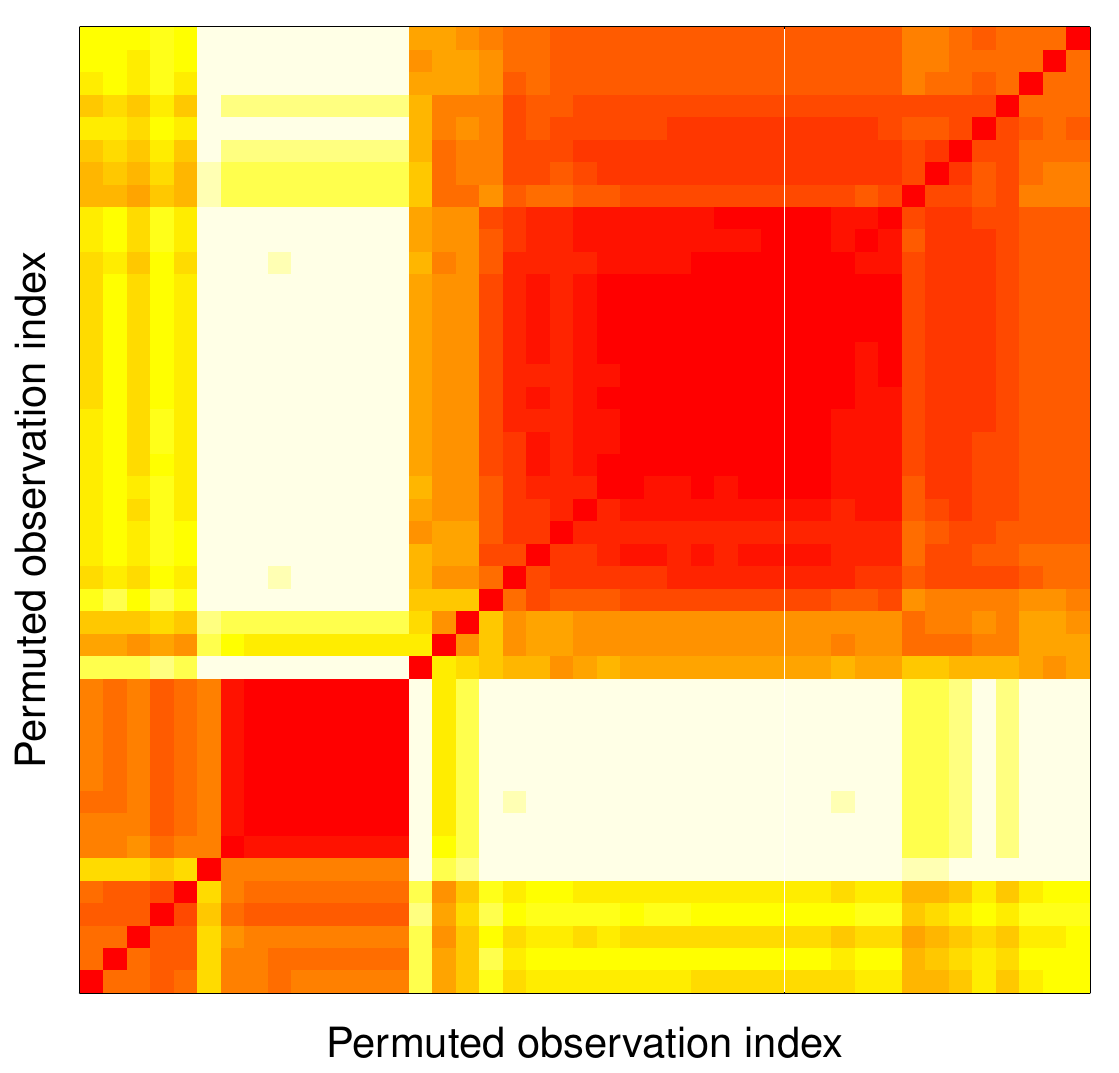}\label{fig:adni_EmE_psm3}}\\
	\subfigure[$y$-cluster 1: Education]{\includegraphics[width=0.32\textwidth]{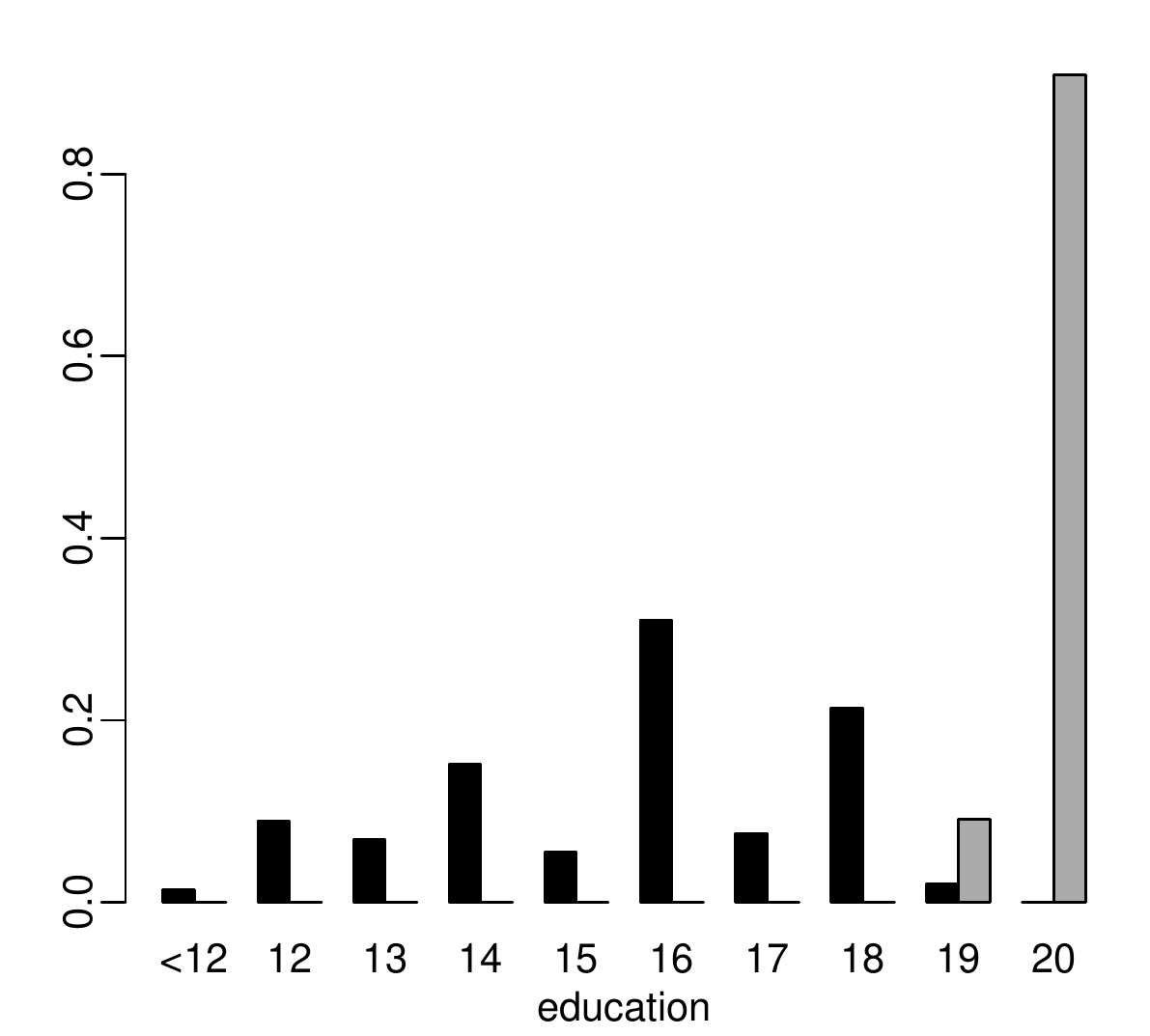}\label{fig:EmE_edu1_VI}} 
	\subfigure[$y$-cluster 2: Education]{\includegraphics[width=0.32\textwidth]{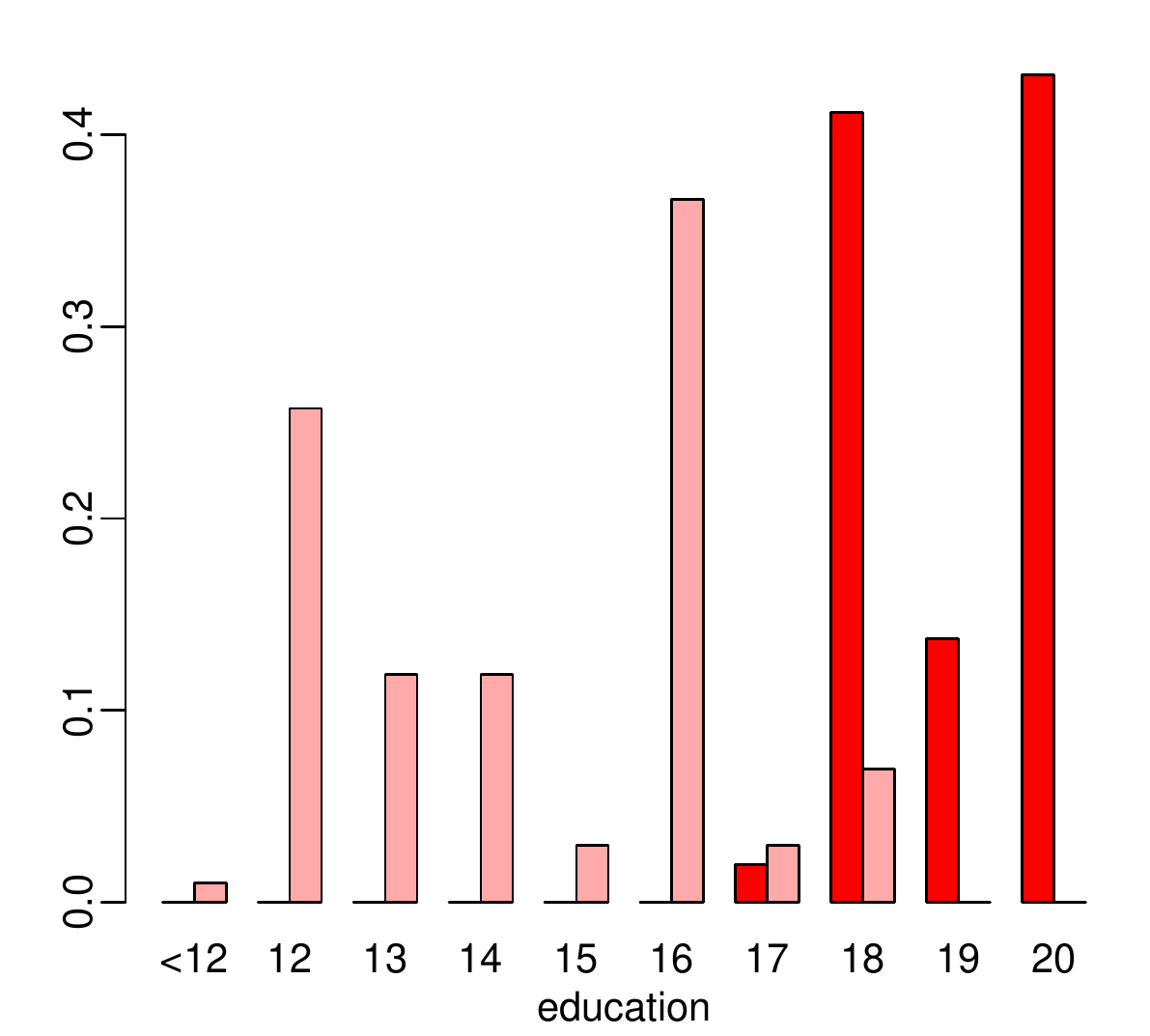}\label{fig:EmE_edu2_VI}} 
	\subfigure[$y$-cluster 3: Education]{\includegraphics[width=0.32\textwidth]{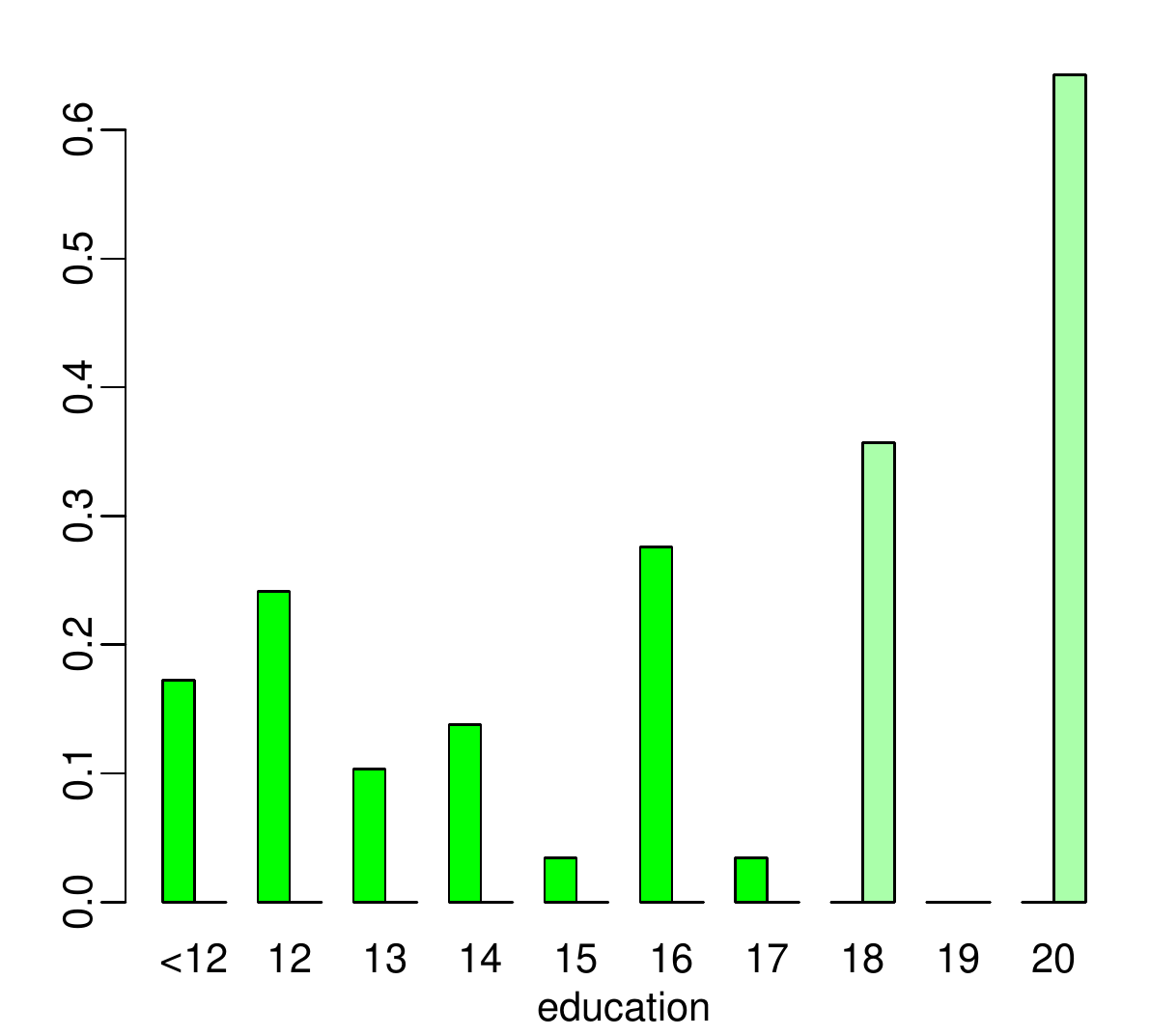}\label{fig:EmE_edu3_VI}}
	\caption{Alzheimer's challenge. First row: heat maps of the posterior similarity matrices for the $x$-clustering within each $y$-cluster for the enriched model. Second row:  a visualization of the VI $x$-clustering estimate within each $y$-cluster through side-by-side bar plots for education. Colour corresponds to the $y$-cluster, while shading corresponds to the $x$-cluster.}
	\label{fig:adni_psmx}
			\end{center}
\vskip -0.2in
\end{figure}

For comparison with the best performers of this subchallenge, the GuanLab and ADDT teams, we implemented the models using publicly available packages in \pkg{R}. For the GuanLab model, we used the \pkg{svm} function of the \pkg{e1071} package \citep{e1071}. For the ADDT model, we used the \pkg{rlm} function of the \pkg{MASS} package \citep{MASS}. For both mixtures of experts, posterior medians, i.e. the point estimate under the absolute error loss, are used to predict MMSE scores, which are appropriate due to the heavy left tail of the predictive densities. 

Heat maps of the posterior similarity matrices are provided in Figure \ref{fig:adni_psm}, and visualizations of the VI clusterings through side-by-side bar plots of MMSE baseline, MMSE follow-up, education, diagnosis, APOE4, gender and age are provided in Figure \ref{fig:adni_VIcluster}, with colours representing clusters. Interestingly, the enriched model identifies three clusters consisting mostly of cognitively normal (black), mild cognitive impairment (red), and AD (green) individuals, similar to the GuanLab model, with slight modifications considering the other variables, particularly, MMSE baseline and follow-up scores. For example, one late MCI individual is allocated to the AD (green) cluster in Figure \ref{fig:EmE_DX_VI} due to the observed sharp drop in MMSE from 27 at baseline to 8 at follow-up. Additionally, we observe that the relative proportion of individuals in the red and green clusters increases with higher APOE4, but does not (marginally) depend on gender and age. 

The DP, on the other hand, further subdivides clusters due to multimodality in education. Similarly, for the enriched model, the VI estimate of $x$-clustering within each VI estimated $y$-cluster, contains two $x$-clusters due to multimodality in education. Figure \ref{fig:adni_psmx} depicts the heap map of the posterior similarity matrix for the $x$-clustering within each estimated $y$-cluster and also shows the VI estimate of $x$-clustering within each VI estimated $y$-cluster for education, with each estimated $x$-clustering containing two clusters. 

We can further appreciate the difference between the deterministic clustering of the GuanLab model and the stochastic clustering of the enriched model in Figure \ref{fig:adni_VIcluster_prob}, which shows the allocation probabilities of a new test point for MMSE baseline scores of 20-30 and diagnosis of CN (Figure \ref{fig:EmE_clusterprob_CN}), eMCI (Figure \ref{fig:EmE_clusterprob_eMCI}), lMCI (Figure \ref{fig:EmE_clusterprob_lMCI}), AD (Figure \ref{fig:EmE_clusterprob_AD}), with other inputs marginalised. As opposed to the GuanLab model which classifies new individuals based on diagnosis, we observe that CN individuals with baseline $\text{MMSE}\geq 27$ have the highest probability of being allocated to the black cluster, while this baseline MMSE cutoff is increased to $28$ and $30$ for eMCI and lMCI individuals, respectively. Below these respective cutoffs, CN, eMCI, and lMCI individuals have the highest probability of being allocated to the red cluster (apart from lMCI individuals with baseline MMSE of 20 that are allocated to the green cluster with highest probability). Instead, AD individuals have the highest probability of belonging to the red cluster for baseline $\text{MMSE}\geq 25$ and to the green cluster otherwise. We note that for CN individuals with low MMSE baseline (not observed), there is a small probability of allocation to a new (blue) cluster. 

\begin{figure}[!h]
\begin{center}
	\subfigure[ CN and APOE4=0]{\includegraphics[width=0.34\textwidth]{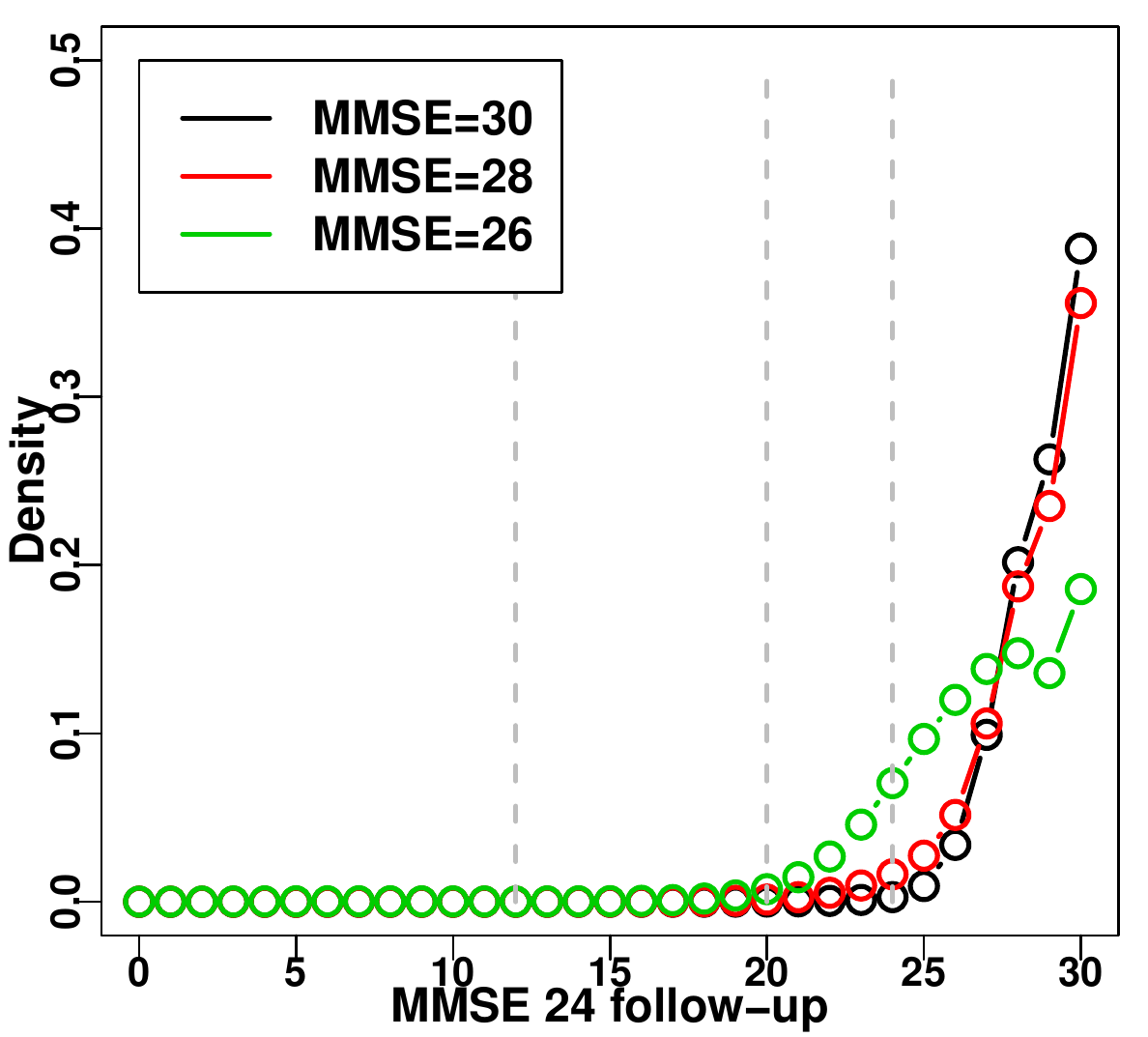}\label{fig:CN0_conditional}}
	\subfigure[ CN and APOE4=2]{\includegraphics[width=0.34\textwidth]{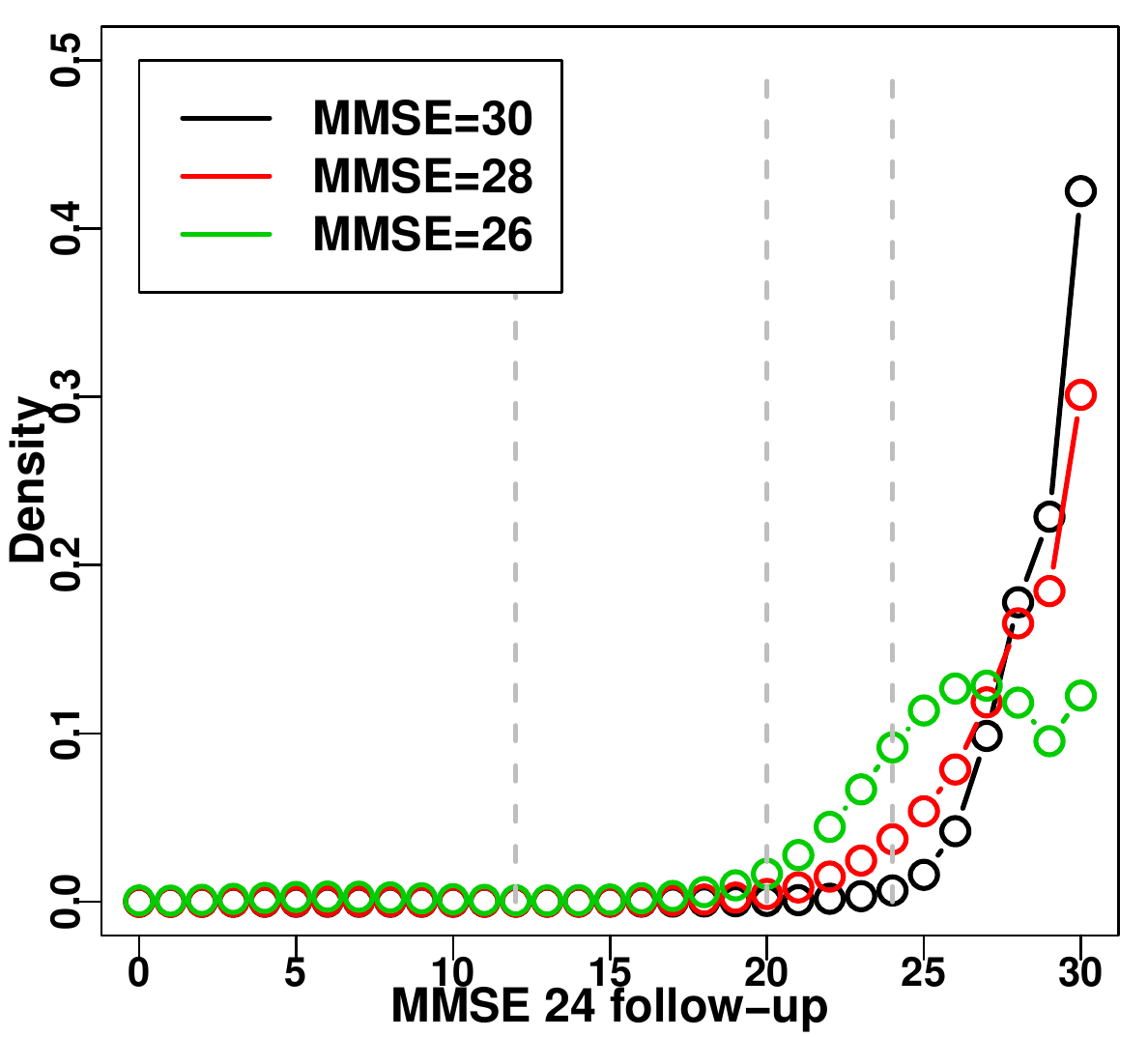}\label{fig:CN2_conditional}}
	\\ \vspace{-4mm}
	\subfigure[ eMCI and APOE4=0]{\includegraphics[width=0.34\textwidth]{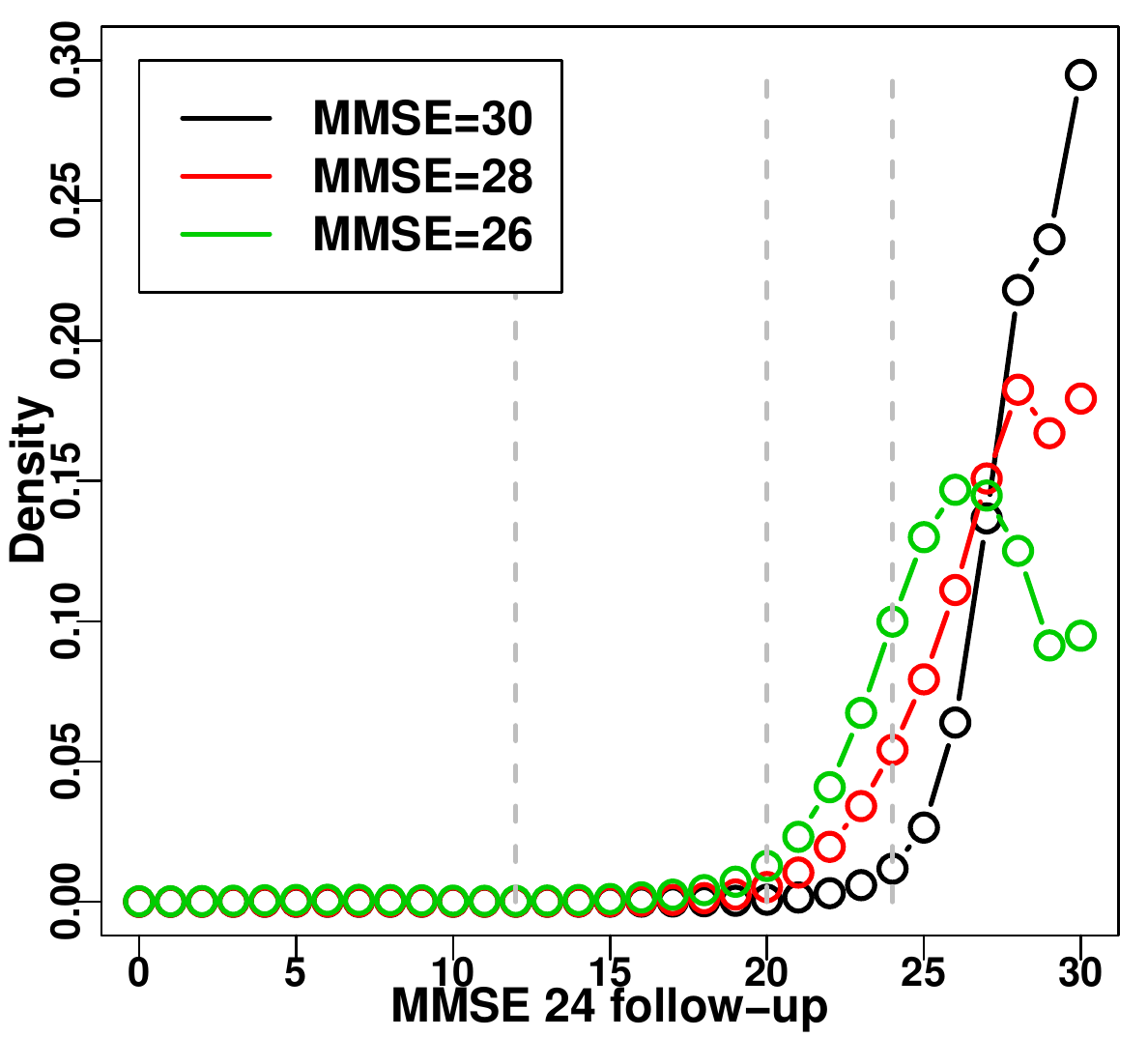}\label{fig:eMCI0_conditional}}
	\subfigure[ eMCI and APOE4=2]{\includegraphics[width=0.34\textwidth]{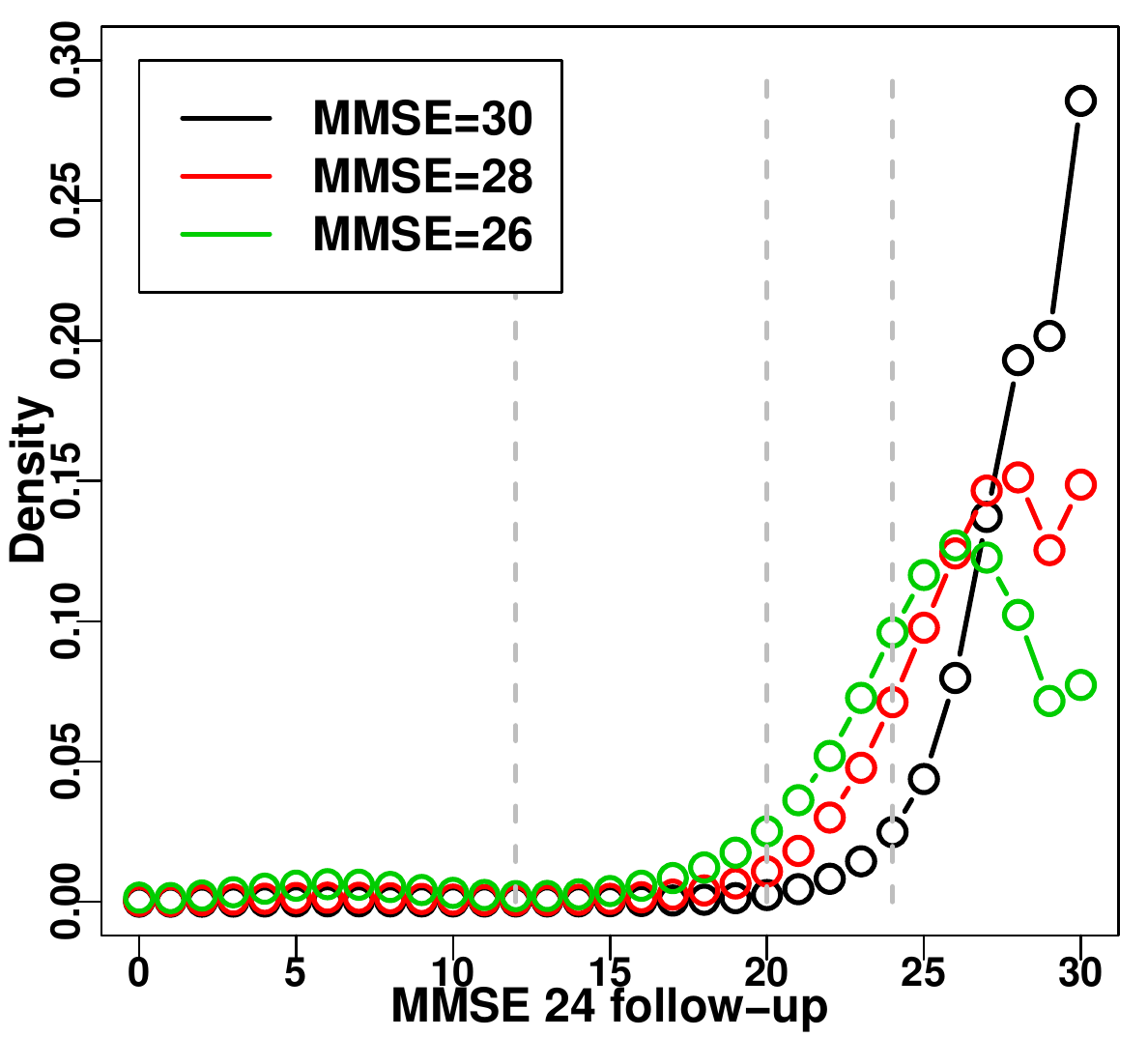}\label{fig:eMCI2_conditional}}
	\\ \vspace{-4mm}
	\subfigure[ lMCI and APOE4=0]{\includegraphics[width=0.34\textwidth]{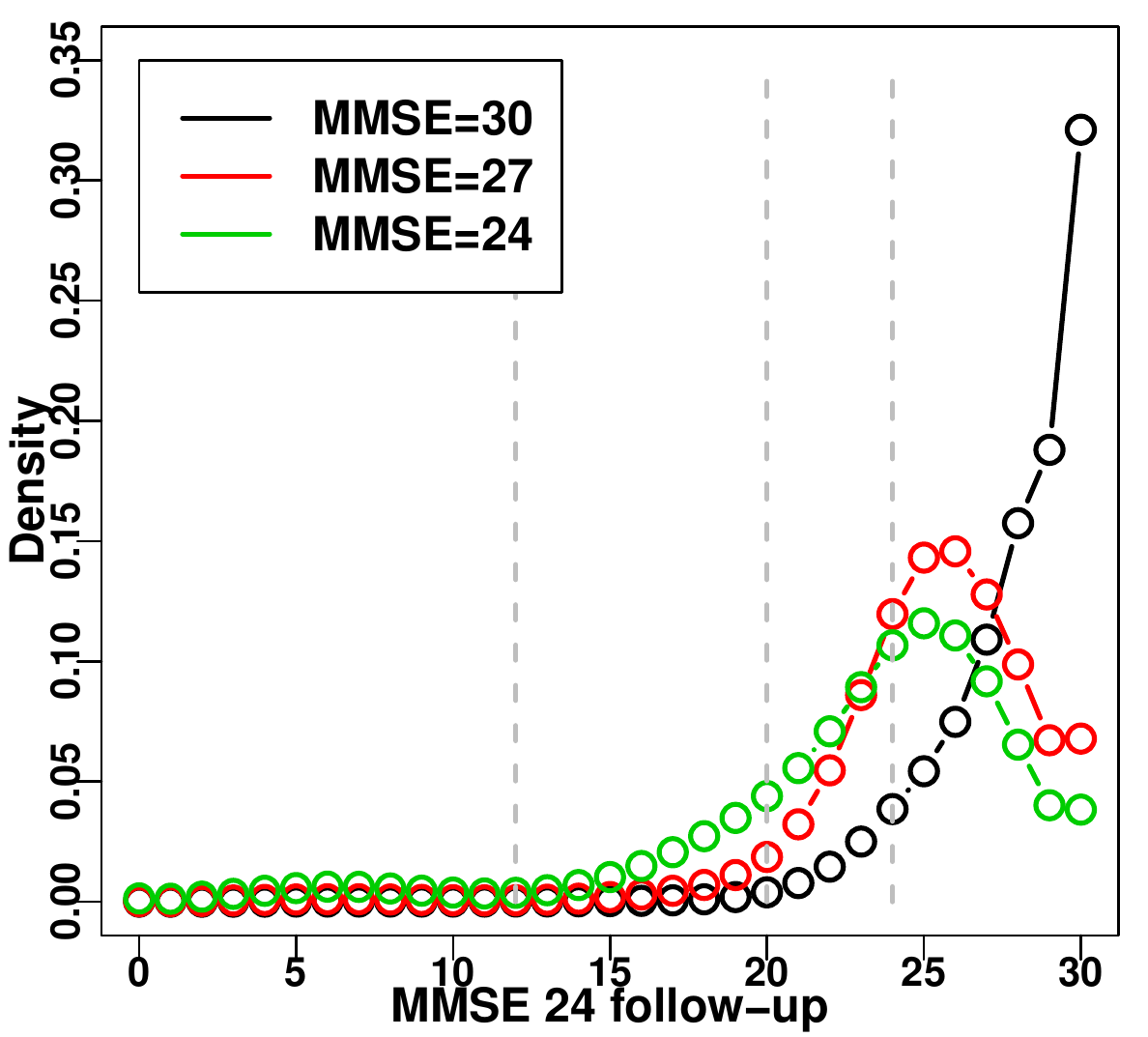}\label{fig:lMCI0_conditional}}
	\subfigure[ lMCI and APOE4=2]{\includegraphics[width=0.34\textwidth]{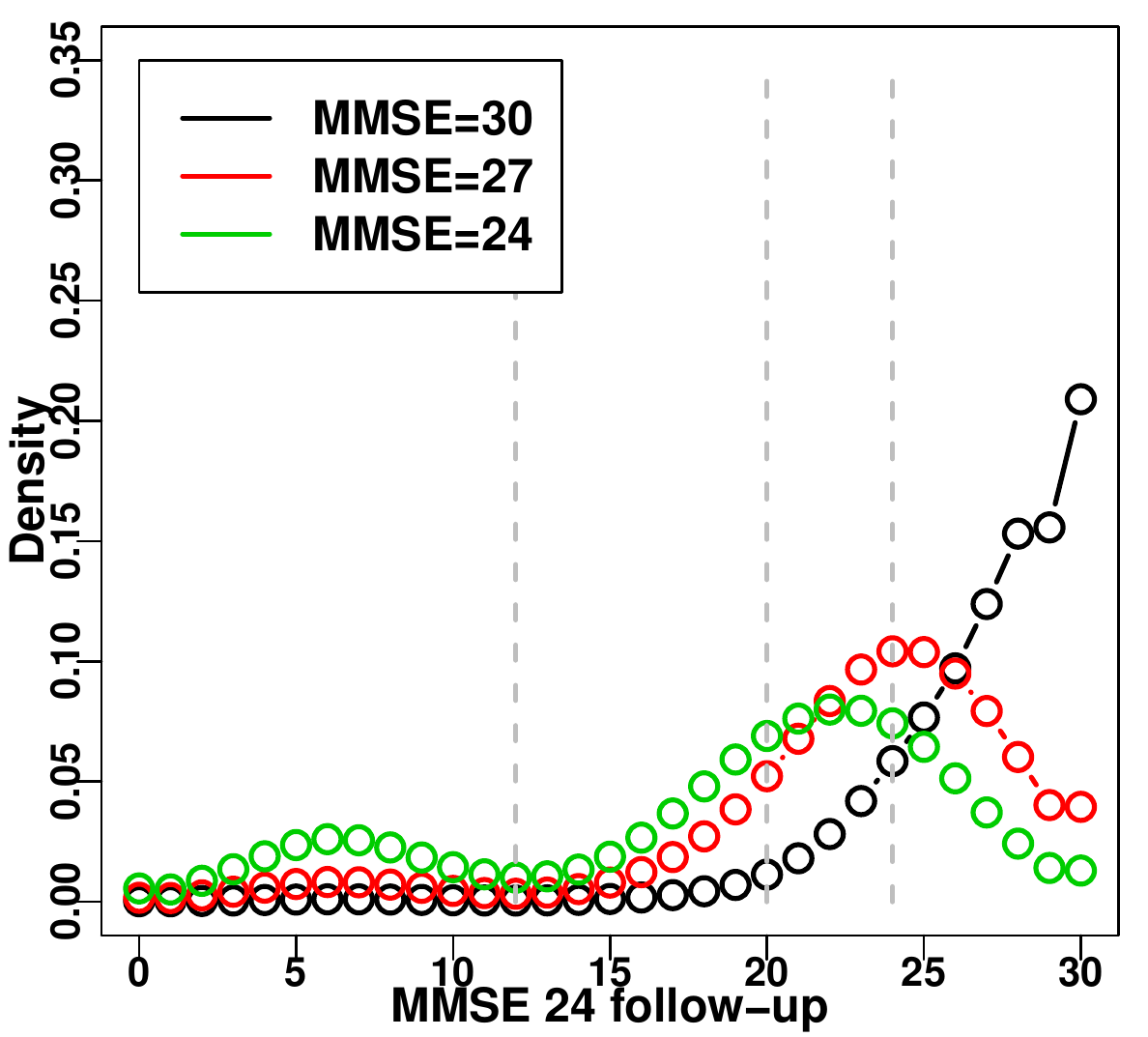}\label{fig:lMCI2_conditional}}
	\\ \vspace{-4mm}
	\subfigure[ AD and APOE4=0]{\includegraphics[width=0.34\textwidth]{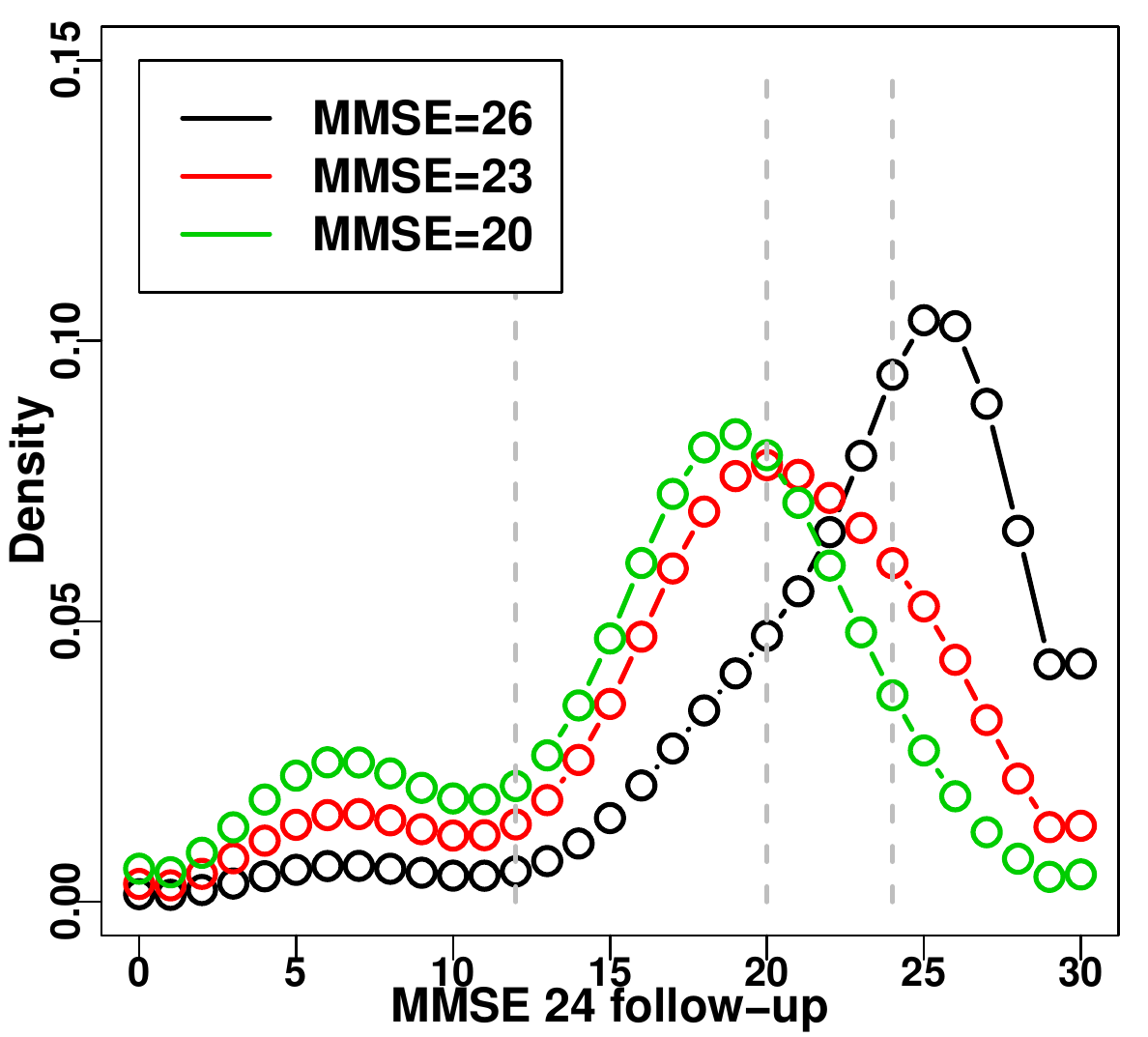}\label{fig:AD0_conditional}}
	\subfigure[ AD and APOE4=2]{\includegraphics[width=0.34\textwidth]{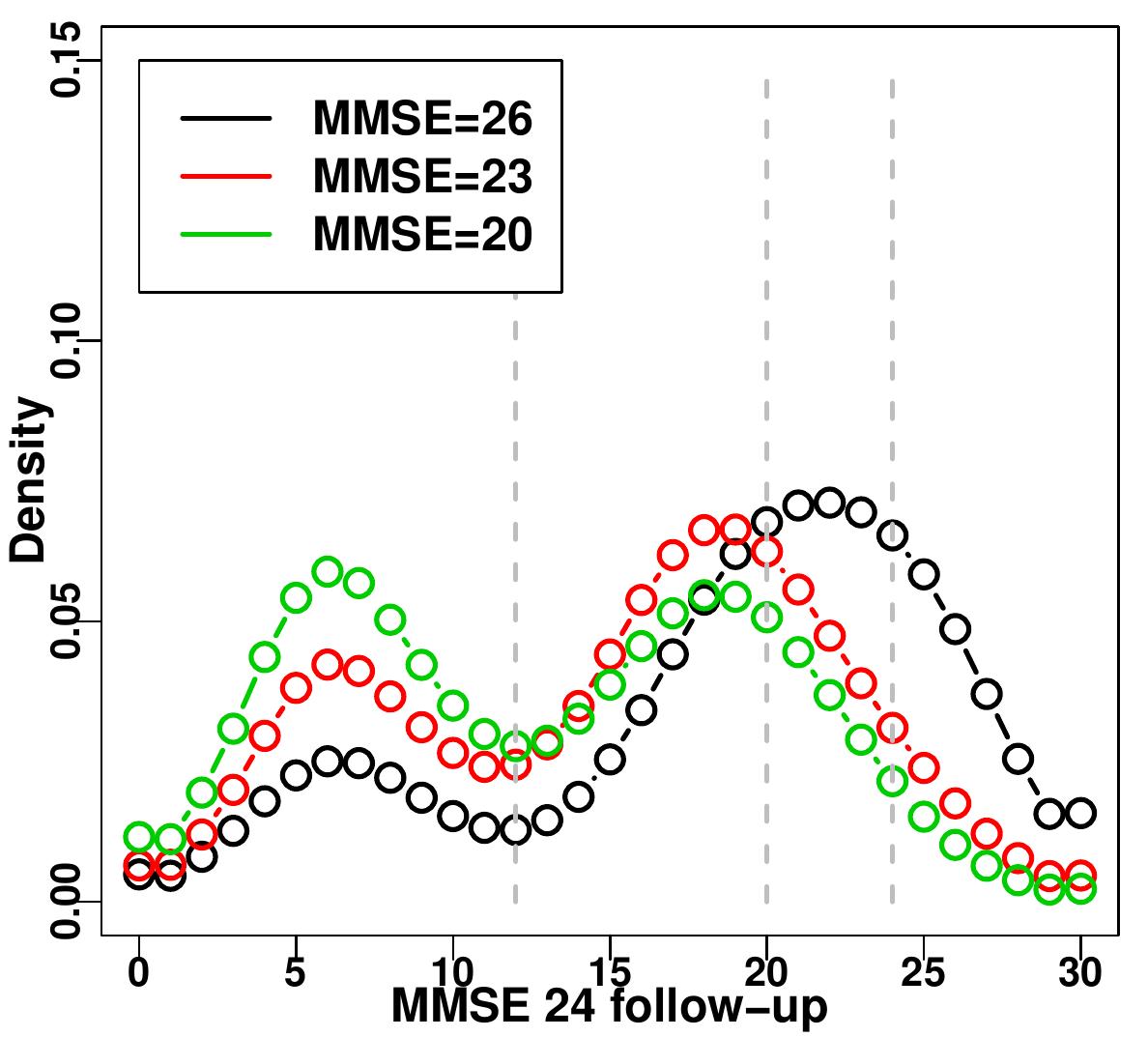}\label{fig:AD2_conditional}}
	\vspace{-4mm}
	\caption{Alzheimer's challenge. Predictive distribution of MMSE 24-month follow-up for different combinations of MMSE baseline, diagnosis, and APOE4, with other inputs marginalised for the enriched model. Columns represent APOE4 types of 0 and 2, whilst rows represent diagnosis.  Dashed lines indicate established cutoffs for MMSE: $\geq 25$ suggests no dementia; $20-24$ suggests mild dementia; $13-19$ suggests moderate dementia; $\leq 12$ suggests severe dementia.}
	\label{fig:adni_predictiveEME2}
			\end{center}
\end{figure}

Figure  \ref{fig:adni_predictiveEME2} shows how the predictive densities of MMSE follow-up scores change given different combinations of baseline MMSE, diagnosis, and APOE4. For CN individuals, the differences between APOE4 type are minor and the posterior mass is very concentrated on high follow-up MMSE scores given high baseline MMSE scores. More evident differences between APOE4 type are visible for more severe diagnosis, and in general, we observe a greater decrease in follow-up scores with more uncertainty for more severe dementia and increased APOE4. In particular, for AD patients that are carriers of APOE4, there is a visible probability of progressing to severe dementia (MMSE$\leq12$), that increases with decreased baseline MMSE.

\small
\bibliographystyle{plainnat}
\bibliography{bibFile}